%% file: corl_2026_template_cameraready/example_template.tex
\documentclass{article}

\PassOptionsToPackage{sort&compress}{natbib}
\PassOptionsToPackage{pagebackref=true}{hyperref}

\usepackage[final]{corl_2026} 
\usepackage{amsmath}  \usepackage{amssymb}  \usepackage{booktabs} \usepackage{pifont}   \usepackage{float}
\usepackage{graphicx} \usepackage{pgfplots} \pgfplotsset{compat=1.18}
\usepackage{verbatim} \usepackage{fancyvrb} \usepackage{listings} \lstdefinestyle{promptbox}{
  basicstyle=\footnotesize\ttfamily,
  breaklines=true,
  breakatwhitespace=false,
  breakindent=0pt,
  frame=single,
  columns=fullflexible,
  keepspaces=true,
  showstringspaces=false,
  upquote=true,
  literate={—}{{---}}1
    {→}{{$\rightarrow$}}1
    {…}{{\ldots}}1
    {’}{{'}}1
    {‘}{{`}}1
    {“}{{``}}1
    {”}{{''}}1,
}

\input{preamble}

\hypersetup{pdftitle={An Agentic Framework for Proactive Robots with Procedural Assistance Memory},pdfauthor={Dohyun Lee, Minho Park, Jeonghoon Park, Dongyoon Hwang, Junha Hyung, Sungwon Hwang, Byungkun Lee, Jaegul Choo},pdfsubject={Proactive robot assistance with Procedural Assistance Memory},pdfkeywords={Proactive Robot, Agentic Framework, Self-Evolving Memory}}

\usepackage[dvipsnames]{xcolor} 
\usepackage{dsfont}

\title{An Agentic Framework for Proactive Robots with Procedural Assistance Memory}

\author{
Dohyun Lee$^*$, Minho Park$^*$, Jeonghoon Park, Dongyoon Hwang, Junha Hyung, \\
\textbf{Sungwon Hwang, Byungkun Lee, Jaegul Choo} \\[3pt]
KAIST AI \\[2pt]
{\small \texttt{\{aiclaudev, m.park\}@kaist.ac.kr}} \\
{\small $^*$ indicates equal contribution.}
}

\begin{document}
\maketitle

\begin{abstract}
Proactive robots should anticipate user needs and provide assistance before being explicitly asked. A straightforward approach is to infer human intentions from commonsense knowledge, but the same observed behavior can imply different needs across users, potentially leading to unwanted assistance.
We observe that everyday routines in the real world exhibit recurring patterns, so learning these patterns can reduce ambiguity in proactive decisions.
Based on this, we propose a memory-based agentic framework in which a VLM task predictor
reads a \emph{Procedural Assistance Memory} (PAM) alongside the current observation
to decide when to assist and what task to perform, and a VLA executor performs the predicted task.
PAM is structured as a finite-state machine over activity phases, per-state assistance
rules, and event-driven transitions; a reflection VLM revises PAM from each day's
interaction trace without weight updates or retraining.
In our 5-day desk-assistance evaluation, memory-free baselines remain conservative, with limited recall on proactive tasks.
Our system improves consistently with accumulated experience, achieving the highest
precision, recall, and F1 and reducing first-fire delay from approximately 5 seconds
on Day 1 to 1 second by Day 5.
Our approach also adapts to changes in the user's routine through continued memory reflection, recovering from memory--routine misalignment after a single memory update.
The project page is available at \url{https://DAVIAN-Robotics.github.io/PAM}. \end{abstract}


\section{Introduction}
\label{sec:intro}

A truly useful assistant in everyday life does not only respond to requests, but also anticipates when help would be valuable.
This may involve handing over an object at the right moment, putting away an item after it has been used, or preparing something that will be needed in the next step.
By contrast, robot assistants are still commonly evaluated as systems that execute explicit user commands~\citep{openvla, pi05}.
For robots to play the role of capable assistants in household, and home-office settings, they must move beyond waiting for instructions and learn to recognize when intervention would be helpful.
This has motivated growing interest in \emph{proactive robot assistance}: robots that infer when help may be useful and act in advance of explicit commands.

One line of work in proactive robot assistance predicts task progress in structured sequential tasks~\citep{pace, liu2023proactive, stouraitis2023predictive}. However, these approaches are largely confined to robots that perform a single task by following a fixed protocol (e.g., chair assembly), and are ill-suited to generalist robots tasked with open-ended housekeeping. Recently emerging generalist Vision-Language-Action models (VLAs) \citep{openvla, gr00t, black2024pi_0, pi05} may appear to offer an alternative, but robotics datasets remain scarce across the diverse scenarios such settings demand. As a result, end-to-end models that map perception directly to action face clear limitations even when equipped with reasoning, and proactive tasks, where intent is often underspecified, have received little attention. 
In response, recent work has adopted a System 1–2 design~\citep{kahneman2011thinking} that explicitly separates a high-level reasoning module from a VLA that executes atomic tasks \citep{cui2025openhelix, song2025hume, chen2025fast, han2024dual}. 
This separation simplifies learning, since the VLM decides when and what to help with while the VLA need only carry out the prescribed action. We adopt this structure, which in turn places the burden of proactivity squarely on the high-level VLM.

\begin{figure}[!t]
\centering
\includegraphics[width=\linewidth]{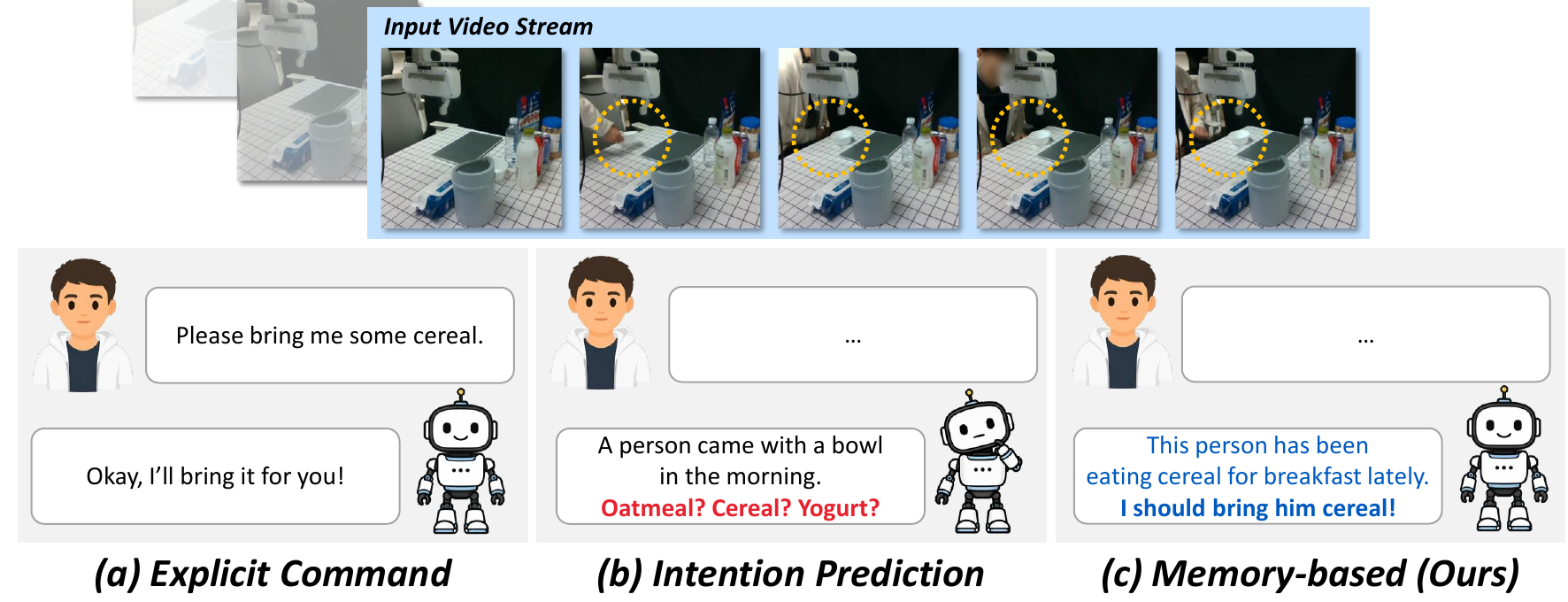}
\caption{
\textbf{Motivation for memory-based proactive assistance.} Visually similar observations map to different correct actions depending on user-specific habits, and our PAM memory resolves this ambiguity by conditioning the task predictor on accumulated behavioral context.
}
\label{fig:1_intro_motivation}\vspace{-16pt}
\end{figure}

The central challenge, then, is accurately predicting human intention at the high level~\citep{buyukgoz2022two}, and recent approaches leverage pretrained, large-scale VLMs to do so in open-ended settings~\citep{eventdriven, roboomni, intentionvla}.
Yet zero-shot prediction remains unreliable when the same observable behavior corresponds to different desired outcomes across users: one person setting out a mug wants coffee, another wants hot water for tea (\Cref{fig:1_intro_motivation}).
A fixed, pretrained predictor cannot resolve such ambiguity, because the missing information is not in the scene but in the individual user's habits.

Crucially, this missing information can be learned during deployment. In housekeeping settings, an individual's behavior is highly repetitive, so even a handful of past interactions reveal stable preferences, and corrective feedback gathered once can be reused whenever a similar situation recurs. This suggests that a proactive robot should not treat each deployment as zero-shot, but should accumulate experience about its specific user over time. We therefore investigate a test-time learning approach for the high-level VLM, building a robot that personalizes quickly from few interactions, continues adapting after deployment, and carries user feedback forward into future behavior.

To this end, we introduce \emph{Procedural Assistance Memory} (PAM), a continually updated deployment-time memory that records when assistance is needed and what assistive task should be issued for a particular user.
PAM structures this memory as a finite-state machine over the user's activity, where states represent activity phases, transitions represent progress between phases, and per-state assistance entries store the assistive opportunities associated with each phase.
In our five-day scenario, the user repeatedly performs a daily activity; PAM is updated across these days so that the robot can provide increasingly appropriate assistance. Our contributions are threefold:

\begin{enumerate}[leftmargin=1.5em]
    \item We propose a \textbf{memory-based agentic framework} for proactive robot assistance.
    By conditioning the high-level VLM on user-specific routine context, it decides both \emph{when} to assist and \emph{what} task to issue without explicit user requests.

    \item We propose \textbf{Procedural Assistance Memory} (PAM), a memory following a finite-state-machine structure with activity phases, transitions, and per-state assistance entries.
    It uses user and activity context to reduce ambiguity in assistance needs.

    \item We develop a \textbf{self-evolving memory loop} whose reflector revises PAM offline after each day using observed behavior and user feedback.
    This enables the robot to learn the user's routine, adapt to changes, and incorporate user feedback without retraining.
\end{enumerate}

\section{Related Work}
\label{sec:related}

\subsection{Vision-Language-Action Models}
\label{sec:rw_vla}
Robot manipulation has advanced through visuomotor imitation-learning policies~\citep{act, chi2023diffusion} and generalist VLAs~\citep{openvla, black2024pi_0, pi05, gr00t}, with subsequent work exploring chain-of-thought reasoning~\citep{ecot, cotvla, rth}, spatial grounding~\citep{tdvla, spatialvla}, and world-model predictions~\citep{vpp, dreamvla, starry}. 
However, jointly optimizing action execution and high-level reasoning in a single network causes the two objectives to compete under limited robot data.
In response, recent work adopts a System 1--2 design~\citep{kahneman2011thinking} that explicitly separates a slow reasoning module from a fast action executor~\citep{han2024dual, song2025hume, chen2025fast, cui2025openhelix, gr00t}. The reasoning module interprets the scene and reasons about how to solve the task, while the executor generates low-level actions for robot control conditioned on the high-level outputs. We build on this division of roles, using a VLM to predict proactive tasks and a VLA to execute the selected atomic skills.

\subsection{Proactive Assistance in Embodied HRI}
\label{sec:rw_proactive}

A prominent line of work on proactive robots studies assistance in structured sequential tasks, where the human activity is already organized into known subtasks.
In these settings, robots estimate task progress or subtask completion to provide help at the right moment~\citep{liu2023proactive, stouraitis2023predictive}, such as handovers in assembly~\citep{pace} or ingredient delivery in cooking~\citep{litsouschef}. 
This direction is especially effective when the task sequence, progress measure, and assistance opportunities are known in advance.
Another line of work models object usage patterns in daily environments to anticipate future assistance needs~\citep{stom,predicting-routine-object-usage,longproactive}.
These methods move beyond single structured tasks, but they typically require substantial interaction logs or curated observations for training, limiting their flexibility across diverse everyday routines and rapidly changing user behavior.

With the recent progress of foundation models, proactive assistance has begun to move toward more general intent inference.
Some systems rely on non-verbal cues such as visual scene changes~\citep{eventdriven}, human-object interactions~\citep{hoi4abot, rahimi2026intentvlm}, or gaze~\citep{gazeguided, gaus2026act} to infer assistive intent, while others draw on verbal cues such as indirect instructions~\citep{intentionvla}, speech~\citep{roboomni}, or dialogue~\citep{sarch2025grounding,helper}.
However, intent inference from momentary cues alone remains ambiguous: the same cue can imply different assistance needs depending on the user, context, and activity history.
We address this by conditioning the proactive decision on a user-specific memory accumulated across days, so that assistance is grounded not only in the current cue but also in how this user has typically acted and been assisted.

\subsection{Evolving Memory for Test-Time Adaptation}
\label{sec:rw_memory}

Recent LLM-agent systems use external memory to support test-time adaptation without weight updates, allowing an agent to accumulate and reuse experience across episodes.
These works span verbal self-reflection~\citep{reflexion}, reusable skill libraries~\citep{voyager}, workflow or experience extraction~\citep{expel,awm}, and long-term memory~\citep{memorybank,genagents}.
However, their memories are generally designed for text-centric agents, games, or instruction-following tasks, rather than for deciding when a robot should proactively assist during an evolving human activity.
PAM organizes assistive opportunities into activity phases, per-state assistance entries, and transitions, preserving the context in which each opportunity arises. The task predictor uses this context to decide when assistance is needed and what assistive task should be issued given the user's current activity and learned routine.

\begin{figure}[!t]
\centering
\includegraphics[width=\linewidth]{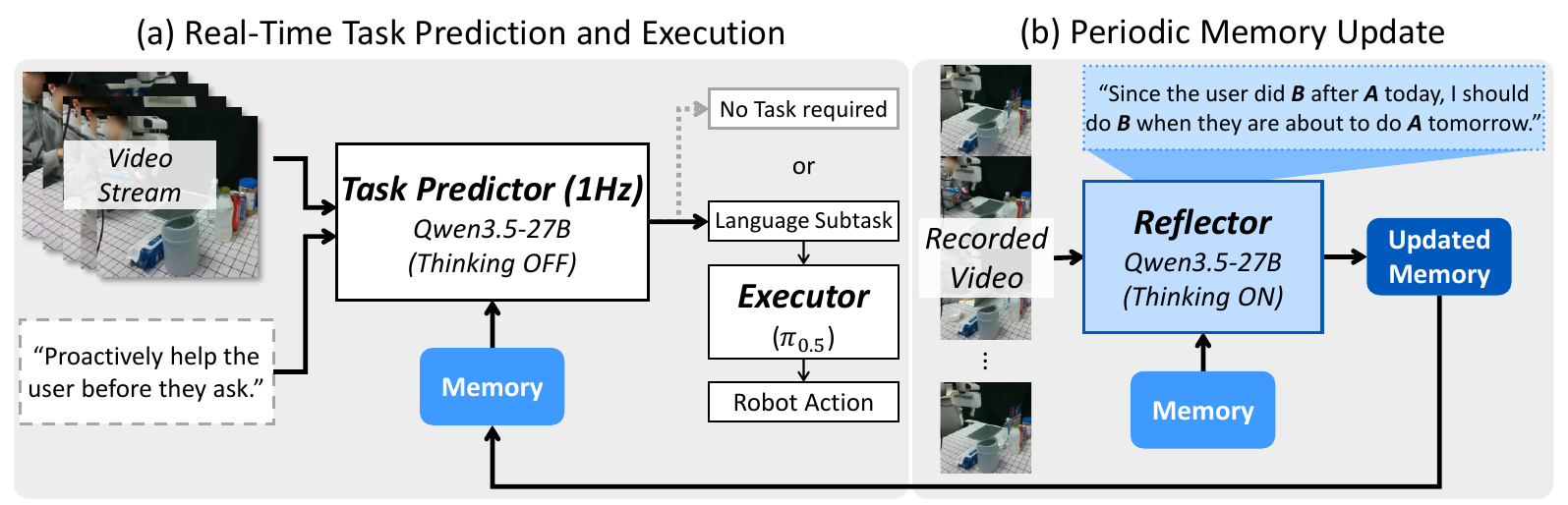}
\vspace{-0.3cm}
\caption{
\textbf{Overview of our framework.} (a) A VLM task predictor reads PAM and recent frames to select assistive tasks for the VLA executor. (b) At the end of each day, a VLM memory reflector updates PAM from the day's video by adding or revising states, trigger-action pairs, and transitions.
}
\vspace{0.2cm}
\label{fig:3_method_overview}
\end{figure}

\section{Method}
\label{sec:method}

Our goal is to learn and store a user's routine in memory, then use it to decide when to assist and what task to perform in real time. The system should also adapt quickly to routine changes without retraining. To this end, we design a proactive assistance system built around Procedural Assistance Memory (PAM) (\Cref{fig:3_method_overview}). A VLM task predictor uses observations and PAM to predict whether assistance is needed and, if so, selects a task for the VLA executor to perform. After each day, a VLM memory reflector updates PAM offline using the day's recorded video.

\newpage
\subsection{Real-Time Task Prediction and Execution via System 1--2 Architecture}
\label{sec:method_tasksetter}
Our task predictor serves as System~2, selecting assistive tasks without task-specific user instructions. At each 1-Hz step, it receives the latest $W$ video frames, PAM as text, and the fixed instruction, \textit{``Proactively help the user before they ask.''} It uses the routine stored in PAM to interpret the observations and outputs either a task expressed in natural language or \texttt{NO\_ACT}. Explicit thinking mode is disabled for real-time inference.

The predicted task is passed to the VLA executor, which serves as System~1 and generates low-level robot actions. For example, the predictor selects \textit{``bring milk''} as the user pours cereal. The executor then carries out this task by picking up the milk and placing it next to the user.

\subsection{Procedural Assistance Memory (PAM)}
\label{sec:method_mem}

\begin{wrapfigure}{r}{0.4\textwidth}
\vspace{-0.5cm}
\centering

\includegraphics[width=0.4\textwidth]{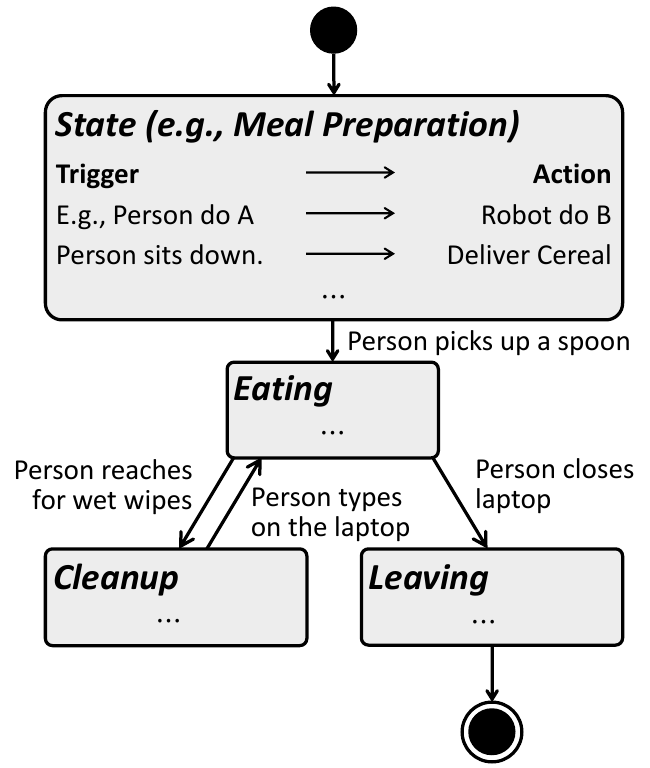}
\vspace{-0.3cm}
\caption{\textbf{An example of PAM}, comprising states, trigger--action pairs, and transitions.}
\label{fig:method:fsm}
\vspace{-0.2cm}
\end{wrapfigure}

PAM should be designed to store information about when to assist the user and what assistance to provide. The simple solution is a flat list of trigger--action pairs, but this provides limited context for interpreting cues within the user's routine. For example, an empty bowl on the desk may indicate that the robot should bring cereal when the user is about to eat or put the bowl away when they have finished eating. To reduce this ambiguity, we organize PAM using a finite-state-machine (FSM) structure: states describe activity phases, per-state trigger--action pairs link cues to assistive tasks, and transitions specify the events that connect phases (\Cref{fig:method:fsm}). Transitions capture observable changes in the activity; for instance, picking up a spoon indicates the shift from meal preparation to eating. A trigger should occur before the user begins an action that the robot can perform for them. For example, when the user sits down holding an empty bowl, the robot brings cereal before the user fetches it.\par PAM is serialized as structured text (e.g., JSON) in the predictor's prompt. The memory provides routine context, while the VLM selects assistive tasks. The VLM reads all state descriptions, trigger--action pairs, and transitions alongside recent observations. The predictor uses PAM as contextual guidance to assess whether assistance is appropriate given the current observations. It selects a task only if it is immediately executable. Otherwise, it outputs \texttt{NO\_ACT} to indicate that no assistance is needed.

\newpage
\subsection{Offline Memory Reflection and Update}
\label{sec:method_workflow}
At the end of each day, a memory reflector updates PAM offline with reasoning enabled. It receives sampled frames from the day's video, the previous PAM as text, a summary of the day's action history, and user feedback when available.

As illustrated in \Cref{fig:3_method_overview}(b), the recording includes actions that the user performed without robot assistance. The reflector examines these segments to identify missed opportunities and infer recurring patterns. For example, seeing the user bring out an empty bowl and then pour cereal can lead it to infer that the robot should bring cereal when the user brings out an empty bowl. It adds the resulting assistance rule to PAM and revises existing states, rules, or transitions when observations or feedback indicate a change in the routine.

The reflector returns the updated PAM for use on the following day. PAM remains fixed during each day; adaptation occurs through these daily revisions. The prompts used for the task predictor and memory reflector are provided in Appendices~\ref{appn:infer_prompt} and~\ref{appn:reflect_prompt}.

\section{Experiments}
\label{sec:exp}

In the following sections, we aim to resolve the following research questions:
\begin{enumerate}[leftmargin=2.5em, label=\textbf{RQ\arabic*.}, ref=RQ\arabic*]
    \item Does PAM improve proactive task prediction over memory-free baselines? (\Cref{sec:exp:vlm})
    \item Does structuring memory as an FSM yield greater improvements across days than a flat list of rules? (\Cref{sec:exp:vlm})
    \item Does continued reflection enable adaptation to changes in the user's routine? (\Cref{sec:exp:memory})
    \item Does the System 1--2 decomposition improve task execution success compared to monolithic VLAs? (\Cref{sec:exp:vla})
\end{enumerate}

\begin{figure}[!t]
\centering
\includegraphics[width=\linewidth]{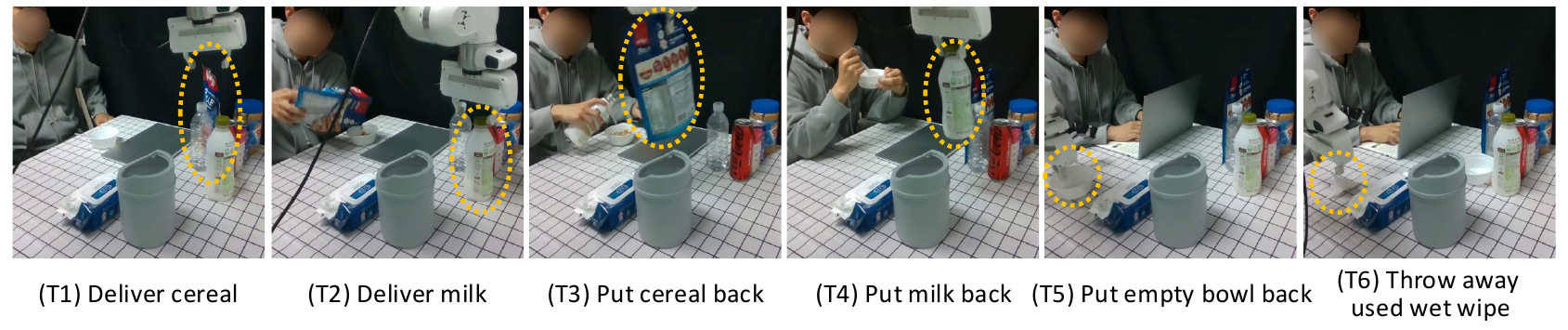}
\vspace{-0.6cm}
\caption{
\textbf{Tasks and dataset.} The six proactive tasks form a continuous desk-activity sequence: (T1) Cereal$\rightarrow$Person, (T2) Milk$\rightarrow$Person, (T3) Cereal$\rightarrow$Storage, (T4) Milk$\rightarrow$Storage, (T5) Bowl$\rightarrow$Storage, (T6) Used wet wipe$\rightarrow$Trash. 
}
\label{fig:exp:dataset}
\end{figure}

\subsection{Experimental Setup}
\label{sec:exp:setup}

\paragraph{Tasks.}We evaluate proactive assistance in a morning desk routine (\Cref{fig:exp:dataset}). The user arrives at the desk carrying an empty bowl, sits down, pours cereal and then milk into the bowl, and eats breakfast. After eating, the user sets the empty bowl aside and starts working on a laptop, then wipes their mouth with a wet wipe, leaves the used wipe on the desk, and eventually leaves the desk. To assist the user throughout this routine, the robot must perform six proactive tasks, namely (T1) deliver cereal (Cereal$\rightarrow$Person), (T2) carry over milk (Milk$\rightarrow$Person), (T3) place back cereal (Cereal$\rightarrow$Storage), (T4) place back milk (Milk$\rightarrow$Storage), (T5) place back the empty bowl (Bowl$\rightarrow$Storage), and (T6) throw away the used wet wipe (Used wet wipe$\rightarrow$Trash). The tasks include both preparing items for the user's next activity and clearing items after use. Thus, the same objects can require different robot actions as the routine progresses, making the timing of assistance part of the task. We sample frames from the recorded routine videos at 1 fps and label each frame with the task that the robot should perform to assist the user. We use another scenario to evaluate adaptation to routine changes in Section~\ref{sec:exp:memory}, with details of this scenario provided in Appendix~\ref{appn:another_scenario}.

\paragraph{Evaluation protocol.}
We formalize proactive task prediction as a per-frame classification problem. 
Importantly, the label set comprises not only the six tasks above but also a \texttt{NO\_ACT} class indicating that no assistance is needed.
We report precision, recall, and F1 score per class (the six proactive tasks and \texttt{NO\_ACT}) along with their macro averages. We also report the first-fire delay, measured from the time at which proactive assistance should be provided until the model first predicts the corresponding task. If the model does not predict a task, we instead measure the delay from the time assistance should be provided until the user completes the task without robot assistance. Together, these metrics measure whether assistance is appropriate and whether it is predicted early enough to be useful.
To assess the effectiveness of our PAM, we conduct a multi-day evaluation over 5 days (Fig.~\ref{fig:5days}), in which the PAM persists across days and is updated from feedback from previous days.

\paragraph{Implementation details.}
We use Qwen 3.5~\citep{qwen3.5} as the VLM for both the task predictor and the reflector. The reflector runs once at the end of each day with thinking mode enabled to reason over the day's observations and revise PAM for use on the following day. The task predictor runs with thinking mode disabled to support timely assistance. At each prediction step, it predicts a task or \texttt{NO\_ACT} from a window of $W{=}5$ frames sampled at stride $2$ from the 1\,fps stream. Each reflector call takes approximately 1 minute, while each task predictor call takes approximately 1 second. For the executor, we collect 10 demonstrations per task and use them to finetune $\pi_{0.5}$~\citep{pi05}. Generation hyperparameters are fixed across all conditions and reported in Appendix~\ref{app:eval-settings}. 

\begin{figure}[!t]
\centering
\includegraphics[width=0.95\linewidth]{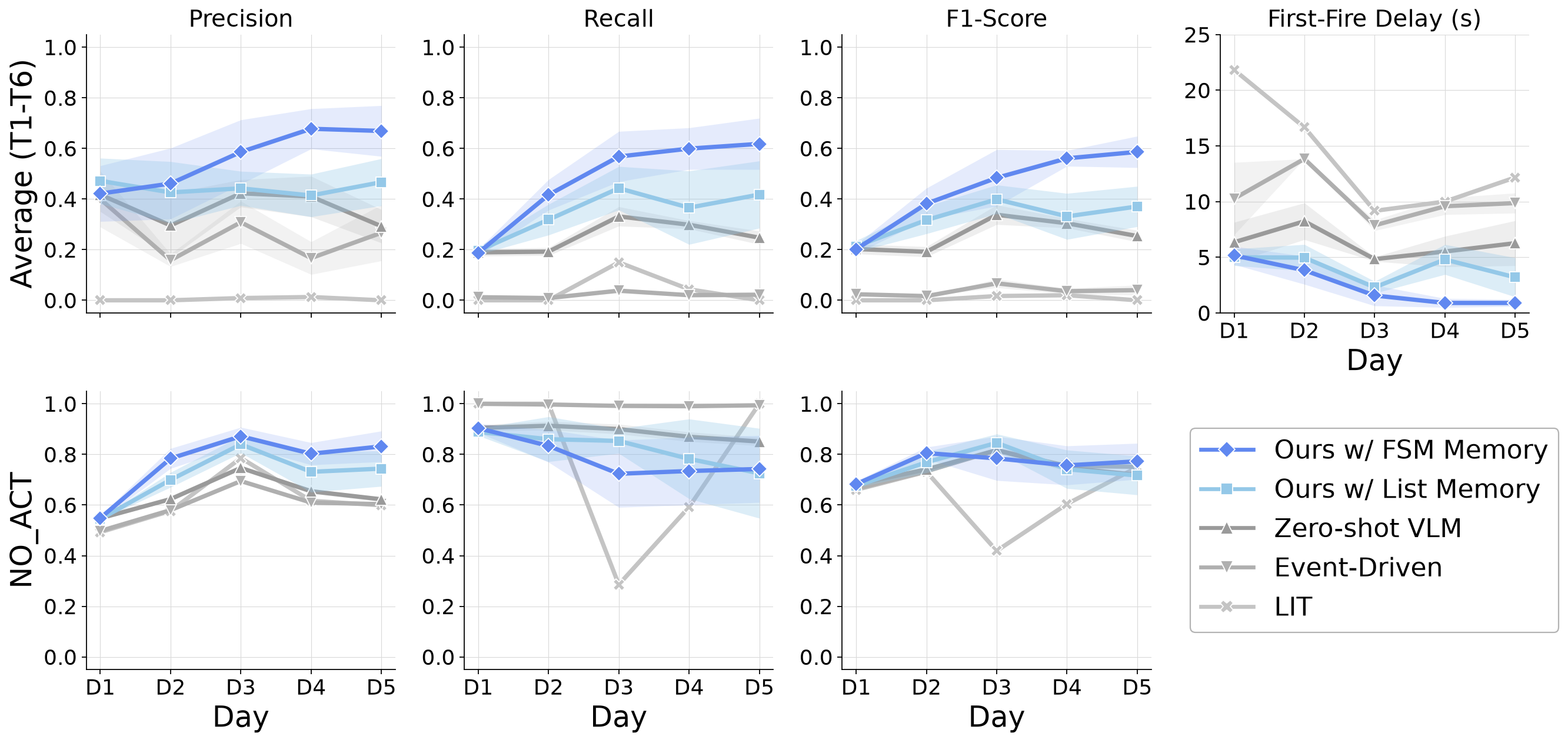}
\caption{
\textbf{Multi-day proactive task prediction (Days 1--5).} Memory-free baselines show limited T1--T6 recall overall, while Ours (FSM Memory) consistently improves across days, achieving the highest F1 and reducing first-fire delay from ${\sim}5$\,s on Day 1 to ${\sim}1$\,s by Day 5.
}
\label{fig:exp:vlm}
\vspace{-0.25cm}
\end{figure}

\paragraph{Baselines.} 
For the proactive task prediction baseline, we provide following:

\begin{itemize}[leftmargin=1.5em]
    \item \textbf{LIT}~\citep{litsouschef}. Builds a task graph over detected objects and estimates the current progress along the graph to predict the next task. This represents structured, progress-tracking proactivity.
    \item \textbf{Event-driven VLM}~\citep{eventdriven}. Predicts the next task only when a scene change is detected via optical flow, by comparing the pre- and post-change frames. This represents reactive, change-triggered proactivity without any cross-day memory.
    \item \textbf{Zero-shot VLM}. A state-of-the-art VLM that predicts the next proactive task directly from the current observation, without any adaptation. This isolates the benefit of our test-time memory against a strong but static predictor.
    \item \textbf{Ours w/ List Memory}. A variant of our framework that replaces the FSM-structured memory with a flat list of trigger-action pairs accumulated over deployment, while keeping the rest of the pipeline identical. This isolates the contribution of the FSM structure itself.
\end{itemize}Additionally, we provide two VLA baselines, which we finetune on 10 episodes in which the robot performs all six assistance tasks throughout the user's routine. We compare their final task success rate on the proactive tasks:
\begin{itemize}[leftmargin=1.5em]
    \item $\boldsymbol{\pi_{0.5}}$. The state-of-the-art open-source pretrained VLA. We prompt it with ``Proactively help the user before they ask.'' at both training and test time throughout the entire scenario.
    \item $\boldsymbol{\pi_{0.5}}$ \textbf{(Reasoning)}. Since $\pi_{0.5}$ can decompose a high-level instruction into low-level atomic actions, we finetune it with ``Proactively help the user before they ask.'' as the high-level prompt and the corresponding atomic command (\eg, ``pick the cereal and place it next to the user'') as the low-level task, leveraging its reasoning capability.
\end{itemize}

\begin{figure}[!t]
\centering
\includegraphics[width=0.95\linewidth]{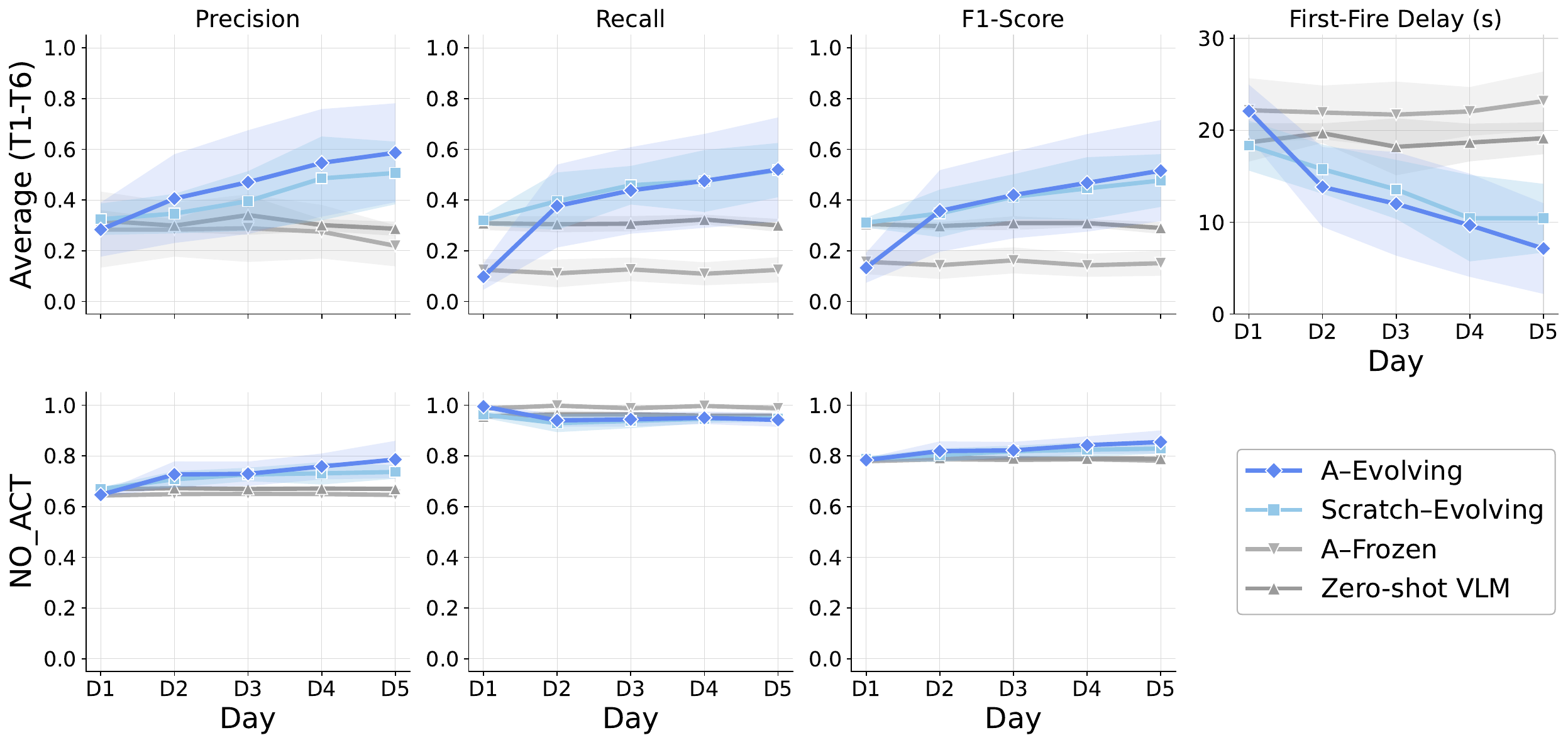}
\caption{\textbf{Memory adaptation (Scenario A$\rightarrow$B).} Scenario A refers to the scenario shown in Figure~\ref{fig:exp:dataset}, and Scenario B refers to the scenario shown in Figure~\ref{fig:scenario2_desc}. A-Frozen fixes A's Day-5 memory, while A-Evolving updates it at the end of each day in Scenario B. Scratch-Evolving builds memory from scratch on B; Zero-shot VLM uses no memory.}
\label{fig:exp:warmstart_A2B}
\end{figure}

\subsection{Proactive Prediction across Memory Designs}
\label{sec:exp:vlm}

\paragraph{Effectiveness of Memory.}
As shown in \Cref{fig:exp:vlm}, baselines without memory (LIT~\citep{litsouschef}, Event-Driven~\citep{eventdriven}, Zero-shot VLM~\citep{qwen3.5}) predict conservatively: lacking prior knowledge of the user's routine, they default to \texttt{NO\_ACT} whenever the observation is ambiguous, which inflates \texttt{NO\_ACT} recall but suppresses proactiveness on T1--T6.
LIT and Event-Driven rarely fire any proactive task, yielding near-zero recall, and the Zero-shot VLM plateaus from Day 2 onward with no mechanism to internalize recurring patterns.
All three remain essentially flat across the five days, confirming that without memory there is no path from repeated exposure to better proactive behavior.
In contrast, our memory-based variants (Ours w/ List, Ours w/ FSM) leverage repeated exposure to improve over the five days, though to different degrees, as we analyze next.

\paragraph{Ablation of Memory Design.}
Among the two memory designs, Ours w/ List Memory saturates early, since the flat list offers no abstraction over recurring states as it grows. In contrast, Ours w/ FSM Memory consistently improves across days, achieving the highest precision, recall, and F1 on T1--T6 while reducing first-fire delay from roughly 5 seconds on Day 1 to about 1 second by Day 5. This shows that the gains come from the FSM structure itself, not simply from adding memory. The trade-off is that FSM Memory's eagerness to act lowers its \texttt{NO\_ACT} recall relative to the conservative baselines, although a correspondingly higher \texttt{NO\_ACT} precision keeps its \texttt{NO\_ACT} F1 on par with the others.
Reducing false alarms further while preserving proactiveness remains an open direction. Per-task results are provided in Appendix~\ref{appn:per_task}.

\newpage
\subsection{Memory Adaptation and Evolution}
\label{sec:exp:memory}

\paragraph{Adaptation to Routine Changes.}
When a user's routine changes, previously learned assistance rules may become inappropriate, making rapid adaptation important. Our method addresses this by updating PAM through reflection without retraining the models. We refer to the cereal routine shown in Figure~\ref{fig:exp:dataset} as Scenario A. Scenario B involves a different routine (Figure~\ref{fig:scenario2_desc}). For example, the user takes vitamins instead of eating cereal. We evaluate how effectively and quickly PAM adapts over five target days when a user who has followed Scenario A suddenly switches to Scenario B.

Both A-Frozen and A-Evolving begin with Scenario A's Day~5 memory. A-Frozen keeps this memory unchanged, whereas A-Evolving updates it through reflection at the end of each day in Scenario B. Figure~\ref{fig:exp:warmstart_A2B} compares these conditions with Scratch-Evolving, which builds memory from scratch in Scenario B, and the Zero-shot VLM, which uses no memory. On Day~1, A-Evolving has lower task recall and F1 than the Zero-shot VLM, suggesting that memory from the previous routine can initially cause confusion in predicting proactive tasks in the new scenario. After a single reflection at the end of Day~1, however, its recall and F1 reach levels comparable to Scratch-Evolving on Day~2. They remain close over the subsequent days, while A-Frozen stays below the Zero-shot VLM. First-fire delay also decreases with continued reflection. These results show that PAM can quickly adapt an existing memory to a different routine, recovering from the initial mismatch after one memory update. Results for the reverse B$\rightarrow$A experiment are provided in Appendix~\ref{app:routine_change}.

\paragraph{Qualitative Memory Evolution.}

\begin{figure}[!t]
\centering
\includegraphics[width=0.95\linewidth]{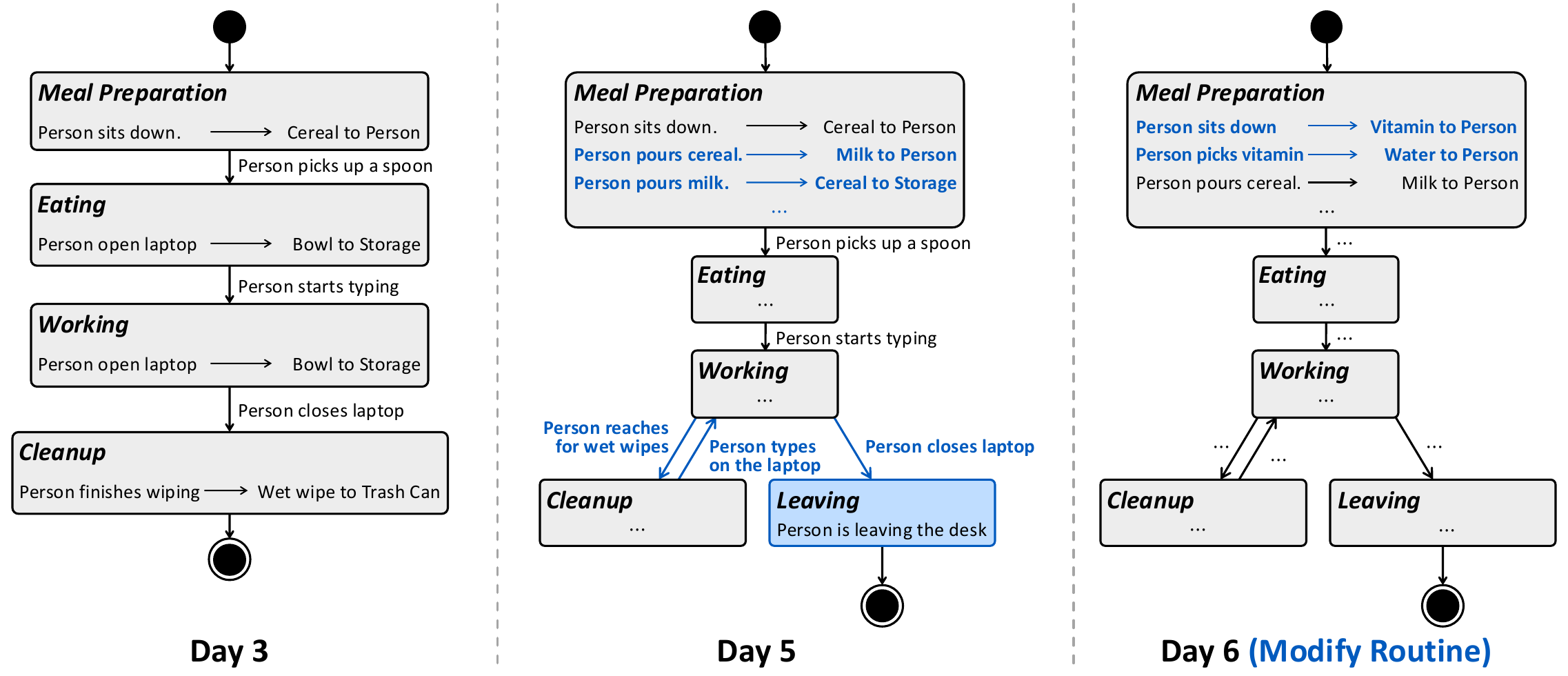}
\caption{
\textbf{Qualitative analysis of FSM-based Procedural Assistance Memory evolution.}
From Day 3--5, missing trigger-action pairs are recovered and per-state rules are completed. When the user's routine changes, end-of-day reflection revises the affected transitions and rules accordingly.
}
\label{fig:exp:memory}
\end{figure}

We qualitatively investigate the FSM-based Procedural Assistance Memory (PAM) in \Cref{fig:exp:memory}, which highlights both the efficiency and flexibility of our memory structure.
Comparing Day 3 and Day 5, the FSM-based PAM becomes increasingly accurate: trigger-action pairs that were missing on Day 3 are recovered on Day 5, with each state capturing its full set of triggers and actions (\eg, the pair ``person pours cereal'' $\rightarrow$ ``milk to person'' is added).
Our FSM-based PAM also adapts dynamically when the user's routine changes. For instance, even when the user sits down with a bowl as before, taking vitamins before eating the cereal that the robot has just delivered causes the transition ``person sits down'' $\rightarrow$ ``cereal to person'' to be revised, after reflection, to ``person sits down'' $\rightarrow$ ``vitamin to person''.

\vspace{-0.2cm}
\subsection{Comparison with Monolithic VLAs}
\label{sec:exp:vla}

\begin{figure}[!t]
\centering
\vspace{-0.3cm}
\includegraphics[width=0.95\linewidth]{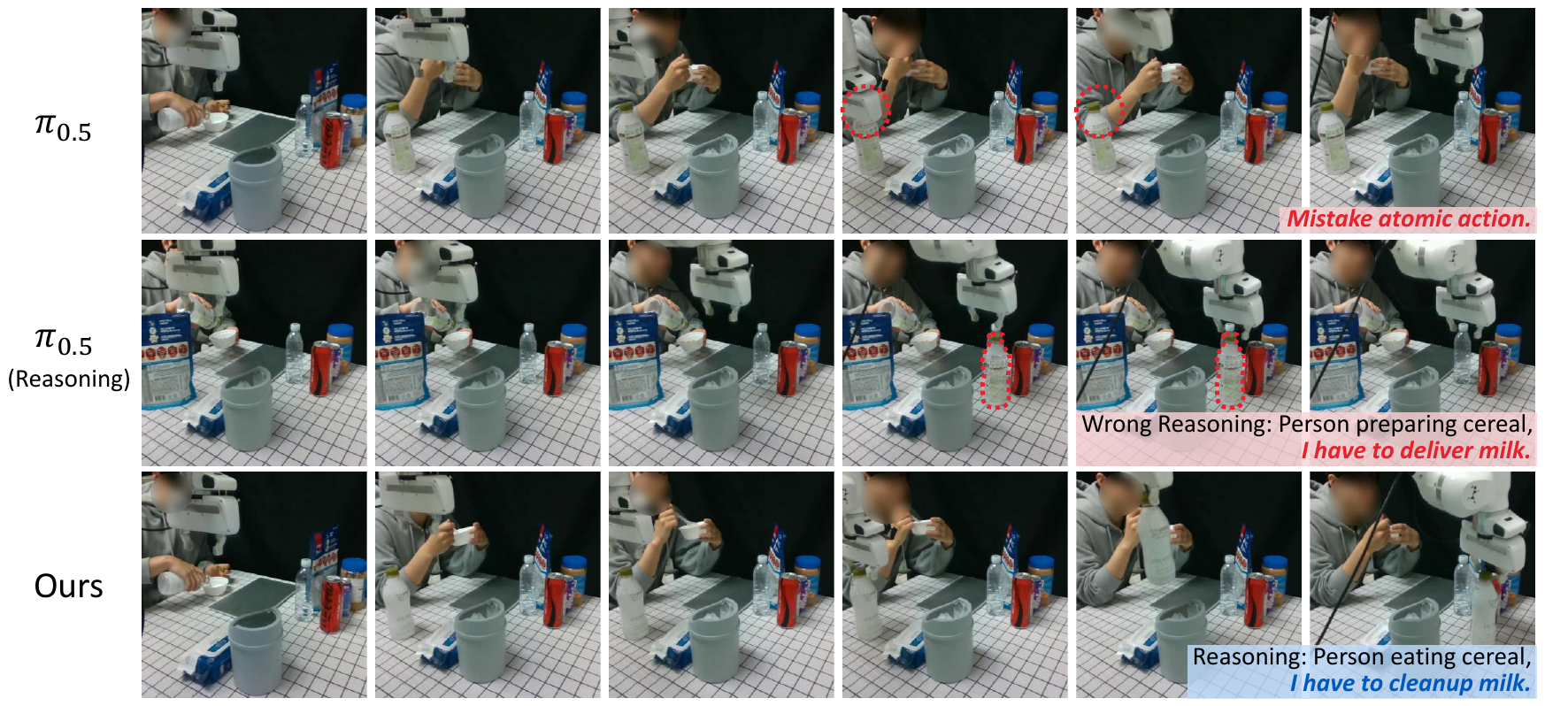}
\caption{
\textbf{Qualitative comparison of closed-loop execution.}
Our System 1--2 framework correctly selects and executes proactive tasks, while $\pi_{0.5}$ and $\pi_{0.5}$ (Reasoning) produce incorrect actions due to joint optimization of reasoning and low-level control under limited data.
}
\label{fig:exp:vla_quali}
\end{figure}

As shown in Table~\ref{tab:exp:vla} and \Cref{fig:exp:vla_quali}, $\pi_{0.5}$, with or without reasoning, suffers from lower success rates on the atomic tasks themselves, since the end-to-end model is forced to jointly learn robot actions and VLM-style reasoning. For example, Milk$\rightarrow$Storage in \Cref{fig:exp:vla_quali} is among the harder atomic skills, as it requires grasping the rounded milk cap; $\pi_{0.5}$ trained with a single prompt across all tasks lacks the capacity to specialize on such individual skills, dragging down per-task success.

This degradation is especially pronounced in the reasoning variant under our low-data regime. Even though we provide $\pi_{0.5}$ (Reasoning) with frame-level supervision specifying which language subtask should be executed, the model still frequently produces incorrect reasoning labels at inference time. For instance, when the correct prediction is Cereal$\rightarrow$Storage but the user happens to be holding the milk, the visible milk often misleads the model into reasoning Milk$\rightarrow$Storage, and such reasoning errors in turn make the generated action even more likely to fail. In contrast, our framework delegates atomic action generation to the strong pretrained $\pi_{0.5}$ backbone and offloads reasoning to Qwen 3.5, achieving higher accuracy on both axes.

\begin{table}[!t]
\centering
\caption{\textbf{Task-wise success rate comparison with monolithic VLA}~\citep{pi05} with and without subtask reasoning. Bold = best per task.}
\vspace{3pt}
\label{tab:exp:vla}
\resizebox{\textwidth}{!}{\begin{tabular}{lcccccc}
\toprule
Method & Cereal$\rightarrow$Person & Milk$\rightarrow$Person & Cereal$\rightarrow$Storage & Milk$\rightarrow$Storage & Bowl$\rightarrow$Storage & Wet
Wipe$\rightarrow$Trash \\
\drule
$\pi_{0.5}$~\citep{pi05} & 7 / 10 & 1 / 10 & 6 / 10 & 0 / 10 & 5 / 10 & 4 / 10 \\
$\pi_{0.5}$ (Reasoning) & 6 / 10 & 0 / 10& 0 / 10 & 0 / 10 & 5 / 10 & 5 / 10 \\
Ours & \textbf{9 / 10} & \textbf{3 / 10} & \textbf{8 / 10} & \textbf{5 / 10} & \textbf{8 / 10} & \textbf{8 / 10} \\
\bottomrule
\end{tabular}
}
\end{table}

\section{Conclusion}
\label{sec:conclusion}
\vspace{-0.2cm}

We introduced Procedural Assistance Memory (PAM), an FSM-based memory that personalizes proactive robot assistance through daily reflection without weight updates. Across five evaluation days, PAM improves proactive prediction and response timing, while its structured memory adapts to changes in the user's routine. Ablations demonstrate the benefits of FSM structure and continued reflection, and the System 1--2 decomposition improves task execution over monolithic VLAs. These results support structured, evolving memory as a practical approach to longitudinal robot personalization. We hope this work encourages further research on proactive robot assistance.

\paragraph{Limitation.}
Two failure modes recur in our analysis. First, FSM Memory occasionally fires when the user briefly handles a learned trigger object without committing to that step, which lowers its \texttt{NO\_ACT} recall (\Cref{fig:exp:vlm}); in such genuinely ambiguous situations the right behavior is often to ask the user rather than to act, but soliciting clarification lies outside our current framework. Second, due to the limited capacity of current open-source VLAs, our evaluation is restricted to seen scenarios rather than novel objects, layouts, or users; recent advances in generalist VLAs~\citep{intelligence2025pi06, intelligence2026pi07, generalist2026gen1, kim2026rldx, liu2026rdt2, wu2026pragmatic, apanasevich2026green} suggest that scaling to broader scenarios is an increasingly feasible next step.

\clearpage
\acknowledgments{This work was supported by Institute for Information \& communications Technology Planning \& Evaluation(IITP) grant funded by the Korea government(MSIT) (RS-2019-II190075, Artificial Intelligence Graduate School Program(KAIST)). This work was supported by the National Research Foundation of Korea(NRF) grant funded by the Korea government(MSIT) (No. RS-2025-00555621). This research was supported by the ``Advanced GPU Utilization Support Program'' funded by the Government of the Republic of Korea (Ministry of Science and ICT).
}

\bibliography{example}

\clearpage

\input{supp_input}

\end{document}

%% file: preamble.tex
\usepackage{kotex}
\usepackage{lipsum}
\usepackage{pifont}\usepackage[normalem]{ulem}
\usepackage{mathtools}

\usepackage{array}
\newcolumntype{P}[1]{>{\centering\arraybackslash}p{#1}}
\newcolumntype{M}[1]{>{\centering\arraybackslash}m{#1}}
\newcolumntype{L}[1]{>{\hspace{0.5em}\raggedright\arraybackslash}m{#1}}
\newcolumntype{R}[1]{>{\raggedleft\arraybackslash}m{#1}}

\usepackage{xspace}

\makeatletter
\DeclareRobustCommand\onedot{\futurelet\@let@token\@onedot}
\def\@onedot{\ifx\@let@token.\else.\null\fi\xspace}

\def\eg{\emph{e.g}\onedot}

\makeatother

\PassOptionsToPackage{colorlinks=true, linkcolor=blue, citecolor=blue, urlcolor=blue, pagebackref=true}{hyperref}
\definecolor[named]{ACMDarkBlue}{cmyk}{1,0.58,0,0.21}
\definecolor{darkgreen}{RGB}{25,200,25}
\hypersetup{colorlinks,
  linkcolor=ACMDarkBlue,  citecolor=ACMDarkBlue,  urlcolor=ACMDarkBlue,   filecolor=ACMDarkBlue
}

\makeatletter
\let\PAMOriginalSortCites\NAT@sort@cites
\def\NAT@sort@cites#1{\PAMOriginalSortCites{#1}\Hy@backout{#1}}
\makeatother

\renewcommand*{\backref}[1]{}

\renewcommand*{\backrefalt}[4]{\ifcase#1\relax\or(Cited on page~#2.)\else(Cited on pages~#2.)\fi}

\usepackage{cleveref}
\Crefformat{figure}{Fig.~#2#1#3}
\Crefformat{equation}{(#2#1#3)}
\Crefformat{table}{Tab.~#2#1#3}

\let\cite\citep

\usepackage{caption}
\DeclareCaptionFormat{tablewithgap}{#1#2\vspace{3pt}#3}

\usepackage{tabularx,booktabs}
\newcolumntype{Y}{>{\centering\arraybackslash}X}
\usepackage{colortbl}
\usepackage{multirow}
\usepackage{soul}
\usepackage{wrapfig}
\newcommand{\drule}{\specialrule{0.2pt}{1pt}{1pt}\specialrule{0.2pt}{0pt}{\belowrulesep}}

\usepackage{footmisc}
\newcommand{\dagfootnote}[1]{\textsuperscript{\dag}\begingroup
    \let\savedfootnoterule\footnoterule
    \renewcommand{\thefootnote}{}\renewcommand{\footnoterule}{\vspace{1ex}\hrule width 0.4\textwidth\vspace{1ex}}\footnotetext{\dag\ #1}\addtocounter{footnote}{-1}\let\footnoterule\savedfootnoterule
  \endgroup
}

\usepackage{algorithm,algpseudocode}
\usepackage[algo2e]{algorithm2e} 
\definecolor{commentcolor}{RGB}{110,154,155}
\definecolor{defcolor}{RGB}{225,81,145}
    
\renewcommand{\arraystretch}{1.1}

\usepackage{titletoc}



\usepackage{enumitem}
\usepackage{flushend}

\usepackage[T1]{fontenc}
\usepackage[utf8]{inputenc}

%% file: supp_input.tex
\appendix

\section*{Appendix}

We first describe the detailed experimental setup, including the evaluation dataset and annotation procedure, the evaluation settings, the training and deployment of VLAs, and the hardware configuration, to ensure reproducibility (\Cref{appn:setup}).
We then report the per-task results of VLM proactiveness, complementing the aggregate metrics reported in the main paper (\Cref{appn:per_task}).
Next, we present a further analysis of the memory update process, examining the effectiveness of memory evolution, the reasoning traces behind each update, and a comparison across different memory structures (\Cref{appn:memory}).
We also present an evaluation on an additional routine and results for the reverse B$\rightarrow$A routine-change experiment (\Cref{appn:diverse_scenario}). Finally, we include the full prompts used for the task predictor (\Cref{appn:infer_prompt}) and the reflector (\Cref{appn:reflect_prompt}) across our List Memory and FSM Memory variants. Below is the table of contents for the appendix.

\startcontents[appendices]
\printcontents[appendices]{l}{1}{\setcounter{tocdepth}{2}}

\section{Detailed Experimental Setup}
\label{appn:setup}

\subsection{Evaluation Dataset and Annotation}
\label{app:eval-dataset}

\begin{figure}[!t]
    \centering
    \includegraphics[width=\linewidth]{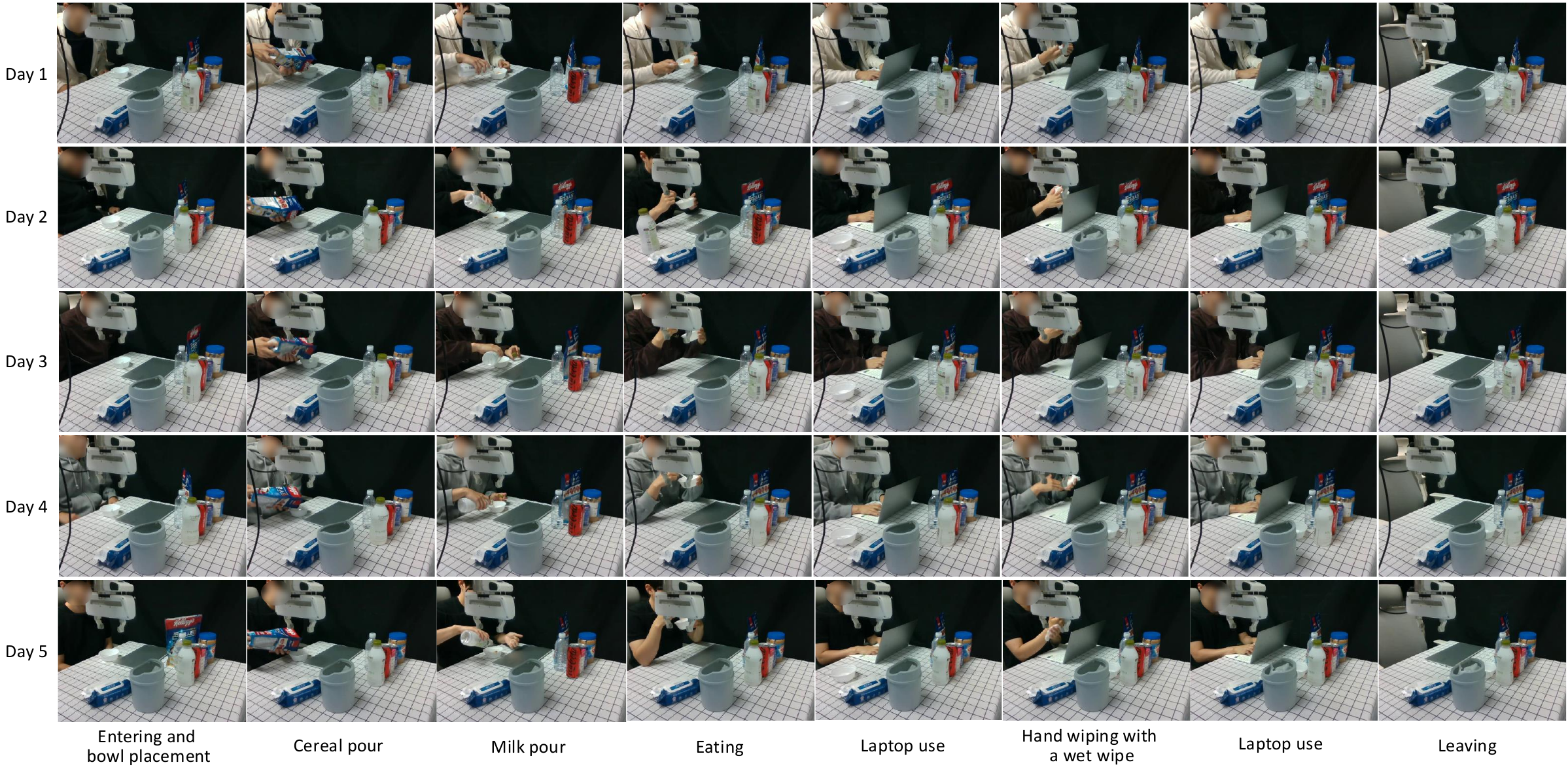}
    \caption{Snapshots of the five evaluation days. Each day captures the same canonical desk routine performed by a single user: bringing out a bowl, pouring cereal and milk, eating, using the laptop with the bowl set aside, wiping the mouth with a wet wipe, and leaving the desk.}
    \label{fig:5days}
\end{figure}

To examine how proactive-task prediction performance evolves with repeated exposure to a user's routine, we record five days of video of a single user performing the same canonical desk routine (\Cref{fig:5days}). Each day follows the same high-level sequence: the user arrives at the desk with a bowl, pours cereal and then milk into the bowl, eats the meal, sets the empty bowl aside while switching to laptop work, wipes their mouth with a wet wipe and leaves it on the desk, and finally leaves the desk. The timing of individual actions and the visible desk state vary naturally across days. These recordings allow us to evaluate whether repeated exposure to the same routine improves task prediction and enables earlier assistance.

\paragraph{Per-frame annotation.}
We treat proactive-task prediction as a per-frame classification problem and therefore label every evaluation frame. Each day's recording lasts roughly $3.5$ to $4.5$ minutes; we uniformly sample at $1$\,fps and obtain $197$ to $270$ frames per day after excluding deleted frames, for a total of $1{,}146$ annotated frames across the five days (\Cref{tab:eval-stats}). The authors manually assign one label per sampled frame from the following seven categories: the six proactive tasks T1--T6 (delivering cereal, delivering milk, putting away cereal, putting away milk, putting away the empty bowl, throwing away the used wet wipe), and \texttt{NO\_ACT} for frames where no intervention is appropriate. A frame receives a task label when, given the current scene and the user's ongoing behavior, that task is the most natural proactive intervention; all other frames are labeled \texttt{NO\_ACT}. The resulting labels provide the ground truth for the proactive task prediction experiments on this routine, and we will release the annotation tool together with the dataset.

\begin{table}[!t]
\centering
\caption{Per-day annotation statistics on the five evaluation days. Frames are extracted at $1$\,fps; deleted frames are excluded from the totals. T1: Cereal$\rightarrow$Person, T2: Milk$\rightarrow$Person, T3: Cereal$\rightarrow$Storage, T4: Milk$\rightarrow$Storage, T5: Bowl$\rightarrow$Storage, T6: Wet wipe$\rightarrow$Trash. \texttt{NO\_ACT} marks frames where no intervention is appropriate.}
\label{tab:eval-stats}
\vspace{0.5cm}
\begin{tabular}{c c c c c c c c c}
\toprule
Day & Frames & \texttt{NO\_ACT} & T1 & T2 & T3 & T4 & T5 & T6 \\
\midrule
1 & 270 & 133 & 12 & 21 &  5 & 34 & 35 & 30 \\
2 & 248 & 143 & 11 & 27 &  6 & 18 & 24 & 19 \\
3 & 224 & 154 &  9 & 23 &  6 &  8 & 14 & 10 \\
4 & 207 & 125 &  8 & 23 &  5 & 16 & 16 & 14 \\
5 & 197 & 118 &  7 & 19 &  4 & 13 & 21 & 15 \\
\midrule
Total & 1{,}146 & 673 & 47 & 113 & 26 & 89 & 110 & 88 \\
\bottomrule
\end{tabular}
\end{table}

\subsection{Evaluation Settings}
\label{app:eval-settings}

Both the task predictor and the reflector use the same Qwen 3.5--27B backbone~\cite{qwen3.5} served via vLLM~\cite{vllm}, but they differ in the input format, the use of thinking, and the call cadence.

\paragraph{Task predictor.}
At each tick the task predictor receives the most recent $W{=}5$ frames sampled at stride $2$ from the 1\,fps stream, i.e., the frames at $t,\,t-2,\,t-4,\,t-6,\,t-8$ seconds, together with the current memory and the wall-clock time of the final frame.
Each frame is the horizontal concatenation of the two back-facing cameras (left and right; \Cref{fig:hardware}); the wrist camera is not exposed to the VLM stack. 
Thinking mode is \emph{disabled}, so the model directly outputs a JSON object containing \texttt{observation}, a brief \texttt{reasoning} field, and \texttt{task\_index}, without a separate thinking trace. The model generates outputs with a temperature of $0.3$. Each call takes approximately $1$ second on average across all five days. The full system prompts driving this stage, one per memory variant (no memory, list, FSM), are provided in \Cref{appn:infer_prompt}.

\paragraph{Reflector.}
The reflector is invoked once at the end of each day. Its visual input is the day's recording uniformly downsampled to $200$ frames, again using the two-view (left and right) back-facing concatenation without the wrist camera; the prior memory, an action-history summary of the day, and (when provided) user feedback are appended as text. Thinking mode is \emph{enabled}, so the model first produces an explicit reasoning trace before emitting the updated memory JSON (FSM or list schema). The model generates outputs with a temperature of $0.3$. Each call takes approximately $1$ minute on average across all five days. The full system prompts for building memory from scratch on the first day and for end-of-day updates on subsequent days, separately for the list and FSM variants, are provided in \Cref{appn:reflect_prompt}.

\subsection{Training and Deployment for VLAs}
\label{app:vla-training-data}

\begin{figure}[!t]
    \centering
    \includegraphics[width=\linewidth]{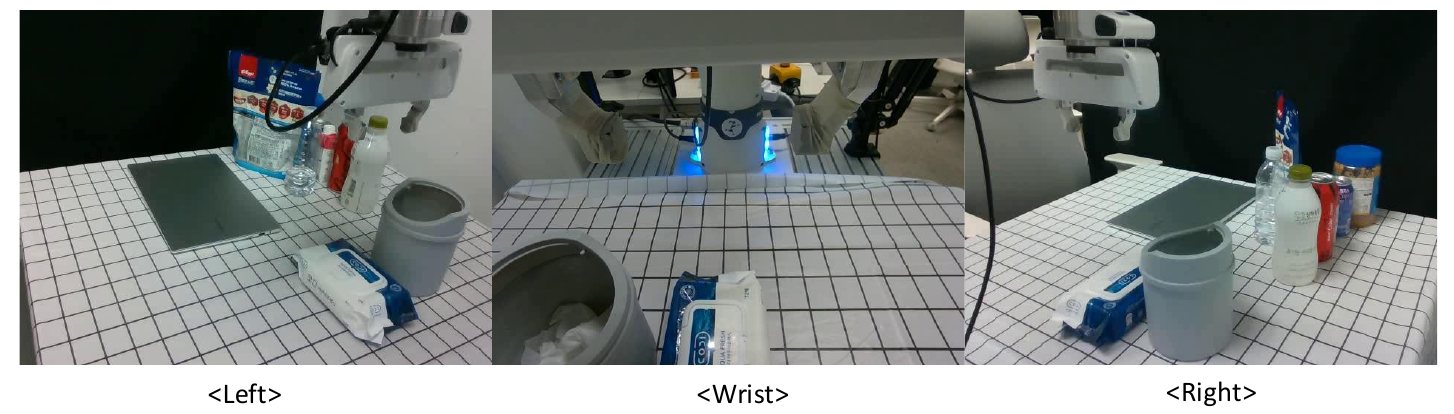}
    \caption{Camera setup. Two back-facing cameras (left and right) observe the user and the desk from behind the Franka arm, while a wrist camera mounted on the end-effector provides close-up grasp context. The VLM-based task predictor and reflector consume only the left and right back views; the VLAs additionally receive the wrist view.}
    \label{fig:hardware}
\end{figure}

All three VLA variants in our study are fine-tuned from the same $\pi_{0.5}$ backbone but differ in what they have to learn. In the cereal routine, we consider six proactive tasks: (T1)~\emph{Pick up the cereal and place it next to the person.}, (T2)~\emph{Pick up the milk and place it next to the person.}, (T3)~\emph{Pick up the cereal and place it in the designated storage area.}, (T4)~\emph{Pick up the milk and place it in the designated storage area.}, (T5)~\emph{Pick up the bowl and place it in the designated storage area.}, and (T6)~\emph{Pick up the used wet wipe and place it in the trash can.}\ All teleoperation demonstrations are recorded on the Franka arm at 20\,fps with three RGB streams (two back-facing cameras and one wrist camera; \Cref{fig:hardware}). Unlike the VLM stack, all three VLA variants receive all three views as input, including the wrist camera that provides close-up grasp context.

\paragraph{$\boldsymbol{\pi_{0.5}}$ (executor).}
Our executor executes one task at a time, as instructed by the task predictor. We therefore collect $10$ teleoperation episodes per task and fine-tune $\pi_{0.5}$ on this set, conditioning each episode on its corresponding instruction taken verbatim from T1--T6. At deployment, the executor is conditioned on the task instruction provided by the task predictor.
\paragraph{$\boldsymbol{\pi_{0.5}}$ (monolithic).}

The monolithic baseline has no external task predictor and must therefore learn both \emph{when} to act and \emph{what} to do inside a single model. To match this requirement, we collect 10 end-to-end teleoperation episodes of the full scenario, in which the user arrives at the desk and proceeds through the routine, and the robot, controlled by an expert teleoperator, performs every proactive task at the appropriate moment without any external instruction. We fine-tune $\pi_{0.5}$ on these full-scenario demonstrations, conditioned on a single fixed high-level instruction, \emph{``Actively and proactively help humans before they even ask.''}, that does not specify which subtask should be issued at any particular moment.
\paragraph{$\boldsymbol{\pi_{0.5}}$ (monolithic + reasoning).}

The reasoning baseline uses the same 10 full-scenario episodes as the monolithic variant, but additionally relies on per-frame subtask supervision so that $\pi_{0.5}$'s native reasoning channel can be trained explicitly.
Each frame is annotated with a subtask string, which is either one of T1--T6 or \emph{``No action is required right now.''}\ when no intervention is appropriate.
During training, the prompt fed to $\pi_{0.5}$ has the form \texttt{Task:\{HL\}.\ Subtask:\{sub\}.\ State:\ldots;\textbackslash nAction:\,} with the same fixed high-level instruction as the monolithic variant, and the total loss combines the standard action-prediction loss with a cross-entropy term on the predicted subtask text (with weight $1.0$); at inference time the model first emits the subtask and then conditions its action on its own subtask prediction.

\paragraph{Training and deployment settings.}
All three VLA variants share the same training and deployment recipe.
We fine-tune $\pi_{0.5}$ for $50{,}000$ optimizer steps with AdamW~\cite{loshchilov2017decoupled} (betas $(0.9, 0.95)$, $\epsilon{=}10^{-8}$, weight decay $0.01$, gradient-norm clipping at $1.0$). 
The learning rate is linearly warmed up from $0$ to $1\times 10^{-4}$ over the first $1{,}000$ steps and then cosine-decayed over the remaining training steps to a minimum of $2.5\times 10^{-6}$. 
Training uses NVIDIA B200 GPUs with an effective batch size of $64$ and no gradient accumulation. 
The action head predicts chunks of $50$ steps, and at deployment we execute the full predicted chunk before requesting a new one, with $10$ denoising steps per chunk.

\subsection{Hardware Setup}
\label{app:hardware}

\paragraph{Robot.}
We use a Franka Research 3 (FR3) 7-DoF manipulator equipped with the built-in parallel-jaw gripper. Low-level control is provided by the Deoxys real-time controller, which exposes operational-space control of the end-effector. The control loop runs at $20$\,Hz during teleoperation and deployment.

\paragraph{Cameras.}
Three Intel RealSense cameras are used for both data collection and deployment, all streaming RGB at 30\,fps and time-aligned to the $20$\,Hz control loop.
Two cameras (\texttt{left\_back} and \texttt{right\_back}) are statically mounted behind the robot at table height, framing the desk and the seated user from two angles, while the third camera is rigidly attached to the wrist and looks down toward the gripper. The two back-facing streams are horizontally concatenated to form the input given to the VLM stack, whereas the VLA variants additionally consume the wrist stream as a third image channel.

\paragraph{Teleoperation interface.}
Teleoperation demonstrations are collected with TactAR~\cite{xue2025reactive}, a Meta Quest 3-based VR teleoperation interface. The operator wears the Quest headset and uses the right hand-held controller; controller pose updates are streamed over HTTP to a FastAPI server, converted from the Unity coordinate frame to the robot base frame, and applied as absolute target end-effector poses with $1{:}1$ translation scaling. 
Gripper open/close is bound to a controller trigger. 
The same setup is used for all three VLA training datasets (executor, monolithic, and monolithic + reasoning).

\section{Per-task Results of VLM Proactiveness}
\label{appn:per_task}

\begin{figure}[!t]
\centering
\includegraphics[width=\linewidth]{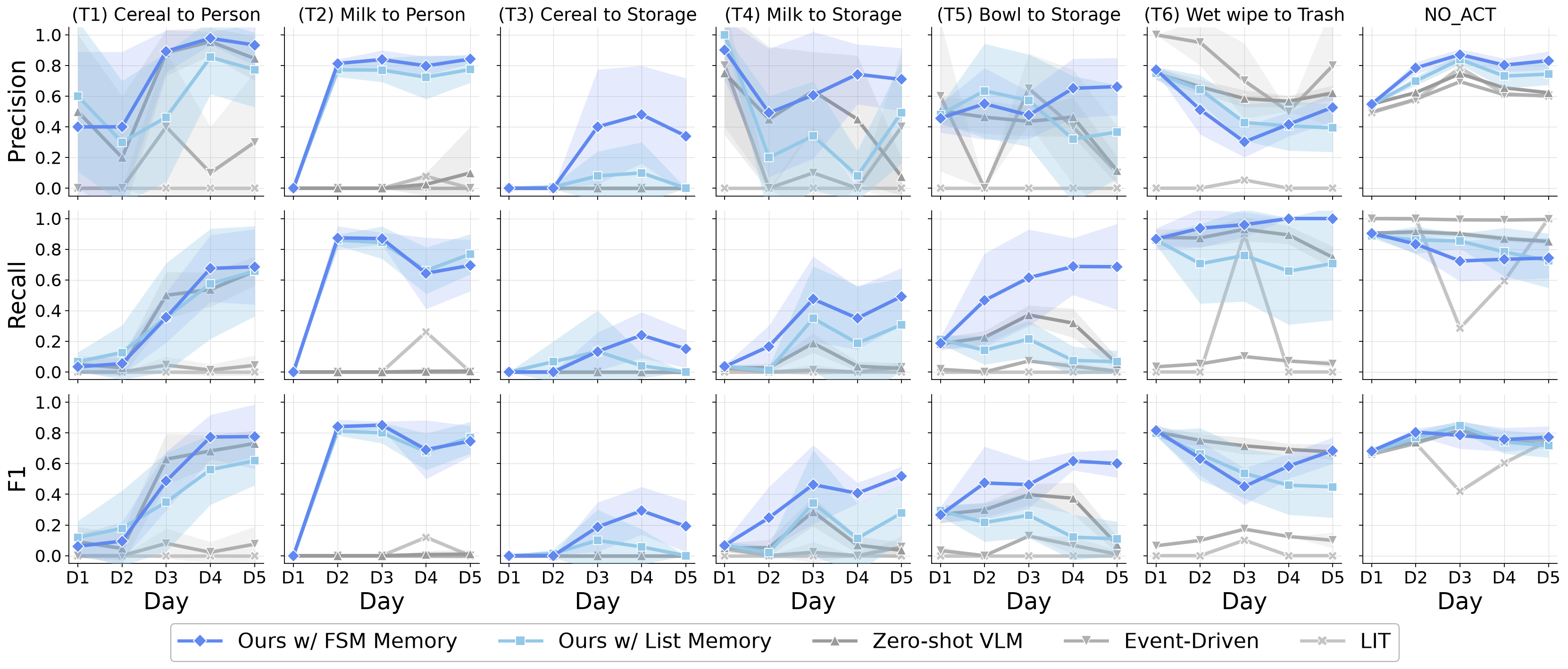}
\caption{Per-task precision, recall, and F1. Memory-based approaches outperform the memory-free baselines on Milk$\rightarrow$Person and Cereal$\rightarrow$Storage while maintaining comparable \texttt{NO\_ACT} F1. FSM Memory provides further gains over List Memory on Cereal$\rightarrow$Storage and Milk$\rightarrow$Storage.}
\label{fig:appn:per_task}
\end{figure}

\paragraph{Memory-augmented variants improve on average across days.}
Both memory variants improve their average F1 across T1--T6 from Day 1 to Day 5, although gains vary by task (\Cref{fig:appn:per_task}). Memory-free baselines (LIT~\cite{litsouschef}, Event-driven~\cite{eventdriven}, and Zero-shot VLM~\cite{qwen3.5}) achieve more limited performance overall.

\paragraph{Memory helps disambiguate assistance.}
Memory is particularly helpful on T2 (Milk$\rightarrow$Person) and T3 (Cereal$\rightarrow$Storage), where assistance depends on the user's habitual pairing or the routine's progression. For example, observing cereal being poured does not by itself establish that this user needs milk next. The gap is smaller on T1 and T6, whose immediate cues are more informative.

\paragraph{FSM structure helps most on storage actions.}
FSM Memory provides substantial gains over List Memory on storage tasks T3--T5. Organizing rules by activity phase helps distinguish similar cues and select the appropriate moment to put items away. This context is less explicit in List Memory, where routine information must be encoded in individual trigger strings. FSM Memory has lower \texttt{NO\_ACT} recall but higher \texttt{NO\_ACT} precision than List Memory, resulting in comparable F1 and reflecting the trade-off discussed in Section~4.2.

\paragraph{Numerical per-task results.}
Tables~\ref{tab:appn:per_task_precision}--\ref{tab:appn:per_task_f1} report per-task results for all five methods on D1--D5 as mean $\pm$ standard deviation, rounded to two decimal places.
\clearpage
\begingroup
\makeatletter
\setlength{\@fptop}{0pt plus 1fil}
\setlength{\@fpbot}{0pt plus 1fil}
\makeatother
\begin{table}[p]
\centering
\caption{Per-task Precision on D1--D5 (mean $\pm$ standard deviation). T1: Cereal$\rightarrow$Person; T2: Milk$\rightarrow$Person; T3: Cereal$\rightarrow$Storage; T4: Milk$\rightarrow$Storage; T5: Bowl$\rightarrow$Storage; T6: Wet wipe$\rightarrow$Trash. \texttt{NO\_ACT} denotes no intervention.}
\label{tab:appn:per_task_precision}
\footnotesize
\setlength{\tabcolsep}{3pt}
\renewcommand{\arraystretch}{1.35}
\scalebox{0.98}{\begin{tabular}{@{}llccccc@{}}
\toprule
Task & Method & D1 & D2 & D3 & D4 & D5 \\
\midrule
T1 & LIT & $0.00\!\pm\!0.00$ & $0.00\!\pm\!0.00$ & $0.00\!\pm\!0.00$ & $0.00\!\pm\!0.00$ & $0.00\!\pm\!0.00$ \\
 & Event-Driven & $0.00\!\pm\!0.00$ & $0.00\!\pm\!0.00$ & $0.40\!\pm\!0.49$ & $0.10\!\pm\!0.30$ & $0.30\!\pm\!0.46$ \\
 & Zero-shot VLM & $0.50\!\pm\!0.50$ & $0.20\!\pm\!0.40$ & $0.88\!\pm\!0.15$ & $0.96\!\pm\!0.09$ & $0.85\!\pm\!0.14$ \\
 & Ours (List Memory) & $0.60\!\pm\!0.49$ & $0.30\!\pm\!0.40$ & $0.46\!\pm\!0.42$ & $0.85\!\pm\!0.24$ & $0.77\!\pm\!0.24$ \\
 & Ours (FSM Memory) & $0.40\!\pm\!0.49$ & $0.40\!\pm\!0.49$ & $0.89\!\pm\!0.14$ & $0.98\!\pm\!0.04$ & $0.93\!\pm\!0.13$ \\
\midrule
T2 & LIT & $0.00\!\pm\!0.00$ & $0.00\!\pm\!0.00$ & $0.00\!\pm\!0.00$ & $0.08\!\pm\!0.00$ & $0.00\!\pm\!0.00$ \\
 & Event-Driven & $0.00\!\pm\!0.00$ & $0.00\!\pm\!0.00$ & $0.00\!\pm\!0.00$ & $0.00\!\pm\!0.00$ & $0.00\!\pm\!0.00$ \\
 & Zero-shot VLM & $0.00\!\pm\!0.00$ & $0.00\!\pm\!0.00$ & $0.00\!\pm\!0.00$ & $0.03\!\pm\!0.08$ & $0.10\!\pm\!0.30$ \\
 & Ours (List Memory) & $0.00\!\pm\!0.00$ & $0.77\!\pm\!0.05$ & $0.77\!\pm\!0.08$ & $0.72\!\pm\!0.14$ & $0.77\!\pm\!0.09$ \\
 & Ours (FSM Memory) & $0.00\!\pm\!0.00$ & $0.81\!\pm\!0.03$ & $0.84\!\pm\!0.06$ & $0.80\!\pm\!0.06$ & $0.84\!\pm\!0.03$ \\
\midrule
T3 & LIT & $0.00\!\pm\!0.00$ & $0.00\!\pm\!0.00$ & $0.00\!\pm\!0.00$ & $0.00\!\pm\!0.00$ & $0.00\!\pm\!0.00$ \\
 & Event-Driven & $0.00\!\pm\!0.00$ & $0.00\!\pm\!0.00$ & $0.00\!\pm\!0.00$ & $0.00\!\pm\!0.00$ & $0.00\!\pm\!0.00$ \\
 & Zero-shot VLM & $0.00\!\pm\!0.00$ & $0.00\!\pm\!0.00$ & $0.00\!\pm\!0.00$ & $0.00\!\pm\!0.00$ & $0.00\!\pm\!0.00$ \\
 & Ours (List Memory) & $0.00\!\pm\!0.00$ & $0.01\!\pm\!0.02$ & $0.08\!\pm\!0.16$ & $0.10\!\pm\!0.20$ & $0.00\!\pm\!0.00$ \\
 & Ours (FSM Memory) & $0.00\!\pm\!0.00$ & $0.00\!\pm\!0.00$ & $0.40\!\pm\!0.37$ & $0.48\!\pm\!0.32$ & $0.34\!\pm\!0.38$ \\
\midrule
T4 & LIT & $0.00\!\pm\!0.00$ & $0.00\!\pm\!0.00$ & $0.00\!\pm\!0.00$ & $0.00\!\pm\!0.00$ & $0.00\!\pm\!0.00$ \\
 & Event-Driven & $0.80\!\pm\!0.40$ & $0.00\!\pm\!0.00$ & $0.10\!\pm\!0.30$ & $0.00\!\pm\!0.00$ & $0.40\!\pm\!0.49$ \\
 & Zero-shot VLM & $0.75\!\pm\!0.40$ & $0.45\!\pm\!0.47$ & $0.63\!\pm\!0.26$ & $0.45\!\pm\!0.42$ & $0.08\!\pm\!0.12$ \\
 & Ours (List Memory) & $1.00\!\pm\!0.00$ & $0.20\!\pm\!0.40$ & $0.34\!\pm\!0.36$ & $0.08\!\pm\!0.16$ & $0.49\!\pm\!0.33$ \\
 & Ours (FSM Memory) & $0.90\!\pm\!0.20$ & $0.49\!\pm\!0.42$ & $0.61\!\pm\!0.41$ & $0.74\!\pm\!0.20$ & $0.71\!\pm\!0.21$ \\
\midrule
T5 & LIT & $0.00\!\pm\!0.00$ & $0.00\!\pm\!0.00$ & $0.00\!\pm\!0.00$ & $0.00\!\pm\!0.00$ & $0.00\!\pm\!0.00$ \\
 & Event-Driven & $0.60\!\pm\!0.49$ & $0.00\!\pm\!0.00$ & $0.65\!\pm\!0.23$ & $0.40\!\pm\!0.37$ & $0.10\!\pm\!0.30$ \\
 & Zero-shot VLM & $0.51\!\pm\!0.09$ & $0.46\!\pm\!0.11$ & $0.44\!\pm\!0.10$ & $0.46\!\pm\!0.13$ & $0.11\!\pm\!0.09$ \\
 & Ours (List Memory) & $0.48\!\pm\!0.06$ & $0.63\!\pm\!0.31$ & $0.57\!\pm\!0.30$ & $0.32\!\pm\!0.41$ & $0.37\!\pm\!0.31$ \\
 & Ours (FSM Memory) & $0.46\!\pm\!0.09$ & $0.55\!\pm\!0.23$ & $0.48\!\pm\!0.16$ & $0.65\!\pm\!0.19$ & $0.66\!\pm\!0.19$ \\
\midrule
T6 & LIT & $0.00\!\pm\!0.00$ & $0.00\!\pm\!0.00$ & $0.05\!\pm\!0.00$ & $0.00\!\pm\!0.00$ & $0.00\!\pm\!0.00$ \\
 & Event-Driven & $1.00\!\pm\!0.00$ & $0.95\!\pm\!0.15$ & $0.70\!\pm\!0.24$ & $0.50\!\pm\!0.00$ & $0.80\!\pm\!0.40$ \\
 & Zero-shot VLM & $0.75\!\pm\!0.04$ & $0.66\!\pm\!0.05$ & $0.58\!\pm\!0.05$ & $0.57\!\pm\!0.04$ & $0.62\!\pm\!0.05$ \\
 & Ours (List Memory) & $0.75\!\pm\!0.04$ & $0.64\!\pm\!0.09$ & $0.43\!\pm\!0.12$ & $0.41\!\pm\!0.16$ & $0.39\!\pm\!0.16$ \\
 & Ours (FSM Memory) & $0.77\!\pm\!0.02$ & $0.51\!\pm\!0.16$ & $0.30\!\pm\!0.10$ & $0.42\!\pm\!0.08$ & $0.53\!\pm\!0.09$ \\
\midrule
\texttt{NO\_ACT} & LIT & $0.49\!\pm\!0.00$ & $0.58\!\pm\!0.00$ & $0.79\!\pm\!0.00$ & $0.62\!\pm\!0.00$ & $0.60\!\pm\!0.00$ \\
 & Event-Driven & $0.50\!\pm\!0.00$ & $0.58\!\pm\!0.00$ & $0.69\!\pm\!0.00$ & $0.61\!\pm\!0.00$ & $0.60\!\pm\!0.00$ \\
 & Zero-shot VLM & $0.55\!\pm\!0.01$ & $0.62\!\pm\!0.01$ & $0.75\!\pm\!0.01$ & $0.65\!\pm\!0.01$ & $0.62\!\pm\!0.01$ \\
 & Ours (List Memory) & $0.54\!\pm\!0.00$ & $0.70\!\pm\!0.03$ & $0.84\!\pm\!0.03$ & $0.73\!\pm\!0.08$ & $0.74\!\pm\!0.07$ \\
 & Ours (FSM Memory) & $0.55\!\pm\!0.01$ & $0.78\!\pm\!0.04$ & $0.87\!\pm\!0.04$ & $0.80\!\pm\!0.04$ & $0.83\!\pm\!0.06$ \\
\bottomrule
\end{tabular}}
\end{table}
\clearpage
\begin{table}[p]
\centering
\caption{Per-task Recall on D1--D5 (mean $\pm$ standard deviation). T1: Cereal$\rightarrow$Person; T2: Milk$\rightarrow$Person; T3: Cereal$\rightarrow$Storage; T4: Milk$\rightarrow$Storage; T5: Bowl$\rightarrow$Storage; T6: Wet wipe$\rightarrow$Trash. \texttt{NO\_ACT} denotes no intervention.}
\label{tab:appn:per_task_recall}
\footnotesize
\setlength{\tabcolsep}{3pt}
\renewcommand{\arraystretch}{1.35}
\scalebox{0.98}{\begin{tabular}{@{}llccccc@{}}
\toprule
Task & Method & D1 & D2 & D3 & D4 & D5 \\
\midrule
T1 & LIT & $0.00\!\pm\!0.00$ & $0.00\!\pm\!0.00$ & $0.00\!\pm\!0.00$ & $0.00\!\pm\!0.00$ & $0.00\!\pm\!0.00$ \\
 & Event-Driven & $0.00\!\pm\!0.00$ & $0.00\!\pm\!0.00$ & $0.04\!\pm\!0.05$ & $0.01\!\pm\!0.04$ & $0.04\!\pm\!0.07$ \\
 & Zero-shot VLM & $0.05\!\pm\!0.06$ & $0.03\!\pm\!0.06$ & $0.50\!\pm\!0.15$ & $0.54\!\pm\!0.11$ & $0.66\!\pm\!0.09$ \\
 & Ours (List Memory) & $0.07\!\pm\!0.06$ & $0.13\!\pm\!0.18$ & $0.36\!\pm\!0.35$ & $0.58\!\pm\!0.36$ & $0.66\!\pm\!0.29$ \\
 & Ours (FSM Memory) & $0.03\!\pm\!0.04$ & $0.05\!\pm\!0.07$ & $0.36\!\pm\!0.16$ & $0.68\!\pm\!0.22$ & $0.69\!\pm\!0.25$ \\
\midrule
T2 & LIT & $0.00\!\pm\!0.00$ & $0.00\!\pm\!0.00$ & $0.00\!\pm\!0.00$ & $0.26\!\pm\!0.00$ & $0.00\!\pm\!0.00$ \\
 & Event-Driven & $0.00\!\pm\!0.00$ & $0.00\!\pm\!0.00$ & $0.00\!\pm\!0.00$ & $0.00\!\pm\!0.00$ & $0.00\!\pm\!0.00$ \\
 & Zero-shot VLM & $0.00\!\pm\!0.00$ & $0.00\!\pm\!0.00$ & $0.00\!\pm\!0.00$ & $0.00\!\pm\!0.01$ & $0.01\!\pm\!0.02$ \\
 & Ours (List Memory) & $0.00\!\pm\!0.00$ & $0.86\!\pm\!0.04$ & $0.84\!\pm\!0.11$ & $0.66\!\pm\!0.15$ & $0.77\!\pm\!0.13$ \\
 & Ours (FSM Memory) & $0.00\!\pm\!0.00$ & $0.87\!\pm\!0.08$ & $0.87\!\pm\!0.04$ & $0.64\!\pm\!0.23$ & $0.69\!\pm\!0.17$ \\
\midrule
T3 & LIT & $0.00\!\pm\!0.00$ & $0.00\!\pm\!0.00$ & $0.00\!\pm\!0.00$ & $0.00\!\pm\!0.00$ & $0.00\!\pm\!0.00$ \\
 & Event-Driven & $0.00\!\pm\!0.00$ & $0.00\!\pm\!0.00$ & $0.00\!\pm\!0.00$ & $0.00\!\pm\!0.00$ & $0.00\!\pm\!0.00$ \\
 & Zero-shot VLM & $0.00\!\pm\!0.00$ & $0.00\!\pm\!0.00$ & $0.00\!\pm\!0.00$ & $0.00\!\pm\!0.00$ & $0.00\!\pm\!0.00$ \\
 & Ours (List Memory) & $0.00\!\pm\!0.00$ & $0.07\!\pm\!0.13$ & $0.13\!\pm\!0.27$ & $0.04\!\pm\!0.08$ & $0.00\!\pm\!0.00$ \\
 & Ours (FSM Memory) & $0.00\!\pm\!0.00$ & $0.00\!\pm\!0.00$ & $0.13\!\pm\!0.12$ & $0.24\!\pm\!0.15$ & $0.15\!\pm\!0.12$ \\
\midrule
T4 & LIT & $0.00\!\pm\!0.00$ & $0.00\!\pm\!0.00$ & $0.00\!\pm\!0.00$ & $0.00\!\pm\!0.00$ & $0.00\!\pm\!0.00$ \\
 & Event-Driven & $0.02\!\pm\!0.01$ & $0.00\!\pm\!0.00$ & $0.01\!\pm\!0.04$ & $0.00\!\pm\!0.00$ & $0.03\!\pm\!0.04$ \\
 & Zero-shot VLM & $0.03\!\pm\!0.02$ & $0.03\!\pm\!0.03$ & $0.19\!\pm\!0.06$ & $0.04\!\pm\!0.03$ & $0.02\!\pm\!0.04$ \\
 & Ours (List Memory) & $0.04\!\pm\!0.01$ & $0.01\!\pm\!0.02$ & $0.35\!\pm\!0.34$ & $0.19\!\pm\!0.38$ & $0.31\!\pm\!0.30$ \\
 & Ours (FSM Memory) & $0.04\!\pm\!0.01$ & $0.17\!\pm\!0.14$ & $0.48\!\pm\!0.28$ & $0.35\!\pm\!0.20$ & $0.49\!\pm\!0.19$ \\
\midrule
T5 & LIT & $0.00\!\pm\!0.00$ & $0.00\!\pm\!0.00$ & $0.00\!\pm\!0.00$ & $0.00\!\pm\!0.00$ & $0.00\!\pm\!0.00$ \\
 & Event-Driven & $0.02\!\pm\!0.01$ & $0.00\!\pm\!0.00$ & $0.07\!\pm\!0.00$ & $0.04\!\pm\!0.03$ & $0.00\!\pm\!0.01$ \\
 & Zero-shot VLM & $0.18\!\pm\!0.04$ & $0.23\!\pm\!0.04$ & $0.37\!\pm\!0.06$ & $0.32\!\pm\!0.09$ & $0.05\!\pm\!0.04$ \\
 & Ours (List Memory) & $0.21\!\pm\!0.03$ & $0.14\!\pm\!0.09$ & $0.21\!\pm\!0.12$ & $0.08\!\pm\!0.09$ & $0.07\!\pm\!0.07$ \\
 & Ours (FSM Memory) & $0.19\!\pm\!0.04$ & $0.47\!\pm\!0.31$ & $0.61\!\pm\!0.31$ & $0.69\!\pm\!0.19$ & $0.69\!\pm\!0.28$ \\
\midrule
T6 & LIT & $0.00\!\pm\!0.00$ & $0.00\!\pm\!0.00$ & $0.90\!\pm\!0.00$ & $0.00\!\pm\!0.00$ & $0.00\!\pm\!0.00$ \\
 & Event-Driven & $0.03\!\pm\!0.00$ & $0.05\!\pm\!0.00$ & $0.10\!\pm\!0.00$ & $0.07\!\pm\!0.00$ & $0.05\!\pm\!0.03$ \\
 & Zero-shot VLM & $0.88\!\pm\!0.05$ & $0.87\!\pm\!0.06$ & $0.93\!\pm\!0.08$ & $0.89\!\pm\!0.06$ & $0.75\!\pm\!0.07$ \\
 & Ours (List Memory) & $0.86\!\pm\!0.03$ & $0.71\!\pm\!0.26$ & $0.76\!\pm\!0.30$ & $0.66\!\pm\!0.35$ & $0.71\!\pm\!0.37$ \\
 & Ours (FSM Memory) & $0.87\!\pm\!0.07$ & $0.94\!\pm\!0.13$ & $0.96\!\pm\!0.08$ & $1.00\!\pm\!0.00$ & $1.00\!\pm\!0.00$ \\
\midrule
\texttt{NO\_ACT} & LIT & $1.00\!\pm\!0.00$ & $1.00\!\pm\!0.00$ & $0.29\!\pm\!0.00$ & $0.59\!\pm\!0.00$ & $1.00\!\pm\!0.00$ \\
 & Event-Driven & $1.00\!\pm\!0.00$ & $1.00\!\pm\!0.00$ & $0.99\!\pm\!0.00$ & $0.99\!\pm\!0.00$ & $0.99\!\pm\!0.00$ \\
 & Zero-shot VLM & $0.90\!\pm\!0.01$ & $0.91\!\pm\!0.02$ & $0.90\!\pm\!0.02$ & $0.87\!\pm\!0.02$ & $0.85\!\pm\!0.02$ \\
 & Ours (List Memory) & $0.89\!\pm\!0.01$ & $0.86\!\pm\!0.09$ & $0.85\!\pm\!0.05$ & $0.78\!\pm\!0.16$ & $0.73\!\pm\!0.18$ \\
 & Ours (FSM Memory) & $0.90\!\pm\!0.02$ & $0.83\!\pm\!0.06$ & $0.72\!\pm\!0.13$ & $0.73\!\pm\!0.13$ & $0.74\!\pm\!0.13$ \\
\bottomrule
\end{tabular}}
\end{table}
\clearpage
\begin{table}[p]
\centering
\caption{Per-task F1 on D1--D5 (mean $\pm$ standard deviation). T1: Cereal$\rightarrow$Person; T2: Milk$\rightarrow$Person; T3: Cereal$\rightarrow$Storage; T4: Milk$\rightarrow$Storage; T5: Bowl$\rightarrow$Storage; T6: Wet wipe$\rightarrow$Trash. \texttt{NO\_ACT} denotes no intervention.}
\label{tab:appn:per_task_f1}
\footnotesize
\setlength{\tabcolsep}{3pt}
\renewcommand{\arraystretch}{1.35}
\scalebox{0.98}{\begin{tabular}{@{}llccccc@{}}
\toprule
Task & Method & D1 & D2 & D3 & D4 & D5 \\
\midrule
T1 & LIT & $0.00\!\pm\!0.00$ & $0.00\!\pm\!0.00$ & $0.00\!\pm\!0.00$ & $0.00\!\pm\!0.00$ & $0.00\!\pm\!0.00$ \\
 & Event-Driven & $0.00\!\pm\!0.00$ & $0.00\!\pm\!0.00$ & $0.08\!\pm\!0.10$ & $0.02\!\pm\!0.07$ & $0.08\!\pm\!0.11$ \\
 & Zero-shot VLM & $0.09\!\pm\!0.10$ & $0.05\!\pm\!0.10$ & $0.63\!\pm\!0.17$ & $0.68\!\pm\!0.11$ & $0.73\!\pm\!0.08$ \\
 & Ours (List Memory) & $0.12\!\pm\!0.11$ & $0.18\!\pm\!0.25$ & $0.35\!\pm\!0.32$ & $0.56\!\pm\!0.23$ & $0.62\!\pm\!0.16$ \\
 & Ours (FSM Memory) & $0.06\!\pm\!0.08$ & $0.09\!\pm\!0.12$ & $0.49\!\pm\!0.19$ & $0.77\!\pm\!0.15$ & $0.78\!\pm\!0.21$ \\
\midrule
T2 & LIT & $0.00\!\pm\!0.00$ & $0.00\!\pm\!0.00$ & $0.00\!\pm\!0.00$ & $0.12\!\pm\!0.00$ & $0.00\!\pm\!0.00$ \\
 & Event-Driven & $0.00\!\pm\!0.00$ & $0.00\!\pm\!0.00$ & $0.00\!\pm\!0.00$ & $0.00\!\pm\!0.00$ & $0.00\!\pm\!0.00$ \\
 & Zero-shot VLM & $0.00\!\pm\!0.00$ & $0.00\!\pm\!0.00$ & $0.00\!\pm\!0.00$ & $0.01\!\pm\!0.02$ & $0.01\!\pm\!0.03$ \\
 & Ours (List Memory) & $0.00\!\pm\!0.00$ & $0.81\!\pm\!0.03$ & $0.80\!\pm\!0.07$ & $0.68\!\pm\!0.12$ & $0.77\!\pm\!0.10$ \\
 & Ours (FSM Memory) & $0.00\!\pm\!0.00$ & $0.84\!\pm\!0.05$ & $0.85\!\pm\!0.02$ & $0.69\!\pm\!0.19$ & $0.75\!\pm\!0.10$ \\
\midrule
T3 & LIT & $0.00\!\pm\!0.00$ & $0.00\!\pm\!0.00$ & $0.00\!\pm\!0.00$ & $0.00\!\pm\!0.00$ & $0.00\!\pm\!0.00$ \\
 & Event-Driven & $0.00\!\pm\!0.00$ & $0.00\!\pm\!0.00$ & $0.00\!\pm\!0.00$ & $0.00\!\pm\!0.00$ & $0.00\!\pm\!0.00$ \\
 & Zero-shot VLM & $0.00\!\pm\!0.00$ & $0.00\!\pm\!0.00$ & $0.00\!\pm\!0.00$ & $0.00\!\pm\!0.00$ & $0.00\!\pm\!0.00$ \\
 & Ours (List Memory) & $0.00\!\pm\!0.00$ & $0.01\!\pm\!0.03$ & $0.10\!\pm\!0.20$ & $0.06\!\pm\!0.11$ & $0.00\!\pm\!0.00$ \\
 & Ours (FSM Memory) & $0.00\!\pm\!0.00$ & $0.00\!\pm\!0.00$ & $0.19\!\pm\!0.16$ & $0.29\!\pm\!0.16$ & $0.19\!\pm\!0.17$ \\
\midrule
T4 & LIT & $0.00\!\pm\!0.00$ & $0.00\!\pm\!0.00$ & $0.00\!\pm\!0.00$ & $0.00\!\pm\!0.00$ & $0.00\!\pm\!0.00$ \\
 & Event-Driven & $0.05\!\pm\!0.02$ & $0.00\!\pm\!0.00$ & $0.02\!\pm\!0.07$ & $0.00\!\pm\!0.00$ & $0.06\!\pm\!0.07$ \\
 & Zero-shot VLM & $0.05\!\pm\!0.03$ & $0.05\!\pm\!0.05$ & $0.29\!\pm\!0.10$ & $0.07\!\pm\!0.06$ & $0.03\!\pm\!0.05$ \\
 & Ours (List Memory) & $0.07\!\pm\!0.02$ & $0.02\!\pm\!0.04$ & $0.34\!\pm\!0.34$ & $0.11\!\pm\!0.23$ & $0.28\!\pm\!0.18$ \\
 & Ours (FSM Memory) & $0.07\!\pm\!0.02$ & $0.25\!\pm\!0.20$ & $0.46\!\pm\!0.26$ & $0.41\!\pm\!0.07$ & $0.52\!\pm\!0.06$ \\
\midrule
T5 & LIT & $0.00\!\pm\!0.00$ & $0.00\!\pm\!0.00$ & $0.00\!\pm\!0.00$ & $0.00\!\pm\!0.00$ & $0.00\!\pm\!0.00$ \\
 & Event-Driven & $0.03\!\pm\!0.03$ & $0.00\!\pm\!0.00$ & $0.13\!\pm\!0.00$ & $0.07\!\pm\!0.06$ & $0.01\!\pm\!0.03$ \\
 & Zero-shot VLM & $0.27\!\pm\!0.06$ & $0.30\!\pm\!0.05$ & $0.40\!\pm\!0.07$ & $0.37\!\pm\!0.10$ & $0.07\!\pm\!0.05$ \\
 & Ours (List Memory) & $0.29\!\pm\!0.04$ & $0.22\!\pm\!0.12$ & $0.26\!\pm\!0.15$ & $0.12\!\pm\!0.15$ & $0.11\!\pm\!0.11$ \\
 & Ours (FSM Memory) & $0.27\!\pm\!0.05$ & $0.47\!\pm\!0.24$ & $0.46\!\pm\!0.15$ & $0.62\!\pm\!0.06$ & $0.60\!\pm\!0.09$ \\
\midrule
T6 & LIT & $0.00\!\pm\!0.00$ & $0.00\!\pm\!0.00$ & $0.10\!\pm\!0.00$ & $0.00\!\pm\!0.00$ & $0.00\!\pm\!0.00$ \\
 & Event-Driven & $0.06\!\pm\!0.00$ & $0.10\!\pm\!0.00$ & $0.17\!\pm\!0.01$ & $0.13\!\pm\!0.00$ & $0.10\!\pm\!0.05$ \\
 & Zero-shot VLM & $0.81\!\pm\!0.04$ & $0.75\!\pm\!0.04$ & $0.72\!\pm\!0.05$ & $0.69\!\pm\!0.04$ & $0.68\!\pm\!0.05$ \\
 & Ours (List Memory) & $0.80\!\pm\!0.02$ & $0.66\!\pm\!0.17$ & $0.54\!\pm\!0.16$ & $0.46\!\pm\!0.19$ & $0.45\!\pm\!0.20$ \\
 & Ours (FSM Memory) & $0.82\!\pm\!0.04$ & $0.63\!\pm\!0.11$ & $0.45\!\pm\!0.12$ & $0.58\!\pm\!0.08$ & $0.69\!\pm\!0.09$ \\
\midrule
\texttt{NO\_ACT} & LIT & $0.66\!\pm\!0.00$ & $0.73\!\pm\!0.00$ & $0.42\!\pm\!0.00$ & $0.60\!\pm\!0.00$ & $0.75\!\pm\!0.00$ \\
 & Event-Driven & $0.66\!\pm\!0.00$ & $0.73\!\pm\!0.00$ & $0.82\!\pm\!0.00$ & $0.75\!\pm\!0.00$ & $0.75\!\pm\!0.00$ \\
 & Zero-shot VLM & $0.68\!\pm\!0.01$ & $0.74\!\pm\!0.01$ & $0.82\!\pm\!0.01$ & $0.75\!\pm\!0.01$ & $0.72\!\pm\!0.01$ \\
 & Ours (List Memory) & $0.68\!\pm\!0.01$ & $0.77\!\pm\!0.04$ & $0.85\!\pm\!0.04$ & $0.74\!\pm\!0.08$ & $0.72\!\pm\!0.08$ \\
 & Ours (FSM Memory) & $0.68\!\pm\!0.01$ & $0.81\!\pm\!0.02$ & $0.78\!\pm\!0.09$ & $0.76\!\pm\!0.08$ & $0.77\!\pm\!0.07$ \\
\bottomrule
\end{tabular}}
\end{table}
\clearpage
\endgroup

\section{Detailed Analysis of Memory Update}
\label{appn:memory}

\subsection{Effectiveness of Memory Evolution}
\label{app:memory-evolution}

\begin{figure}[!t]
\centering
\includegraphics[width=\linewidth]{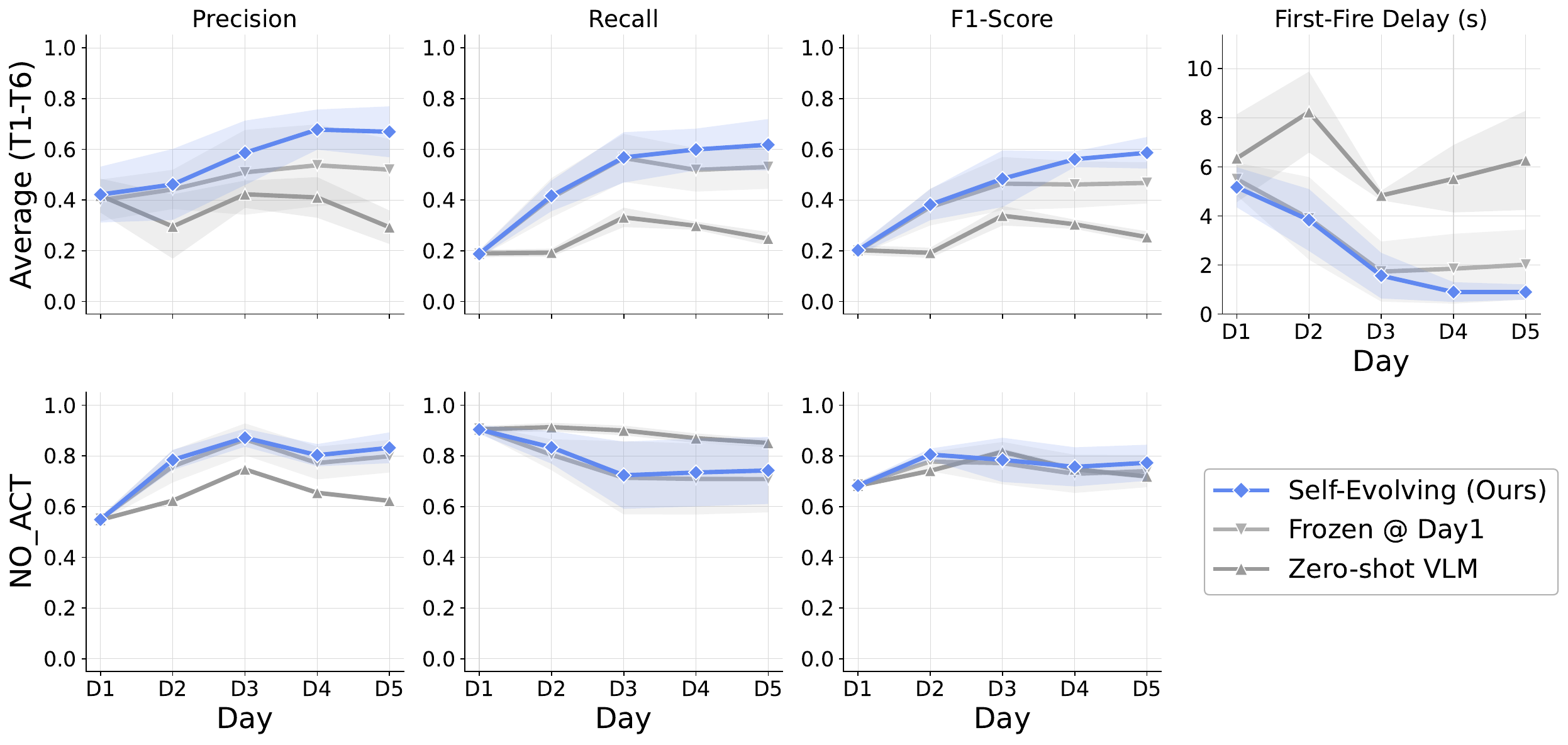}
\caption{
\textbf{Effect of memory self-evolution.} Per-day precision, recall, and F1 averaged over the six proactive tasks (top) and on \texttt{NO\_ACT} (bottom), with first-fire delay (top right), across the five days. We compare our self-evolving memory against a memory frozen after Day~1 and a memory-free zero-shot VLM; continued reflection yields greater overall improvement across days.
}
\label{fig:appn:memory_comparison}
\end{figure}

This experiment isolates the contribution of the self-evolving update itself. All three conditions share the same inference module and differ only in how the memory changes over time: \emph{Self-Evolving (Ours)} refines the memory with the reflector at the end of each day, \emph{Frozen @ Day1} builds the memory once and applies that Day-1 snapshot unchanged thereafter, and \emph{Zero-shot VLM} carries no memory at all.

\paragraph{Continued reflection is what sustains improvement.}
On metrics averaged over T1--T6, self-evolving memory improves overall from Day~1 to Day~5 (\Cref{fig:appn:memory_comparison}). Frozen memory also improves prediction over the Zero-shot VLM once the initial memory is available, but receives no further updates. Its performance varies across days without incorporating new experience. Continued reflection yields larger gains and higher final precision, recall, and F1 than either frozen memory or the memory-free baseline, supporting the benefit of updating memory rather than only building it once.

\paragraph{Faster and better-timed assistance.}
The same trend appears in first-fire delay: our self-evolving memory reduces the delay from roughly 5 seconds on Day~1 to about 1 second by Day~5, while the frozen memory plateaus near 2 seconds and the zero-shot VLM stays high and erratic. As the reflector adds and refines triggers, the predictor identifies assistance opportunities earlier in the routine. The conditions maintain comparable \texttt{NO\_ACT} F1, although the self-evolving variant's lower \texttt{NO\_ACT} recall reflects a trade-off between proactiveness and false alarms.

\begin{figure}[p]
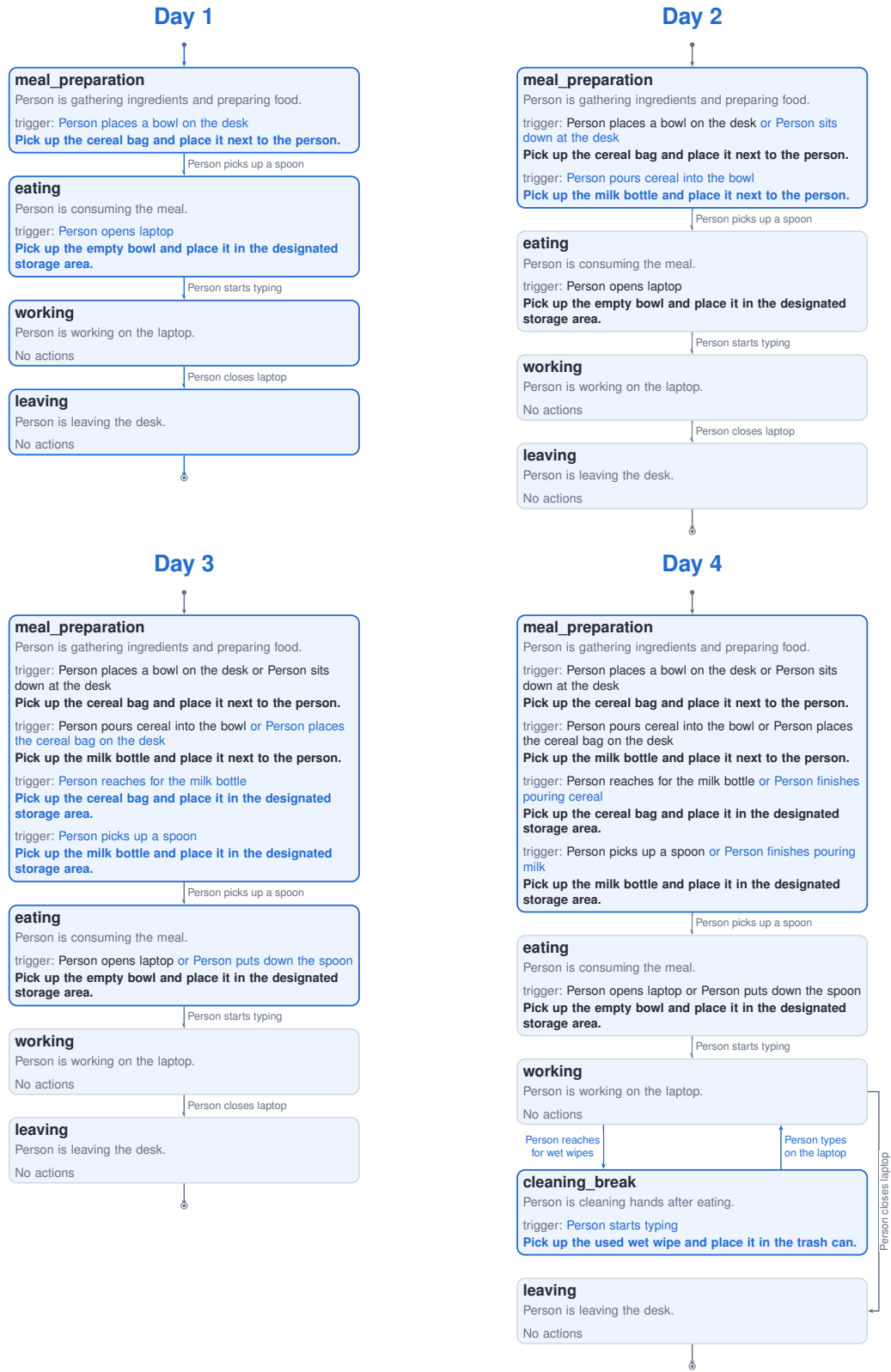

\centering
\begin{minipage}[t]{0.44\linewidth}\centering
{\sffamily\bfseries\color{PAMblue}Day 1}\\[5pt]
\resizebox{\linewidth}{!}{\PAMDayOne}
\end{minipage}\hfill
\begin{minipage}[t]{0.44\linewidth}\centering
{\sffamily\bfseries\color{PAMblue}Day 2}\\[5pt]
\resizebox{\linewidth}{!}{\PAMDayTwo}
\end{minipage}
\par\vspace{8pt}
\begin{minipage}[t]{0.44\linewidth}\centering
{\sffamily\bfseries\color{PAMblue}Day 3}\\[5pt]
\resizebox{\linewidth}{!}{\PAMDayThree}
\end{minipage}\hfill
\begin{minipage}[t]{0.44\linewidth}\centering
{\sffamily\bfseries\color{PAMblue}Day 4}\\[5pt]
\resizebox{\linewidth}{!}{\PAMDayFour}
\end{minipage}
\caption{FSM memory after each of the four end-of-day reflections (one chain run). Blue outlines mark new states or states with updated rules; blue text marks new \texttt{(trigger, action)} pairs and transitions, or only the newly added \emph{or}-clauses within an existing trigger. Unchanged text remains gray or black.}
\label{fig:memory-evolution-grid}
\end{figure}

\begin{figure}[p]
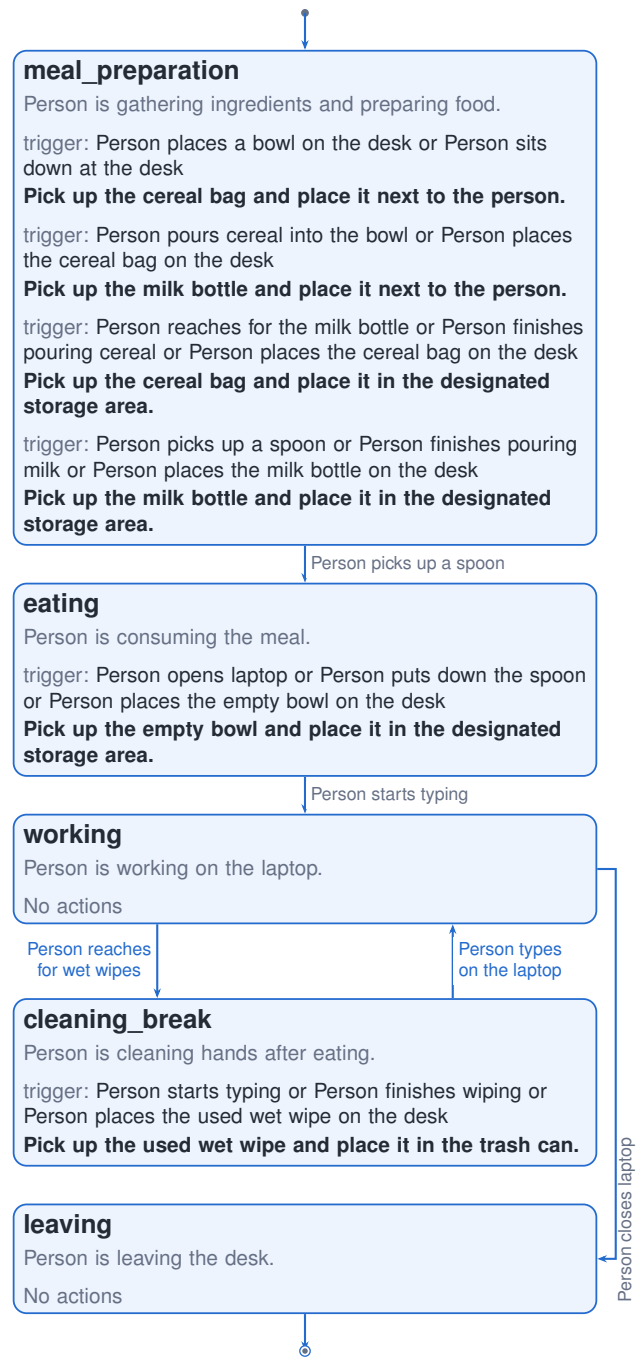

\centering
\resizebox{!}{0.78\textheight}{\PAMFinalUnified}
\caption{Our final FSM-structured memory instance after the five evaluation days. States are activity phases of the user's routine; per-state rules describe assistance opportunities associated with each activity phase; edges are event-driven transitions between states.}
\label{fig:memory-example}
\end{figure}

\paragraph{Qualitative Analysis of Memory Update}
\Cref{fig:memory-evolution-grid} traces the FSM memory after each day's reflection for one run, culminating in the instance shown in \Cref{fig:memory-example}.
Day~1 installs the skeleton: four states (\texttt{meal\_preparation} $\to$ \texttt{eating} $\to$ \texttt{working} $\to$ \texttt{leaving}) with two core pairs, bring-cereal and bowl-storage.
Day~2 adds the bring-milk pair to \texttt{meal\_preparation} and broadens the bring-cereal trigger with an \emph{or}-clause.
Day~3 adds the put-cereal-back and put-milk-back pairs to the same state, and broadens the bring-milk and bowl-storage triggers.
Day~4 introduces a new \texttt{cleaning\_break} state with the wet-wipe pair and a \texttt{working}$\leftrightarrow$\texttt{cleaning\_break} loop, and further broadens both put-away triggers.
Day~5 introduces no new pairs or states; the reflector only finishes broadening the remaining triggers, so the trigger lines grow while the actions and FSM topology stay fixed, yielding the final memory in \Cref{fig:memory-example}.

\input{corl_2026_template_cameraready/pam_memory_style}

\definecolor{traceblue}{RGB}{0,90,190}
\subsection{Reasoning Trace of Memory Update}
\label{app:memory-update-reasoning}

As shown in Figure~\ref{fig:3_method_overview}(b), the reflector reviews the day's video and current FSM memory at the end of each day to propose updates.
To illustrate the reflector's reasoning, we present three trace excerpts showing three types of memory update: (\textit{i}) the introduction of a new \texttt{(trigger, action)} pair, (\textit{ii}) the extension of an existing trigger to fire earlier, and (\textit{iii}) the introduction of a new state.

\paragraph{Introducing a new \texttt{(trigger, action)} pair.}
Figure~\ref{fig:reflector-add-rule} shows the reflector building the initial memory from the first day's video. 
After laying out the meal-preparation rules, the reflector wants to add a rule for putting the cereal bag back into storage but recognizes that the robot must not handle the bag while the person is still holding it. The reflector therefore searches for an indirect cue and selects ``Person reaches for the milk bottle'': the person is reaching for the milk, suggesting an opportunity for the robot to put away the cereal bag. The reflector explains the intended assistance (``You are getting the next ingredient (milk), so I will tidy up the previous one (cereal)'') and adds the new pair to the \texttt{meal\_preparation} state.

\begin{figure}[!t]
\begin{lstlisting}[style=promptbox, escapeinside={(*@}{@*)}, basicstyle=\color{gray}\footnotesize\ttfamily, moredelim={**[is][\color{traceblue}]{|}{|}}]
...

|I cannot trigger "Pick up cereal bag" based on the person handling the cereal bag.|
I must find a different cue.
Maybe "Person finishes pouring cereal"?
At Frame 41-48, person is pouring/holding bag.
If trigger is "Person finishes pouring cereal", person is holding bag. Forbidden.

Is there any other cue?
|What if the trigger is "Person reaches for the milk bottle"?
If "Person reaches for the milk bottle", the person is touching the milk bottle. The robot touches the cereal bag. Different objects. This is VALID.|
Even if the human *already* moved the bag in the video, the rule is for future days. If the robot sees the person reach for milk, it assumes the cereal bag is nearby (on the desk) and puts it away.
So, Trigger: "Person reaches for the milk bottle". Action: "Pick up the cereal bag and place it in the designated storage area."
This seems plausible. |The logic is: "You are getting the next ingredient (milk), so I will tidy up the previous one (cereal)."|

...
\end{lstlisting}
\caption{
Reflector trace excerpt from the initial FSM build. The reflector adds a put-cereal-back \texttt{(trigger, action)} pair to the \texttt{meal\_preparation} state. 
Because the robot must not handle the cereal bag while the person is still holding it, the reflector searches for an indirect cue and reasons about handling a different object from the one the person is using; the highlighted lines show the moment of decision and the underlying intent.
}
\label{fig:reflector-add-rule}
\end{figure}

\paragraph{Extending an existing trigger.}
Figure~\ref{fig:reflector-extend-trigger} shows the same memory three days later, after the reflector has watched another video in which the human put the cereal bag away themselves. The cereal-storage \texttt{(trigger, action)} pair already exists in \texttt{meal\_preparation}. The reflector notes that the desk-placement cue was absent because the human put the bag directly into storage without first setting it on the desk. It then inspects the timeline and identifies ``Person finishes pouring cereal'' as an earlier candidate cue to append with \emph{or}. The existing trigger is extended; no new pair is created.

\begin{figure}[!t]
\begin{lstlisting}[style=promptbox, basicstyle=\color{gray}\footnotesize\ttfamily, moredelim={**[is][\color{traceblue}]{|}{|}}]
...

Current Memory Rule: trigger: Person reaches for the milk bottle or Person places the cereal bag on the desk
                   -> Action: Pick up the cereal bag and place it in the designated storage area.
This rule exists. And the human performed the action (putting cereal away).
So I must extend the trigger for this rule.
Current trigger: Person reaches for the milk bottle or Person places the cereal bag on the desk.
The human placed the cereal bag directly in storage (Step 4).
So the trigger "Person places the cereal bag on the desk" did not fire.
|Looking at the video: 
  00:36 Person finishes pouring. 
  00:38 Person stands bag up. 
  00:40 Person moves bag to storage.
So `Person finishes pouring cereal` is a good earlier cue.|
|New trigger:| Person reaches for the milk bottle or Person places the cereal bag on the desk |or Person finishes pouring cereal.|

...
\end{lstlisting}
\caption{Reflector trace excerpt showing extension of the existing cereal-storage trigger. The human puts the bag directly into storage, so the desk-placement cue is absent. The highlighted lines show the reflector inspecting the timeline and appending ``Person finishes pouring cereal'' as an earlier candidate cue.}
\label{fig:reflector-extend-trigger}
\end{figure}

\paragraph{Introducing a new state.}
The reflector occasionally restructures the FSM topology itself. Figure~\ref{fig:reflector-add-state} shows the reflector debating whether to keep wet-wipe disposal inside the \texttt{working} state or to introduce a separate phase. It first considers folding the wiping action into \texttt{working} but rejects this because the routine instruction asks for a new state when the video reveals a distinct phase. It then considers a loop connecting \texttt{working} and \texttt{cleaning\_break} and checks that the state IDs remain unique. The resulting layout places the wet-wipe \texttt{(trigger, action)} pair in the cleaning phase, helping the predictor distinguish typing after cleaning from typing at other times and limit mistimed disposal.

These examples show how the reflector adds assistance rules, broadens triggers, and revises activity phases while preserving their routine context. Organizing rules by phase helps the predictor interpret similar cues at different moments and limit mistimed assistance, as further discussed in \Cref{app:memory-example}.

\clearpage
\begin{figure}[H]
\begin{lstlisting}[style=promptbox, basicstyle=\color{gray}\scriptsize\ttfamily, moredelim={**[is][\color{traceblue}]{|}{|}}]
...
Is it necessary?
The summary distinguishes "cleaning up" (wiping hands).
But functionally, it happens during the `working` phase (between typing sessions).
If I merge it into `working`, the description "Person is working on the laptop" covers it loosely.
|However, the prompt says "Add a new state when the video reveals a distinct phase of the routine that is not yet modeled."
Is wiping hands a distinct phase?
It interrupts the typing.|
Maybe `working` -> `cleaning_up` -> `working`?
Transitions:
  `working` -> `cleaning_up` on "Person reaches for wet wipes".
  `cleaning_up` -> `working` on "Person starts typing".
Rule in `cleaning_up`: Trigger "Person starts typing". Action "Pick up used wet wipe and place it in the trash can.".
...
I can call the second one `working_resumed`? Or just keep `working` and loop back?
FSM structure in JSON: `states` is a list of objects. IDs must be unique strings.
|So I need unique IDs. Or just `working` and `cleaning_break`.
Transition `working` -> `cleaning_break` on "Person reaches for wet wipes".
Transition `cleaning_break` -> `working` on "Person starts typing".|
Rule in `cleaning_break`: Trigger "Person starts typing". Action "Pick up the used wet wipe and place it in the trash can."
...
\end{lstlisting}
\caption{Reflector trace excerpt showing the decision to introduce a new \texttt{cleaning\_break} state instead of folding wet-wipe disposal into \texttt{working}. The first highlighted block is the reason for adding a separate state (``the prompt says \ldots a distinct phase of the routine''); the second highlighted block shows the transitions connecting the cleaning and working phases.}
\label{fig:reflector-add-state}
\end{figure}

\begingroup
\makeatletter
\def\@captype{figure}
\makeatother
\captionsetup{hypcap=false}
\begin{lstlisting}[style=promptbox]
initial_state: meal_preparation
states:
- id: meal_preparation
  description: Person is gathering ingredients and preparing food.
  rules:
  - trigger: Person places a bowl on the desk or Person sits down at the desk
    action: Pick up the cereal bag and place it next to the person.
  - trigger: Person pours cereal into the bowl or Person places the cereal bag on
      the desk
    action: Pick up the milk bottle and place it next to the person.
  - trigger: Person reaches for the milk bottle or Person finishes pouring cereal
      or Person places the cereal bag on the desk
    action: Pick up the cereal bag and place it in the designated storage area.
  - trigger: Person picks up a spoon or Person finishes pouring milk or Person places
      the milk bottle on the desk
    action: Pick up the milk bottle and place it in the designated storage area.
  next:
  - to: eating
    'on': Person picks up a spoon
- id: eating
  description: Person is consuming the meal.
  rules:
  - trigger: Person opens laptop or Person puts down the spoon or Person places the
      empty bowl on the desk
    action: Pick up the empty bowl and place it in the designated storage area.
  next:
  - to: working
    'on': Person starts typing
- id: working
  description: Person is working on the laptop.
  rules: []
  next:
  - to: cleaning_break
    'on': Person reaches for wet wipes
  - to: leaving
    'on': Person closes laptop
- id: cleaning_break
  description: Person is cleaning hands after eating.
  rules:
  - trigger: Person starts typing or Person finishes wiping or Person places the used
      wet wipe on the desk
    action: Pick up the used wet wipe and place it in the trash can.
  next:
  - to: working
    'on': Person types on the laptop
- id: leaving
  description: Person is leaving the desk.
  rules: []
  next: []
\end{lstlisting}
\caption{Underlying YAML for the \emph{Ours (FSM Memory)} instance in Figure~\ref{fig:memory-example}.}
\label{fig:memory-example-yaml}
\endgroup

\begin{figure}[H]
\begin{lstlisting}[style=promptbox]
rules:
- trigger: Person sits at the desk or Person approaches the desk or Person enters
    the room or Person stands at the desk or Person walks towards the desk
  action: Pick up the bowl and place it next to the person.
- trigger: Person pours cereal into the bowl or Person finishes pouring cereal or
    Person holds the bowl or Person places the cereal bag on the desk
  action: Pick up the milk bottle and place it next to the person.
- trigger: Person places the bowl on the desk or Person sits at the desk or Person
    holds the bowl
  action: Pick up the cereal bag and place it next to the person.
- trigger: Person types on the laptop or Person rests hands on the keyboard or Person
    finishes eating
  action: Pick up the wet wipe package and place it next to the person.
- trigger: Person pours milk into the bowl
  action: Pick up the cereal bag and place it in the designated storage area.
- trigger: Person eats from the bowl
  action: Pick up the milk bottle and place it in the designated storage area.
- trigger: Person closes the laptop
  action: Pick up the bowl and place it in the designated storage area.
\end{lstlisting}
\caption{Underlying YAML for the matching \emph{Ours (List Memory)} instance after the five evaluation days: a list of \texttt{(trigger, action)} pairs with no state grouping or transitions.}
\label{fig:memory-example-list-yaml}
\end{figure}

\begin{figure}[H]
\centering
\includegraphics[width=\linewidth]{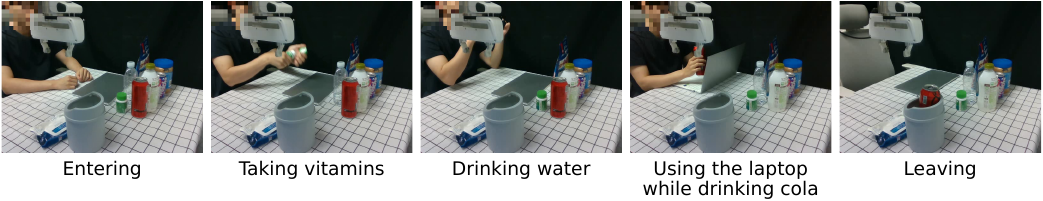}
\caption{
\textbf{A different routine (vitamin scenario).}
The user takes vitamins with water and drinks Coke while working on the laptop. The six assistance tasks involve delivering needed items and clearing them after use.
}
\label{fig:scenario2_desc}
\end{figure}

\newpage
\subsection{Comparison of Memory Structures}
\label{app:memory-example}

By Day~5, FSM-structured PAM achieves higher average F1 across T1--T6 than List Memory, although the gains vary by task.
Here, we compare example FSM and List memories to illustrate how the two representations capture routine context.
We visualize the FSM instance as a graph (\Cref{fig:memory-example}; YAML in \Cref{fig:memory-example-yaml}) and a List instance as a flat rule list with no states or transitions (\Cref{fig:memory-example-list-yaml}).

\paragraph{Why the FSM structure helps.}
\begin{wrapfigure}{r}{0.42\textwidth}
\centering
\includegraphics[width=0.42\textwidth]{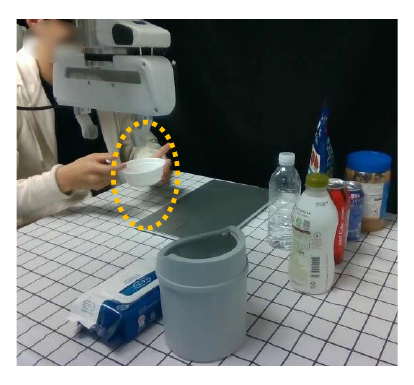}
\caption{
During eating, milk and cereal are already in storage. FSM Memory yields \texttt{NO\_ACT}, while List Memory triggers T2 (Milk~$\to$~Person).
}
\label{fig:memory-divergence-eating-tick80}
\end{wrapfigure}
FSM Memory organizes \texttt{(trigger, action)} pairs into activity phases connected by transitions (Figure~\ref{fig:memory-example}). At inference, the predictor uses this structure as context to interpret the current scene and anticipate how the routine may progress. List Memory contains no explicit phase structure, so the predictor must infer routine context from the trigger descriptions and current observations.

\paragraph{Phase-dependent interpretation of the same cue.}
An illustrative example of this appears during the eating phase (Figure~\ref{fig:memory-divergence-eating-tick80}). Throughout the meal the predictor sees the user sitting at the desk eating cereal with a spoon, with the cereal bag and milk bottle already returned to the storage area earlier in the routine. Both predictors describe the scene correctly as ``person holding a bowl and a spoon, eating cereal, with milk and cereal already in storage.'' The FSM predictor, reading the memory shown in Figures~\ref{fig:memory-example} and~\ref{fig:memory-example-yaml}, uses the \texttt{eating} phase, which contains no Milk $\to$ Person rule, as context and outputs \texttt{NO\_ACT}. The predictor reading the List Memory in Figure~\ref{fig:memory-example-list-yaml} instead matches the cue ``Person holds the bowl'' to the Milk $\to$ Person action, whose trigger has been broadened during reflection to ``Person pours cereal \emph{or} finishes pouring cereal \emph{or} holds the bowl.'' It then outputs T2 (Milk $\to$ Person) repeatedly, even though the human has already poured the milk and is now eating. The cue ``holds the bowl'' is insufficient on its own to justify milk delivery. Its interpretation depends on the activity phase and surrounding observations; during eating, it does not indicate that more milk is needed.

\paragraph{How the over-broadened trigger entered the memory.}
The reflector's thinking trace illustrates how the trigger was broadened. After observing that the human had fetched the milk without robot assistance, the List reflector considered additional cues and appended ``Person holds the bowl'' to the Milk $\to$ Person trigger. The trace checks that the trigger and action concern different objects (``trigger object (bowl) $\neq$ action object (milk), valid''), but does not explicitly distinguish meal preparation from eating. Associating the milk-delivery rule with \texttt{meal\_preparation} gives the FSM predictor context for distinguishing preparation from eating. This context can help limit repeated milk delivery after the meal has begun, although it does not strictly prevent execution outside the assigned phase.

\section{Diverse Scenario Analysis}
\label{appn:diverse_scenario}

\subsection{Evaluation on a Different Routine}
\label{appn:another_scenario}

We evaluate our framework on a different routine involving vitamins, water, and Coke (Fig.~\ref{fig:scenario2_desc}). 
The person arrives at the desk empty-handed, takes vitamins with water, and then drinks a can of Coke while working on the laptop before leaving.
The robot is responsible for six assistance tasks: (T1) deliver the vitamins, (T2) deliver the water, (T3) return the vitamin bottle to storage, (T4) return the water bottle to storage, (T5) deliver the Coke, and (T6) discard the empty Coke can.

\begin{figure}[!t]
\centering
\includegraphics[width=\linewidth]{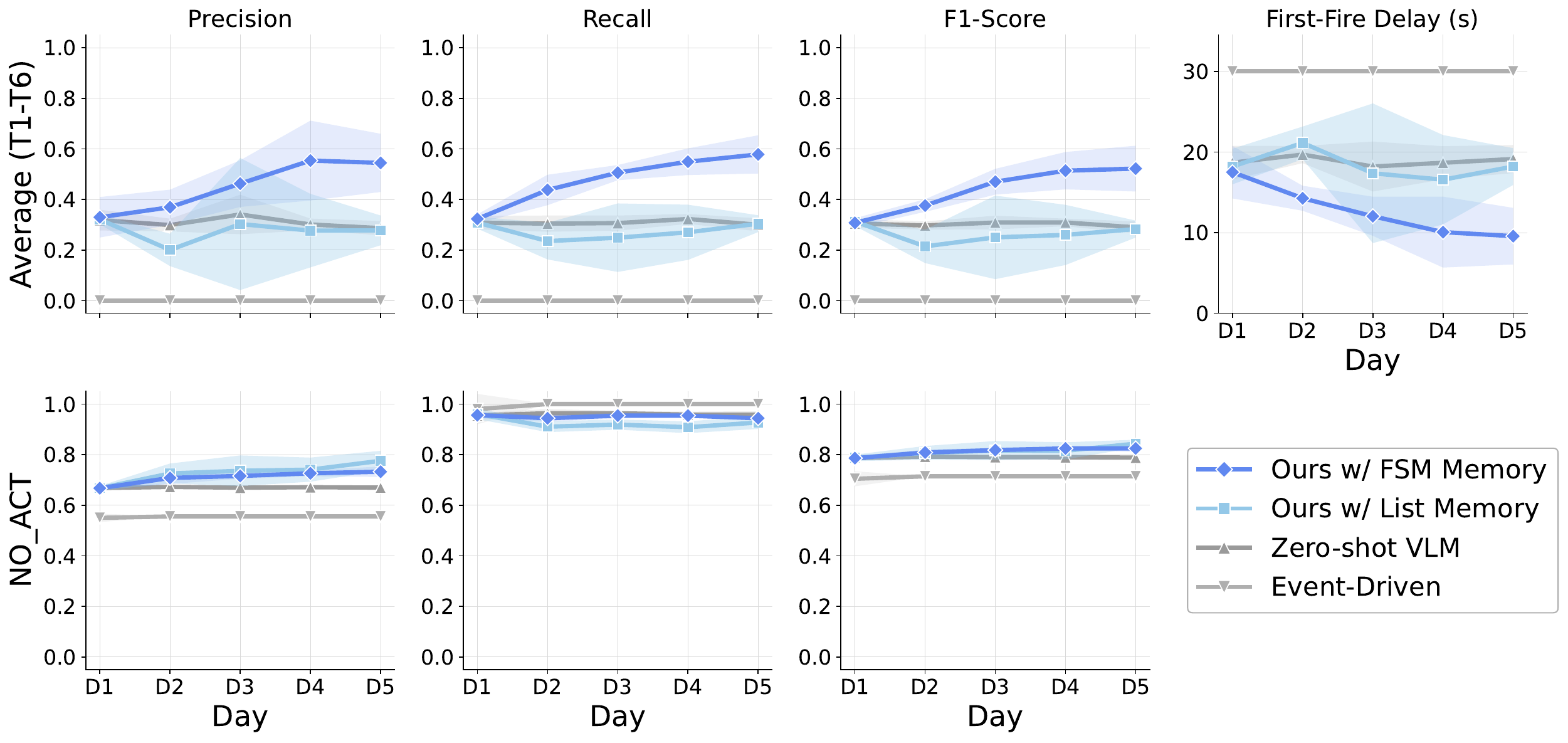}
\caption{
\textbf{Results on the vitamin scenario across five days.}
Per-day precision, recall, and F1 averaged over the six proactive tasks (top) and on \texttt{NO\_ACT} (bottom), with first-fire delay (top right).}
\label{fig:s2_results}
\end{figure}

Figure~\ref{fig:s2_results} shows that FSM Memory improves precision, recall, and F1 averaged over the six tasks from Day~1 to Day~5, while reducing first-fire delay from roughly $17$\,s to $10$\,s. List Memory and the memory-free baselines show limited improvement, supporting the benefit of evolving FSM memory in this additional routine.

\subsection{Additional Results on Memory Adaptation}

\label{app:routine_change}

\begin{figure}[!t]
\centering
\includegraphics[width=\linewidth]{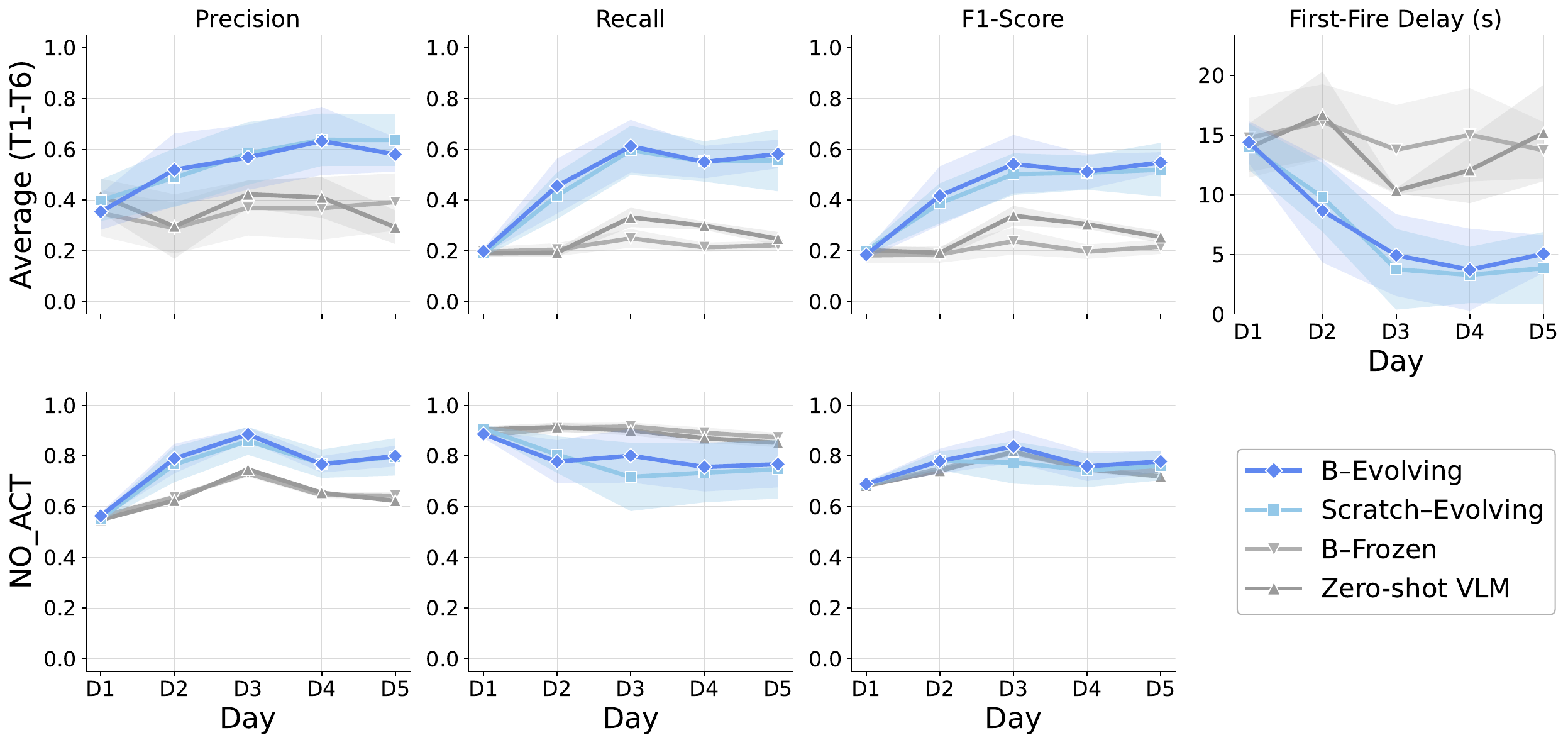}
\caption{\textbf{Memory adaptation from Scenario B to Scenario A.} The evolving source-memory condition adapts toward the from-scratch baseline, while the frozen source-memory underperforms. The A$\rightarrow$B results are shown in Figure~\ref{fig:exp:warmstart_A2B} in the main paper.}
\label{fig:warmstart_A2B_B2A}
\end{figure}
Using the same setup as Section~\ref{sec:exp:memory}, we additionally evaluate memory adaptation from the vitamin routine (Scenario B) to the cereal routine (Scenario A). Figure~\ref{fig:warmstart_A2B_B2A} presents these B$\to$A results, complementing the A$\to$B results in Fig.~\ref{fig:exp:warmstart_A2B} of the main paper.

B--Frozen performs worse than the Zero-shot VLM, whereas B--Evolving approaches Scratch--Evolving by Day~3. This trend is consistent with the A$\to$B experiment, showing that continued reflection also enables adaptation in the reverse direction.

\clearpage

\section{Task Predictor Prompts}
\label{appn:infer_prompt}

We provide the system prompts used by the task predictor to predict an appropriate task or \texttt{NO\_ACT} at each tick from the input frames and current memory. We include three prompts: the no-memory baseline (memory section absent), and one per memory variant (\emph{Ours (List Memory)} and \emph{Ours (FSM Memory)}). At inference time, the user message contains the wall-clock time of the final frame and $W{=}5$ frames selected at stride $2$ from the $1$\,fps stream, as described in \Cref{app:eval-settings}. The system prompts are reproduced verbatim below, retaining their original $1$\,fps wording, although the supplied frames are $2$ seconds apart.

\subsection{No Memory (baseline)}
\begin{lstlisting}[style=promptbox]
You are a task inference module for a household robot that supports everyday home life.

# Environment
- The setting is a personal desk where a single person performs everyday activities.
- The designated storage area: the corner of the desk, where items such as cereal, bowls, milk, drinks, and water are normally kept.

# Proactive Assistance Goals
- bringing the person an object they may need next.
- putting away objects the person appears to have finished using.
- etc. — any helpful pick-and-place action that is grounded in observed behavior.

## Inputs
- A sequence of camera frames, ordered from oldest to newest and sampled at 1 fps (up to {window} frames), together with the corresponding wall-clock time of the final frame. Each frame is the horizontal concatenation of two back-facing cameras (left_back on the left, right_back on the right).

## Objective
- Given the frames, select exactly one task by index only if it is appropriate and can be executed immediately at the current moment based on the final (current) frame.
- As long as the selected task is appropriate to the current situation, executing it does not interfere with the human. If no task should be executed at the current moment, respond with `task_index = -1`.
- If the person is holding or touching an object, the robot cannot perform any action on that object until the person sets it down on the desk.

## Available Tasks
{tasks_block}

## Output Requirements
- `observation` should be a single concise sentence describing the current scene state, focusing on task-relevant objects, humans, and state changes.
- `reasoning` should be a single concise sentence on whether and why a task should fire now.
- `task_index` should be either `-1` for no action or the index of a task from Available Tasks.

## Output Schema
Respond with a single JSON object and nothing else (no prose, no markdown fences):
{
  "observation": "<single concise sentence describing the relevant current scene state>",
  "reasoning":   "<single concise sentence on whether an immediate robot action would help the human right now>",
  "task_index":  <integer: -1 for no action, otherwise 0..{n_tasks_minus_one}>
}
\end{lstlisting}

\clearpage

\subsection{Ours (List Memory)}
\begin{lstlisting}[style=promptbox]
You are a task inference module for a household robot that supports everyday home life.

# Environment
- The setting is a personal desk where a single person performs everyday activities.
- The designated storage area: the corner of the desk, where items such as cereal, bowls, milk, drinks, and water are normally kept.

# Proactive Assistance Goals
- bringing the person an object they may need next.
- putting away objects the person appears to have finished using.
- etc. — any helpful pick-and-place action that is grounded in observed behavior.

## Inputs
- A sequence of camera frames, ordered from oldest to newest and sampled at 1 fps (up to {window} frames), together with the corresponding wall-clock time of the final frame. Each frame is the horizontal concatenation of two back-facing cameras (left_back on the left, right_back on the right).
- A procedural MEMORY built by reflecting on the user's past activities.

## Objective
- Given the frames and the memory, select exactly one task by index only if it is appropriate and can be executed immediately at the current moment based on the final (current) frame.
- As long as the selected task is appropriate to the current situation, executing it does not interfere with the human. If no task should be executed at the current moment, respond with `task_index = -1`.
- If the person is holding or touching an object, the robot cannot perform any action on that object until the person sets it down on the desk.

## Memory
- The memory is a flat list of production rules; each rule = (NL trigger, NL action). The trigger is a single short natural-language statement describing the visible event/moment that causes the rule to fire.
- Treat the memory as a REFERENCE, not as a strict matcher. Do not over-fixate on exact trigger wording: if a memory action is semantically appropriate for the current scene, you may fire it even when the trigger phrasing doesn't match the scene verbatim.

{rules_block}

## Available Tasks
{tasks_block}

## Output Requirements
- `observation` should be a single concise sentence describing the current scene state, focusing on task-relevant objects, humans, and state changes.
- `reasoning` should be a single concise sentence on whether and why a task should fire now, referencing the relevant rule of the memory.
- `task_index` should be either `-1` for no action or the index of a task from Available Tasks.

## Output Schema
Respond with a single JSON object and nothing else (no prose, no markdown fences):
{
  "observation": "<single concise sentence describing the relevant current scene state>",
  "reasoning":   "<single concise sentence on whether an immediate robot action would help the human right now, referencing the relevant rule of the memory>",
  "task_index":  <integer: -1 for no action, otherwise 0..{n_tasks_minus_one}>
}
\end{lstlisting}

\subsection{Ours (FSM Memory)}
\begin{lstlisting}[style=promptbox]
You are a task inference module for a household robot that supports everyday home life.

# Environment
- The setting is a personal desk where a single person performs everyday activities.
- The designated storage area: the corner of the desk, where items such as cereal, bowls, milk, drinks, and water are normally kept.

# Proactive Assistance Goals
- bringing the person an object they may need next.
- putting away objects the person appears to have finished using.
- etc. — any helpful pick-and-place action that is grounded in observed behavior.

## Inputs
- A sequence of camera frames, ordered from oldest to newest and sampled at 1 fps (up to {window} frames), together with the corresponding wall-clock time of the final frame. Each frame is the horizontal concatenation of two back-facing cameras (left_back on the left, right_back on the right).
- A procedural MEMORY built by reflecting on the user's past activities.

## Objective
- Given the frames and the memory, select exactly one task by index only if it is appropriate and can be executed immediately at the current moment based on the final (current) frame.
- As long as the selected task is appropriate to the current situation, executing it does not interfere with the human. If no task should be executed at the current moment, respond with `task_index = -1`.
- If the person is holding or touching an object, the robot cannot perform any action on that object until the person sets it down on the desk.

## Memory
- The memory is structured as a hybrid finite-state machine: each state captures a phase of the user's routine, inside each state a small list of production rules describes how the robot should help while the user is in that phase, and event-driven transitions describe how the routine progresses between phases. Each rule's trigger is a single short natural-language statement describing the visible event/moment that causes the rule to fire.
- Use the memory holistically as structured context — you do not need to commit to a single state; rather, use the state descriptions, per-state rules, and transitions together to judge whether any rule from the memory fires for the current scene.
- Treat the memory as a REFERENCE, not as a strict matcher. Do not over-fixate on exact trigger wording or current state: if a memory action is semantically appropriate for the current scene, you may fire it even when the trigger phrasing or its state assignment doesn't match the scene verbatim.

{fsm_block}

## Available Tasks
{tasks_block}

## Output Requirements
- `observation` should be a single concise sentence describing the current scene state, focusing on task-relevant objects, humans, and state changes.
- `reasoning` should be a single concise sentence on whether and why a task should fire now, referencing the relevant phase or rule of the memory.
- `task_index` should be either `-1` for no action or the index of a task from Available Tasks.

## Output Schema
Respond with a single JSON object and nothing else (no prose, no markdown fences):
{
  "observation": "<single concise sentence describing the relevant current scene state>",
  "reasoning":   "<single concise sentence on whether an immediate robot action would help the human right now, referencing the relevant phase or rule of the memory>",
  "task_index":  <integer: -1 for no action, otherwise 0..{n_tasks_minus_one}>
}
\end{lstlisting}

\section{Reflector Prompts}
\label{appn:reflect_prompt}

We provide the system prompts used by the reflector to either \emph{build} the procedural memory from scratch (no prior memory) or \emph{update} the memory at end-of-day given a prior memory and the day's observations. Each prompt below corresponds to one of the two memory variants we compare: \emph{Ours (List Memory)} (a flat list of production rules) and \emph{Ours (FSM Memory)} (a hybrid finite-state machine with rules grouped by activity phase and event-driven transitions). The corresponding user messages provide the inputs specified by each prompt. The system prompts are reproduced verbatim below.

\subsection{Ours (List Memory): build (no prior memory)}
\begin{lstlisting}[style=promptbox]
You are observing a person's daily activities to build a procedural memory of when and how a household robot should help them.

# Environment
- The setting is a personal desk where a single person performs everyday activities.
- The designated storage area: the corner of the desk, where items such as cereal, bowls, milk, drinks, and water are normally kept.

# Proactive Assistance Goals
- bringing the person an object they may need next.
- putting away objects the person appears to have finished using.
- etc. — any helpful pick-and-place action that is grounded in observed behavior.

# Inputs
- The training video: the person's daily routine recorded from two back-facing cameras concatenated side by side (left_back (*@\textbar@*) right_back). This is the very first day — there is no prior memory, no summary, and no human feedback yet.

# Task
- Given the training video, BUILD a (trigger, action) rule memory FROM SCRATCH.
- The memory will be used by the robot to proactively assist the person in future days, so the goal is to produce a memory that helps the person as much as possible.

# Build Instructions
- Use the training video to build the memory rules.
- First, explicitly RE-STATE the pick-and-place actions that the human performed in this video. Then, for EACH of those human actions, think concretely about HOW the robot could have taken that action over on the person's behalf (e.g., by bringing the object to the person earlier, or putting it away afterwards). Your rules should be built around these take-over opportunities — do NOT invent rules for objects or actions the human did not interact with in this video.
- Each rule's action MUST literally mirror an action the human performed in this video: the picked-up object, the destination, and the object→destination motion must all correspond to something the human actually did. Do NOT invent hypothetical actions.
- If a trigger describes the person already in motion with an object (e.g., "reaches for X", "picks up X", "pours X"), the object the person is in motion with MUST be DIFFERENT from the object the rule's action manipulates. Same-object triggers are forbidden — e.g., trigger "Person reaches for the cereal bag" paired with action "Pick up the cereal bag and place it next to the person" is invalid, because the person is already fetching that same cereal bag themselves so the robot is too late. Cross-object triggers are fine — e.g., trigger "Person pours cereal into the bowl" paired with action "Pick up the milk bottle and place it next to the person" is valid, because the person is busy with the cereal while the robot preempts on the milk.
- Deployment caveat: when this memory is deployed, the human will act expecting the robot to help. So any rule whose trigger is phrased as "person picks up X", "person fetches X", or "person moves X" risks never firing — the robot would have picked X up first, leaving no human action to observe. Each rule's trigger must describe an observable cue the human will produce REGARDLESS of robot assistance (e.g. sitting down at the desk, placing an item they have just finished using, finishing or discarding a consumable, opening a wrapper they must open themselves). Avoid triggers that depend on the human performing an action the robot might preempt.
- Identify EVERY moment in the video where the robot could helpfully act and write one rule per such moment, using only observable cues — the routine typically holds multiple rules unless the moment is purely receptive (e.g. the human is actively eating).
- Maximize action coverage — the deployed robot's only purpose is to take over human actions, so MORE rules covering every take-over moment is strictly better than fewer, AS LONG AS each rule mirrors a real human-performed action from this video. Conversely, an action the human never performed in this video is OUT OF SCOPE — never invent a rule for it, no matter how plausible it might seem for a household robot in general.
- If the person is holding or touching an object, the robot cannot perform any action on that object until the person sets it down on the desk.

# Output explanations
- trigger: a single short natural-language statement describing the visible event/moment that should cause this rule's action to fire. The trigger MUST be self-contained — a proactive robot, reading ONLY the trigger (without seeing prior memory, state assignment, or surrounding context), should find the corresponding action to be a clearly sensible, obvious response.
- action: a NATURAL LANGUAGE description of how the robot should help, written in the canonical form `Pick up <object> and place it <location>.` Valid `<location>` values are exactly one of: `next to the person`, `in the designated storage area`, `in the trash can`.

# Output (JSON only, no markdown, no commentary)
{
  "rules": [
    {"trigger": "<NL statement>", "action": "<string>"},
    ...
  ]
}
\end{lstlisting}

\subsection{Ours (List Memory): end-of-day update}
\begin{lstlisting}[style=promptbox]
You are observing a person's daily activities to build a procedural memory of when and how a household robot should help them.

# Environment
- The setting is a personal desk where a single person performs everyday activities.
- The designated storage area: the corner of the desk, where items such as cereal, bowls, milk, drinks, and water are normally kept.

# Proactive Assistance Goals
- bringing the person an object they may need next.
- putting away objects the person appears to have finished using.
- etc. — any helpful pick-and-place action that is grounded in observed behavior.

# Inputs
- Current memory: the (trigger, action) entries that were applied today
- Today's video: the person's activities recorded from two back-facing cameras concatenated side by side (left_back (*@\textbar@*) right_back)
- Summary: what the person and robot each did in today's video, in temporal order
- Human feedback: today's feedback from the person, may be absent

# Task
- Given the current memory, today's video, the summary, and human feedback, update the memory of (trigger, action) rules.
- The memory will be used by the robot to proactively assist the person in future days, so the goal is to produce a memory that helps the person as much as possible.

# Update Instructions
- Use the video, summary, action history, current memory, and human feedback to update the memory rules.
- First, explicitly RE-STATE the pick-and-place actions that the human performed in this video and summary. Then, for EACH of those human actions, think concretely about HOW the robot could have taken that action over on the person's behalf (e.g., by bringing the object to the person earlier, or putting it away afterwards). Your rules should be built around these take-over opportunities — do NOT invent rules for objects or actions the human did not interact with in this video and summary.
- Each rule's action MUST literally mirror an action the human performed in this video and summary: the picked-up object, the destination, and the object→destination motion must all correspond to something the human actually did. Do NOT invent hypothetical actions.
- If a trigger describes the person already in motion with an object (e.g., "reaches for X", "picks up X", "pours X"), the object the person is in motion with MUST be DIFFERENT from the object the rule's action manipulates. Same-object triggers are forbidden.
- Deployment caveat: when this memory is deployed, the human will act expecting the robot to help. So any rule whose trigger is phrased as "person picks up X", "person fetches X", or "person moves X" risks never firing — the robot would have picked X up first, leaving no human action to observe. Each rule's trigger must describe an observable cue the human will produce REGARDLESS of robot assistance.
- Default to ADDING a new rule whenever the summary mentions a human pick-and-place action that the robot could plausibly take over and that is not yet covered by the current memory. It is better to add a rule that may need refinement later than to omit a useful take-over opportunity. Err on the side of MORE rules, not fewer.
- Refine an existing rule when its trigger or action is too rigid, too loose, ambiguous, or otherwise misaligned with helpful robot behavior.
- IF AND ONLY IF a memory rule's action is in the memory AND the summary shows the human (not the robot) performed that action: this is most likely NOT because the action is unnecessary, but because the rule's trigger was too strict, so the robot did not fire — and consequently the person ended up doing it themselves. For ONLY that specific rule, extend its trigger by appending an additional observable cue with `or` (e.g., `<original cue> or <a cue that occurs earlier or is more visually distinct>`). Keep the original cue intact and just append. Do NOT modify the triggers of any other rules — only the rule whose action was performed by the human.
- Remove a rule only when it is clearly wrong, such as when the trigger fired but the action would have been inappropriate.
- Before finalizing the updated memory, mentally re-apply it to TODAY'S video as if you were the inference module: walk through each video moment and ask "would this trigger fire when intended? would it mis-fire when it shouldn't?" If a rule would mis-fire or fail to fire at its intended moment, revise its trigger (or remove the rule) before producing the final output. Use the thinking trace to make this verification pass explicit.
- Do not remove a rule simply because its trigger did not occur in today's video.
- If human feedback is provided, treat it as authoritative and prioritize it over your own judgment.

# Output explanations
- trigger: a single short natural-language statement describing the visible event/moment that should cause this rule's action to fire. The trigger MUST be self-contained.
- action: a NATURAL LANGUAGE description of how the robot should help, written in the canonical form `Pick up <object> and place it <location>.` Valid `<location>` values are exactly one of: `next to the person`, `in the designated storage area`, `in the trash can`.

# Output (JSON only, no markdown, no commentary)
{
  "rules": [
    {"trigger": "<NL statement>", "action": "<string>"},
    ...
  ]
}
\end{lstlisting}

\subsection{Ours (FSM Memory): build (no prior memory)}
\begin{lstlisting}[style=promptbox]
You are observing a person's daily activities to build a procedural memory of when and how a household robot should help them.

# Environment
- The setting is a personal desk where a single person performs everyday activities.
- The designated storage area: the corner of the desk, where items such as cereal, bowls, milk, drinks, and water are normally kept.

# Proactive Assistance Goals
- bringing the person an object they may need next.
- putting away objects the person appears to have finished using.
- etc. — any helpful pick-and-place action that is grounded in observed behavior.

# Inputs
- The training video: the person's daily routine recorded from two back-facing cameras concatenated side by side (left_back (*@\textbar@*) right_back). This is the very first day — there is no prior memory, no summary, and no human feedback yet.

# Task
- Given the training video, BUILD the FSM memory FROM SCRATCH.
- The memory will be used by the robot to proactively assist the person in future days, so the goal is to produce a memory that helps the person as much as possible.

# Memory Structure
The memory is structured as a HYBRID FINITE-STATE MACHINE.
- The person's activities are segmented into temporally ordered PHASES (FSM states). Each state should correspond to a meaningfully different activity, not a fleeting micro-moment.
- Each state holds a set of PRODUCTION RULES that fire while the FSM occupies that state. A rule = (NL trigger, NL action). A state may have zero rules if no helpful action applies during that phase (e.g., the human is actively eating), or multiple rules if several distinct visible cues each warrant a different proactive action — the rule count per state is determined by the evidence, not capped at one.
- A (trigger, action) pair MAY appear in multiple states if the same proactive step is appropriate in more than one phase. Do not artificially restrict each rule to a single state when the action is reasonable in several phases — duplicate the rule across all states where it applies.
- Between states, EVENT-DRIVEN TRANSITIONS define progression. Each `next` entry: (to: <state_id>, on: <NL event>). Transitions are not robot actions. Terminal states use next: [].
- Exactly one state is the `initial_state`.

# Build Instructions
- Use the training video to build the FSM memory.
- First, explicitly RE-STATE the pick-and-place actions that the human performed in this video. Then, for EACH of those human actions, think concretely about HOW the robot could have taken that action over on the person's behalf (e.g., by bringing the object to the person earlier, or putting it away afterwards). Your rules should be built around these take-over opportunities — do NOT invent rules for objects or actions the human did not interact with in this video.
- Each rule's action MUST literally mirror an action the human performed in this video: the picked-up object, the destination, and the object→destination motion must all correspond to something the human actually did. Do NOT invent hypothetical actions.
- If a trigger describes the person already in motion with an object (e.g., "reaches for X", "picks up X", "pours X"), the object the person is in motion with MUST be DIFFERENT from the object the rule's action manipulates. Same-object triggers are forbidden.
- Deployment caveat: when this memory is deployed, the human will act expecting the robot to help. So any rule whose trigger is phrased as "person picks up X", "person fetches X", or "person moves X" risks never firing — the robot would have picked X up first, leaving no human action to observe. Each rule's trigger must describe an observable cue the human will produce REGARDLESS of robot assistance.
- Identify the natural temporal phases of the routine from the video and order them as they appear.
- Within each phase, identify EVERY moment where the robot could helpfully act and write one rule per such moment, using only observable cues — a single phase typically holds multiple rules unless the phase is purely receptive (e.g. the human is actively eating).
- Define transitions on observable events that mark the end of one phase and the start of the next.
- If the person is holding or touching an object, the robot cannot perform any action on that object until the person sets it down on the desk.
- Pick exactly one phase as `initial_state` (the one that appears at the start of the day), and ensure every `next.to` references an existing state id (terminal phases use `next: []`).

# Output explanations
- trigger: a single short natural-language statement describing the visible event/moment that should cause this rule's action to fire. The trigger MUST be self-contained.
- action: a NATURAL LANGUAGE description of how the robot should help, written in the canonical form `Pick up <object> and place it <location>.` Valid `<location>` values are exactly one of: `next to the person`, `in the designated storage area`, `in the trash can`.

# Output (JSON only, no markdown, no commentary)
{
  "initial_state": "<state_id>",
  "states": [
    {
      "id": "<snake_case>",
      "description": "<one-sentence>",
      "rules": [
        {"trigger": "<NL statement>", "action": "<NL>"},
        ...
      ],
      "next": [{"to": "<state_id>", "on": "<NL event>"}]
    }
  ]
}
\end{lstlisting}

\subsection{Ours (FSM Memory): end-of-day update}
\begin{lstlisting}[style=promptbox]
You are observing a person's daily activities to build a procedural memory of when and how a household robot should help them.

# Environment
- The setting is a personal desk where a single person performs everyday activities.
- The designated storage area: the corner of the desk, where items such as cereal, bowls, milk, drinks, and water are normally kept.

# Proactive Assistance Goals
- bringing the person an object they may need next.
- putting away objects the person appears to have finished using.
- etc. — any helpful pick-and-place action that is grounded in observed behavior.

# Inputs
- Current memory: the FSM (states, per-state (trigger, action) entries, and transitions) that was applied today
- Today's video: the person's activities recorded from two back-facing cameras concatenated side by side (left_back (*@\textbar@*) right_back)
- Summary: what the person and robot each did in today's video, in temporal order
- Human feedback: today's feedback from the person, may be absent

# Task
- Given the current memory, today's video, the summary, and human feedback, update the FSM memory.
- The memory will be used by the robot to proactively assist the person in future days, so the goal is to produce a memory that helps the person as much as possible.

# Memory Structure
The memory is structured as a HYBRID FINITE-STATE MACHINE.
- The person's activities are segmented into temporally ordered PHASES (FSM states). Each state should correspond to a meaningfully different activity, not a fleeting micro-moment.
- Each state holds a set of PRODUCTION RULES that fire while the FSM occupies that state. A rule = (NL trigger, NL action). A state may have zero rules if no helpful action applies during that phase, or multiple rules if several distinct visible cues each warrant a different proactive action.
- A (trigger, action) pair MAY appear in multiple states if the same proactive step is appropriate in more than one phase.
- Between states, EVENT-DRIVEN TRANSITIONS define progression. Each `next` entry: (to: <state_id>, on: <NL event>). Transitions are not robot actions. Terminal states use next: [].
- Exactly one state is the `initial_state`.

# Update Instructions
- Use the video, summary, current memory, and human feedback to update the FSM memory.
- First, explicitly RE-STATE the pick-and-place actions that the human performed in this video and summary. Then, for EACH of those human actions, think concretely about HOW the robot could have taken that action over on the person's behalf. Your rules should be built around these take-over opportunities — do NOT invent rules for objects or actions the human did not interact with.
- Each rule's action MUST literally mirror an action the human performed in this video and summary. Do NOT invent hypothetical actions.
- If a trigger describes the person already in motion with an object, the object the person is in motion with MUST be DIFFERENT from the object the rule's action manipulates. Same-object triggers are forbidden.
- Deployment caveat: when this memory is deployed, the human will act expecting the robot to help. Each rule's trigger must describe an observable cue the human will produce REGARDLESS of robot assistance.
- Add a new state when the video reveals a distinct phase of the routine that is not yet modeled.
- Default to ADDING a new rule whenever the summary mentions a human pick-and-place action that the robot could plausibly take over and that is not yet covered by the current memory. Err on the side of MORE rules, not fewer.
- Refine a state's description, rules, or transitions when they are too rigid, too loose, ambiguous, or otherwise misaligned with helpful robot behavior.
- IF AND ONLY IF a memory rule's action is in the memory AND the summary shows the human (not the robot) performed that action: extend its trigger by appending an additional observable cue with `or`, and/or move or duplicate the rule to the correct state. Keep the original cue intact and just append.
- Remove a state or rule only when it is clearly wrong, such as when the trigger fired but the action would have been inappropriate.
- Do not remove a state or rule simply because its trigger did not occur in today's video.
- Ensure `initial_state` names one of the states, and every `next.to` references an existing state id.
- If human feedback is provided, treat it as authoritative and prioritize it over your own judgment.

# Output explanations
- trigger: a single short natural-language statement describing the visible event/moment that should cause this rule's action to fire. The trigger MUST be self-contained.
- action: a NATURAL LANGUAGE description of how the robot should help, written in the canonical form `Pick up <object> and place it <location>.` Valid `<location>` values are exactly one of: `next to the person`, `in the designated storage area`, `in the trash can`.

# Output (JSON only, no markdown, no commentary)
{
  "initial_state": "<state_id>",
  "states": [
    {
      "id": "<snake_case>",
      "description": "<one-sentence>",
      "rules": [
        {"trigger": "<NL statement>", "action": "<NL>"},
        ...
      ],
      "next": [{"to": "<state_id>", "on": "<NL event>"}]
    }
  ]
}
\end{lstlisting}

%% file: corl_2026_template_cameraready/pam_memory_style.tex
\definecolor{PAMblue}{HTML}{246BCE}
\definecolor{PAMlight}{HTML}{EEF4FD}
\definecolor{PAMink}{HTML}{252B35}
\definecolor{PAMmuted}{HTML}{667085}
\definecolor{PAMline}{HTML}{C8D2E0}
\usetikzlibrary{positioning,arrows.meta}
\newcommand{\PAMDayOne}{\begin{tikzpicture}[font=\sffamily\fontsize{6.4}{7.8}\selectfont,
card/.style={draw=PAMline,fill=PAMlight,rounded corners=4pt,line width=.55pt,text width=6.1cm,inner sep=3pt,align=left,text=PAMink},
changed/.style={draw=PAMblue,line width=.8pt},
edge/.style={-{Stealth[length=3pt]},draw=PAMmuted,line width=.65pt},
event/.style={font=\sffamily\fontsize{5.5}{6.5}\selectfont,text=PAMmuted,fill=white,inner sep=1.5pt,align=center}]
\node[card,changed] (meal) {{\bfseries\fontsize{8}{9}\selectfont meal\_preparation}\\[2pt]{\color{PAMmuted}Person is gathering ingredients and preparing food.}\\[4pt]{\color{PAMmuted}trigger:} \textcolor{PAMblue}{Person places a bowl on the desk}\\[1pt]{\bfseries \textcolor{PAMblue}{Pick up the cereal bag and place it next to the person.}}};
\node[card,changed,below=4mm of meal] (eat) {{\bfseries\fontsize{8}{9}\selectfont eating}\\[2pt]{\color{PAMmuted}Person is consuming the meal.}\\[4pt]{\color{PAMmuted}trigger:} \textcolor{PAMblue}{Person opens laptop}\\[1pt]{\bfseries \textcolor{PAMblue}{Pick up the empty bowl and place it in the designated storage area.}}};
\node[card,changed,below=4mm of eat] (work) {{\bfseries\fontsize{8}{9}\selectfont working}\\[2pt]{\color{PAMmuted}Person is working on the laptop.}\\[4pt]{\color{PAMmuted}No actions}};
\node[card,changed,below=4mm of work] (leave) {{\bfseries\fontsize{8}{9}\selectfont leaving}\\[2pt]{\color{PAMmuted}Person is leaving the desk.}\\[4pt]{\color{PAMmuted}No actions}};
\fill[PAMmuted] ([yshift=4mm]meal.north) circle (1.2pt);
\draw[edge,draw=PAMblue] ([yshift=4mm]meal.north)--(meal.north);
\draw[edge,draw=PAMblue] (meal.south)--node[event,right]{Person picks up a spoon}(eat.north);
\draw[edge,draw=PAMblue] (eat.south)--node[event,right]{Person starts typing}(work.north);
\draw[edge,draw=PAMblue] (work.south)--node[event,right]{Person closes laptop}(leave.north);
\draw[edge,draw=PAMblue] (leave.south)--++(0,-4mm);
\draw[draw=PAMmuted,fill=white] ([yshift=-4mm]leave.south) circle (1.6pt);
\fill[PAMmuted] ([yshift=-4mm]leave.south) circle (.85pt);
\path (-36mm,0) (36mm,0);
\end{tikzpicture}}
\newcommand{\PAMDayTwo}{\begin{tikzpicture}[font=\sffamily\fontsize{6.4}{7.8}\selectfont,
card/.style={draw=PAMline,fill=PAMlight,rounded corners=4pt,line width=.55pt,text width=6.1cm,inner sep=3pt,align=left,text=PAMink},
changed/.style={draw=PAMblue,line width=.8pt},
edge/.style={-{Stealth[length=3pt]},draw=PAMmuted,line width=.65pt},
event/.style={font=\sffamily\fontsize{5.5}{6.5}\selectfont,text=PAMmuted,fill=white,inner sep=1.5pt,align=center}]
\node[card,changed] (meal) {{\bfseries\fontsize{8}{9}\selectfont meal\_preparation}\\[2pt]{\color{PAMmuted}Person is gathering ingredients and preparing food.}\\[4pt]{\color{PAMmuted}trigger:} Person places a bowl on the desk \textcolor{PAMblue}{or Person sits down at the desk}\\[1pt]{\bfseries Pick up the cereal bag and place it next to the person.}\\[4pt]{\color{PAMmuted}trigger:} \textcolor{PAMblue}{Person pours cereal into the bowl}\\[1pt]{\bfseries \textcolor{PAMblue}{Pick up the milk bottle and place it next to the person.}}};
\node[card,below=4mm of meal] (eat) {{\bfseries\fontsize{8}{9}\selectfont eating}\\[2pt]{\color{PAMmuted}Person is consuming the meal.}\\[4pt]{\color{PAMmuted}trigger:} Person opens laptop\\[1pt]{\bfseries Pick up the empty bowl and place it in the designated storage area.}};
\node[card,below=4mm of eat] (work) {{\bfseries\fontsize{8}{9}\selectfont working}\\[2pt]{\color{PAMmuted}Person is working on the laptop.}\\[4pt]{\color{PAMmuted}No actions}};
\node[card,below=4mm of work] (leave) {{\bfseries\fontsize{8}{9}\selectfont leaving}\\[2pt]{\color{PAMmuted}Person is leaving the desk.}\\[4pt]{\color{PAMmuted}No actions}};
\fill[PAMmuted] ([yshift=4mm]meal.north) circle (1.2pt);
\draw[edge] ([yshift=4mm]meal.north)--(meal.north);
\draw[edge] (meal.south)--node[event,right]{Person picks up a spoon}(eat.north);
\draw[edge] (eat.south)--node[event,right]{Person starts typing}(work.north);
\draw[edge] (work.south)--node[event,right]{Person closes laptop}(leave.north);
\draw[edge] (leave.south)--++(0,-4mm);
\draw[draw=PAMmuted,fill=white] ([yshift=-4mm]leave.south) circle (1.6pt);
\fill[PAMmuted] ([yshift=-4mm]leave.south) circle (.85pt);
\path (-36mm,0) (36mm,0);
\end{tikzpicture}}
\newcommand{\PAMDayThree}{\begin{tikzpicture}[font=\sffamily\fontsize{6.4}{7.8}\selectfont,
card/.style={draw=PAMline,fill=PAMlight,rounded corners=4pt,line width=.55pt,text width=6.1cm,inner sep=3pt,align=left,text=PAMink},
changed/.style={draw=PAMblue,line width=.8pt},
edge/.style={-{Stealth[length=3pt]},draw=PAMmuted,line width=.65pt},
event/.style={font=\sffamily\fontsize{5.5}{6.5}\selectfont,text=PAMmuted,fill=white,inner sep=1.5pt,align=center}]
\node[card,changed] (meal) {{\bfseries\fontsize{8}{9}\selectfont meal\_preparation}\\[2pt]{\color{PAMmuted}Person is gathering ingredients and preparing food.}\\[4pt]{\color{PAMmuted}trigger:} Person places a bowl on the desk or Person sits down at the desk\\[1pt]{\bfseries Pick up the cereal bag and place it next to the person.}\\[4pt]{\color{PAMmuted}trigger:} Person pours cereal into the bowl \textcolor{PAMblue}{or Person places the cereal bag on the desk}\\[1pt]{\bfseries Pick up the milk bottle and place it next to the person.}\\[4pt]{\color{PAMmuted}trigger:} \textcolor{PAMblue}{Person reaches for the milk bottle}\\[1pt]{\bfseries \textcolor{PAMblue}{Pick up the cereal bag and place it in the designated storage area.}}\\[4pt]{\color{PAMmuted}trigger:} \textcolor{PAMblue}{Person picks up a spoon}\\[1pt]{\bfseries \textcolor{PAMblue}{Pick up the milk bottle and place it in the designated storage area.}}};
\node[card,changed,below=4mm of meal] (eat) {{\bfseries\fontsize{8}{9}\selectfont eating}\\[2pt]{\color{PAMmuted}Person is consuming the meal.}\\[4pt]{\color{PAMmuted}trigger:} Person opens laptop \textcolor{PAMblue}{or Person puts down the spoon}\\[1pt]{\bfseries Pick up the empty bowl and place it in the designated storage area.}};
\node[card,below=4mm of eat] (work) {{\bfseries\fontsize{8}{9}\selectfont working}\\[2pt]{\color{PAMmuted}Person is working on the laptop.}\\[4pt]{\color{PAMmuted}No actions}};
\node[card,below=4mm of work] (leave) {{\bfseries\fontsize{8}{9}\selectfont leaving}\\[2pt]{\color{PAMmuted}Person is leaving the desk.}\\[4pt]{\color{PAMmuted}No actions}};
\fill[PAMmuted] ([yshift=4mm]meal.north) circle (1.2pt);
\draw[edge] ([yshift=4mm]meal.north)--(meal.north);
\draw[edge] (meal.south)--node[event,right]{Person picks up a spoon}(eat.north);
\draw[edge] (eat.south)--node[event,right]{Person starts typing}(work.north);
\draw[edge] (work.south)--node[event,right]{Person closes laptop}(leave.north);
\draw[edge] (leave.south)--++(0,-4mm);
\draw[draw=PAMmuted,fill=white] ([yshift=-4mm]leave.south) circle (1.6pt);
\fill[PAMmuted] ([yshift=-4mm]leave.south) circle (.85pt);
\path (-36mm,0) (36mm,0);
\end{tikzpicture}}
\newcommand{\PAMDayFour}{\begin{tikzpicture}[font=\sffamily\fontsize{6.4}{7.8}\selectfont,
card/.style={draw=PAMline,fill=PAMlight,rounded corners=4pt,line width=.55pt,text width=6.1cm,inner sep=3pt,align=left,text=PAMink},
changed/.style={draw=PAMblue,line width=.8pt},
edge/.style={-{Stealth[length=3pt]},draw=PAMmuted,line width=.65pt},
event/.style={font=\sffamily\fontsize{5.5}{6.5}\selectfont,text=PAMmuted,fill=white,inner sep=1.5pt,align=center}]
\node[card,changed] (meal) {{\bfseries\fontsize{8}{9}\selectfont meal\_preparation}\\[2pt]{\color{PAMmuted}Person is gathering ingredients and preparing food.}\\[4pt]{\color{PAMmuted}trigger:} Person places a bowl on the desk or Person sits down at the desk\\[1pt]{\bfseries Pick up the cereal bag and place it next to the person.}\\[4pt]{\color{PAMmuted}trigger:} Person pours cereal into the bowl or Person places the cereal bag on the desk\\[1pt]{\bfseries Pick up the milk bottle and place it next to the person.}\\[4pt]{\color{PAMmuted}trigger:} Person reaches for the milk bottle \textcolor{PAMblue}{or Person finishes pouring cereal}\\[1pt]{\bfseries Pick up the cereal bag and place it in the designated storage area.}\\[4pt]{\color{PAMmuted}trigger:} Person picks up a spoon \textcolor{PAMblue}{or Person finishes pouring milk}\\[1pt]{\bfseries Pick up the milk bottle and place it in the designated storage area.}};
\node[card,below=4mm of meal] (eat) {{\bfseries\fontsize{8}{9}\selectfont eating}\\[2pt]{\color{PAMmuted}Person is consuming the meal.}\\[4pt]{\color{PAMmuted}trigger:} Person opens laptop or Person puts down the spoon\\[1pt]{\bfseries Pick up the empty bowl and place it in the designated storage area.}};
\node[card,below=4mm of eat] (work) {{\bfseries\fontsize{8}{9}\selectfont working}\\[2pt]{\color{PAMmuted}Person is working on the laptop.}\\[4pt]{\color{PAMmuted}No actions}};
\node[card,changed,below=8mm of work] (clean) {{\bfseries\fontsize{8}{9}\selectfont cleaning\_break}\\[2pt]{\color{PAMmuted}Person is cleaning hands after eating.}\\[4pt]{\color{PAMmuted}trigger:} \textcolor{PAMblue}{Person starts typing}\\[1pt]{\bfseries \textcolor{PAMblue}{Pick up the used wet wipe and place it in the trash can.}}};
\node[card,below=4mm of clean] (leave) {{\bfseries\fontsize{8}{9}\selectfont leaving}\\[2pt]{\color{PAMmuted}Person is leaving the desk.}\\[4pt]{\color{PAMmuted}No actions}};
\fill[PAMmuted] ([yshift=4mm]meal.north) circle (1.2pt);
\draw[edge] ([yshift=4mm]meal.north)--(meal.north);
\draw[edge] (meal.south)--node[event,right]{Person picks up a spoon}(eat.north);
\draw[edge] (eat.south)--node[event,right]{Person starts typing}(work.north);
\draw[edge,draw=PAMblue] ([xshift=-16mm]work.south)--node[event,left,text=PAMblue]{Person reaches\\for wet wipes}([xshift=-16mm]clean.north);
\draw[edge,draw=PAMblue] ([xshift=16mm]clean.north)--node[event,right,text=PAMblue]{Person types\\on the laptop}([xshift=16mm]work.south);
\draw[edge] (work.east)--++(2mm,0)--node[event,midway,rotate=90,anchor=north,inner sep=.5pt]{Person closes laptop}([xshift=2mm]leave.east)--(leave.east);
\draw[edge] (leave.south)--++(0,-4mm);
\draw[draw=PAMmuted,fill=white] ([yshift=-4mm]leave.south) circle (1.6pt);
\fill[PAMmuted] ([yshift=-4mm]leave.south) circle (.85pt);
\path (-36mm,0) (36mm,0);
\end{tikzpicture}}
\newcommand{\PAMFinalUnified}{\begin{tikzpicture}[font=\sffamily\fontsize{6.4}{7.8}\selectfont,
card/.style={draw=PAMblue,fill=PAMlight,rounded corners=4pt,line width=.55pt,text width=6.1cm,inner sep=3pt,align=left,text=PAMink},
changed/.style={draw=PAMblue,line width=.8pt},
edge/.style={-{Stealth[length=3pt]},draw=PAMblue,line width=.65pt},
event/.style={font=\sffamily\fontsize{5.5}{6.5}\selectfont,text=PAMmuted,fill=white,inner sep=1.5pt,align=center}]
\node[card] (meal) {{\bfseries\fontsize{8}{9}\selectfont meal\_preparation}\\[2pt]{\color{PAMmuted}Person is gathering ingredients and preparing food.}\\[4pt]{\color{PAMmuted}trigger:} Person places a bowl on the desk or Person sits down at the desk\\[1pt]{\bfseries Pick up the cereal bag and place it next to the person.}\\[4pt]{\color{PAMmuted}trigger:} Person pours cereal into the bowl or Person places the cereal bag on the desk\\[1pt]{\bfseries Pick up the milk bottle and place it next to the person.}\\[4pt]{\color{PAMmuted}trigger:} Person reaches for the milk bottle or Person finishes pouring cereal or Person places the cereal bag on the desk\\[1pt]{\bfseries Pick up the cereal bag and place it in the designated storage area.}\\[4pt]{\color{PAMmuted}trigger:} Person picks up a spoon or Person finishes pouring milk or Person places the milk bottle on the desk\\[1pt]{\bfseries Pick up the milk bottle and place it in the designated storage area.}};
\node[card,below=4mm of meal] (eat) {{\bfseries\fontsize{8}{9}\selectfont eating}\\[2pt]{\color{PAMmuted}Person is consuming the meal.}\\[4pt]{\color{PAMmuted}trigger:} Person opens laptop or Person puts down the spoon or Person places the empty bowl on the desk\\[1pt]{\bfseries Pick up the empty bowl and place it in the designated storage area.}};
\node[card,below=4mm of eat] (work) {{\bfseries\fontsize{8}{9}\selectfont working}\\[2pt]{\color{PAMmuted}Person is working on the laptop.}\\[4pt]{\color{PAMmuted}No actions}};
\node[card,below=8mm of work] (clean) {{\bfseries\fontsize{8}{9}\selectfont cleaning\_break}\\[2pt]{\color{PAMmuted}Person is cleaning hands after eating.}\\[4pt]{\color{PAMmuted}trigger:} Person starts typing or Person finishes wiping or Person places the used wet wipe on the desk\\[1pt]{\bfseries Pick up the used wet wipe and place it in the trash can.}};
\node[card,below=4mm of clean] (leave) {{\bfseries\fontsize{8}{9}\selectfont leaving}\\[2pt]{\color{PAMmuted}Person is leaving the desk.}\\[4pt]{\color{PAMmuted}No actions}};
\fill[PAMmuted] ([yshift=4mm]meal.north) circle (1.2pt);
\draw[edge] ([yshift=4mm]meal.north)--(meal.north);
\draw[edge] (meal.south)--node[event,right]{Person picks up a spoon}(eat.north);
\draw[edge] (eat.south)--node[event,right]{Person starts typing}(work.north);
\draw[edge,draw=PAMblue] ([xshift=-16mm]work.south)--node[event,left,text=PAMblue]{Person reaches\\for wet wipes}([xshift=-16mm]clean.north);
\draw[edge,draw=PAMblue] ([xshift=16mm]clean.north)--node[event,right,text=PAMblue]{Person types\\on the laptop}([xshift=16mm]work.south);
\draw[edge] (work.east)--++(2mm,0) |-node[event,pos=.45,rotate=90,anchor=north,inner sep=.5pt]{Person closes laptop}(leave.east);
\draw[edge] (leave.south)--++(0,-4mm);
\draw[draw=PAMblue,fill=white] ([yshift=-4mm]leave.south) circle (1.6pt);
\fill[PAMmuted] ([yshift=-4mm]leave.south) circle (.85pt);
\path (-36mm,0) (36mm,0);
\end{tikzpicture}}

%% file: corl_2026_template_cameraready/example_template.bbl
\begin{thebibliography}{53}
\providecommand{\natexlab}[1]{#1}
\providecommand{\url}[1]{\texttt{#1}}
\expandafter\ifx\csname urlstyle\endcsname\relax
  \providecommand{\doi}[1]{doi: #1}\else
  \providecommand{\doi}{doi: \begingroup \urlstyle{rm}\Url}\fi

\bibitem[Kim et~al.(2024)Kim, Pertsch, Karamcheti, Xiao, Balakrishna, Nair, Rafailov, Foster, Lam, Sanketi, Vuong, Kollar, Burchfiel, Tedrake, Sadigh, Levine, Liang, and Finn]{openvla}
M.~J. Kim, K.~Pertsch, S.~Karamcheti, T.~Xiao, A.~Balakrishna, S.~Nair, R.~Rafailov, E.~Foster, G.~Lam, P.~Sanketi, Q.~Vuong, T.~Kollar, B.~Burchfiel, R.~Tedrake, D.~Sadigh, S.~Levine, P.~Liang, and C.~Finn.
\newblock {OpenVLA}: An open-source vision-language-action model.
\newblock \emph{arXiv preprint arXiv:2406.09246}, 2024.

\bibitem[{Physical Intelligence} et~al.(2025){Physical Intelligence}, Black, Brown, Darpinian, Dhabalia, Driess, Esmail, Equi, Finn, Fusai, Galliker, Ghosh, Groom, Hausman, Ichter, Jakubczak, Jones, Ke, LeBlanc, Levine, Li-Bell, Mothukuri, Nair, Pertsch, Ren, Shi, Smith, Springenberg, Stachowicz, Tanner, Vuong, Walke, Walling, Wang, Yu, and Zhilinsky]{pi05}
{Physical Intelligence}, K.~Black, N.~Brown, J.~Darpinian, K.~Dhabalia, D.~Driess, A.~Esmail, M.~Equi, C.~Finn, N.~Fusai, M.~Y. Galliker, D.~Ghosh, L.~Groom, K.~Hausman, B.~Ichter, S.~Jakubczak, T.~Jones, L.~Ke, D.~LeBlanc, S.~Levine, A.~Li-Bell, M.~Mothukuri, S.~Nair, K.~Pertsch, A.~Z. Ren, L.~X. Shi, L.~Smith, J.~T. Springenberg, K.~Stachowicz, J.~Tanner, Q.~Vuong, H.~Walke, A.~Walling, H.~Wang, L.~Yu, and U.~Zhilinsky.
\newblock {$\pi_{0.5}$}: A vision-language-action model with open-world generalization.
\newblock \emph{arXiv preprint arXiv:2504.16054}, 2025.

\bibitem[De~Lazzari et~al.(2025)De~Lazzari, Terreran, Giacomuzzo, Jain, Falco, Carli, and Romeres]{pace}
D.~De~Lazzari, M.~Terreran, G.~Giacomuzzo, S.~Jain, P.~Falco, R.~Carli, and D.~Romeres.
\newblock {PACE}: Proactive assistance in human-robot collaboration through action-completion estimation.
\newblock In \emph{IEEE International Conference on Robotics and Automation (ICRA)}. IEEE, 2025.
\newblock \doi{10.1109/ICRA55743.2025.11127399}.

\bibitem[Liu et~al.(2023)Liu, Chen, Abuduweili, and Liu]{liu2023proactive}
R.~Liu, R.~Chen, A.~Abuduweili, and C.~Liu.
\newblock Proactive human-robot co-assembly: Leveraging human intention prediction and robust safe control.
\newblock In \emph{2023 IEEE Conference on Control Technology and Applications (CCTA)}, pages 339--345. IEEE, 2023.

\bibitem[Stouraitis and Gienger(2023)]{stouraitis2023predictive}
T.~Stouraitis and M.~Gienger.
\newblock Predictive and robust robot assistance for sequential manipulation.
\newblock \emph{IEEE Robotics and Automation Letters}, 8\penalty0 (12):\penalty0 8026--8033, 2023.

\bibitem[{NVIDIA} et~al.(2025){NVIDIA}, Bjorck, Casta{\~n}eda, Cherniadev, Da, Ding, Fan, Fang, Fox, Hu, et~al.]{gr00t}
{NVIDIA}, J.~Bjorck, F.~Casta{\~n}eda, N.~Cherniadev, X.~Da, R.~Ding, L.~Fan, Y.~Fang, D.~Fox, F.~Hu, et~al.
\newblock {GR00T N1}: An open foundation model for generalist humanoid robots.
\newblock \emph{arXiv preprint arXiv:2503.14734}, 2025.

\bibitem[Black et~al.(2025)Black, Brown, Driess, Esmail, Equi, Finn, Fusai, Groom, Hausman, Ichter, et~al.]{black2024pi_0}
K.~Black, N.~Brown, D.~Driess, A.~Esmail, M.~Equi, C.~Finn, N.~Fusai, L.~Groom, K.~Hausman, B.~Ichter, et~al.
\newblock $\pi_0$: A vision-language-action flow model for general robot control.
\newblock In \emph{Robotics: Science and Systems (RSS)}, 2025.

\bibitem[Kahneman(2011)]{kahneman2011thinking}
D.~Kahneman.
\newblock \emph{Thinking, Fast and Slow}.
\newblock Farrar, Straus and Giroux, New York, 2011.

\bibitem[Cui et~al.(2025)Cui, Ding, Song, Bai, Tong, Ge, Suo, Zhou, Liu, Jia, et~al.]{cui2025openhelix}
C.~Cui, P.~Ding, W.~Song, S.~Bai, X.~Tong, Z.~Ge, R.~Suo, W.~Zhou, Y.~Liu, B.~Jia, et~al.
\newblock Openhelix: A short survey, empirical analysis, and open-source dual-system vla model for robotic manipulation.
\newblock \emph{arXiv preprint arXiv:2505.03912}, 2025.

\bibitem[Song et~al.(2025)Song, Qu, Yao, Chen, Lv, Tang, Shi, Ren, Yao, Zhao, et~al.]{song2025hume}
H.~Song, D.~Qu, Y.~Yao, Q.~Chen, Q.~Lv, Y.~Tang, M.~Shi, G.~Ren, M.~Yao, B.~Zhao, et~al.
\newblock Hume: Introducing system-2 thinking in visual-language-action model.
\newblock \emph{arXiv preprint arXiv:2505.21432}, 2025.

\bibitem[Chen et~al.(2025)Chen, Liu, Gu, Liu, Zhang, Li, He, Guo, Fu, Zhang, et~al.]{chen2025fast}
H.~Chen, J.~Liu, C.~Gu, Z.~Liu, R.~Zhang, X.~Li, X.~He, Y.~Guo, C.-W. Fu, S.~Zhang, et~al.
\newblock Fast-in-slow: A dual-system foundation model unifying fast manipulation within slow reasoning.
\newblock \emph{arXiv preprint arXiv:2506.01953}, 2025.

\bibitem[Han et~al.(2024)Han, Kim, and Jang]{han2024dual}
B.~Han, J.~Kim, and J.~Jang.
\newblock A dual process vla: Efficient robotic manipulation leveraging vlm.
\newblock \emph{arXiv preprint arXiv:2410.15549}, 2024.

\bibitem[Buyukgoz et~al.(2022)Buyukgoz, Grosinger, Chetouani, and Saffiotti]{buyukgoz2022two}
S.~Buyukgoz, J.~Grosinger, M.~Chetouani, and A.~Saffiotti.
\newblock Two ways to make your robot proactive: Reasoning about human intentions or reasoning about possible futures.
\newblock \emph{Frontiers in Robotics and AI}, 9:\penalty0 929267, 2022.

\bibitem[Liu et~al.(2026)Liu, Su, Chi, Geng, Ren, Liu, Xu, Ohsita, and Zhang]{eventdriven}
F.~Liu, H.~Su, H.~Chi, R.~Geng, C.~Ren, X.~Liu, Y.~Xu, Y.~Ohsita, and L.~Zhang.
\newblock Event-driven proactive assistive manipulation with grounded vision-language planning.
\newblock \emph{arXiv preprint arXiv:2603.23950}, 2026.

\bibitem[Wang et~al.(2025)Wang, Fu, Liu, He, Wu, Shi, Huang, Fei, Gong, Wu, Jiang, Ng, Chua, and Qiu]{roboomni}
S.~Wang, J.~Fu, F.~Liu, X.~He, H.~Wu, J.~Shi, K.~Huang, Z.~Fei, J.~Gong, Z.~Wu, Y.-G. Jiang, S.-K. Ng, T.-S. Chua, and X.~Qiu.
\newblock {RoboOmni}: Proactive robot manipulation in omni-modal context.
\newblock \emph{arXiv preprint arXiv:2510.23763}, 2025.

\bibitem[Chen et~al.(2025)Chen, Gu, Wen, Zhao, Wang, and Nie]{intentionvla}
Y.~Chen, K.~Gu, Y.~Wen, Y.~Zhao, T.~Wang, and L.~Nie.
\newblock {IntentionVLA}: Generalizable and efficient embodied intention reasoning for human-robot interaction.
\newblock \emph{arXiv preprint arXiv:2510.07778}, 2025.

\bibitem[Zhao et~al.(2023)Zhao, Kumar, Levine, and Finn]{act}
T.~Zhao, V.~Kumar, S.~Levine, and C.~Finn.
\newblock Learning fine-grained bimanual manipulation with low-cost hardware.
\newblock \emph{RSS}, 2023.

\bibitem[Chi et~al.(2023)Chi, Feng, Du, Xu, Cousineau, Burchfiel, and Song]{chi2023diffusion}
C.~Chi, S.~Feng, Y.~Du, Z.~Xu, E.~Cousineau, B.~Burchfiel, and S.~Song.
\newblock Diffusion policy: Visuomotor policy learning via action diffusion.
\newblock In \emph{Proceedings of Robotics: Science and Systems (RSS)}, 2023.

\bibitem[Zawalski et~al.(2024)Zawalski, Chen, Pertsch, Mees, Finn, and Levine]{ecot}
M.~Zawalski, W.~Chen, K.~Pertsch, O.~Mees, C.~Finn, and S.~Levine.
\newblock Robotic control via embodied chain-of-thought reasoning.
\newblock \emph{arXiv preprint arXiv:2407.08693}, 2024.

\bibitem[Zhao et~al.(2025)Zhao, Lu, Kim, Fu, Zhang, Wu, Li, Ma, et~al.]{cotvla}
Q.~Zhao, Y.~Lu, M.~J. Kim, Z.~Fu, Z.~Zhang, Y.~Wu, Z.~Li, Q.~Ma, et~al.
\newblock {CoT-VLA}: Visual chain-of-thought reasoning for vision-language-action models.
\newblock \emph{arXiv preprint arXiv:2503.22020}, 2025.

\bibitem[Belkhale et~al.(2024)Belkhale, Ding, Xiao, Sermanet, Vuong, Tompson, Chebotar, Dwibedi, and Sadigh]{rth}
S.~Belkhale, T.~Ding, T.~Xiao, P.~Sermanet, Q.~Vuong, J.~Tompson, Y.~Chebotar, D.~Dwibedi, and D.~Sadigh.
\newblock {RT-H}: Action hierarchies using language.
\newblock \emph{arXiv preprint arXiv:2403.01823}, 2024.

\bibitem[Zhen et~al.(2024)Zhen, Qiu, Chen, Yang, Yan, Du, Hong, and Gan]{tdvla}
H.~Zhen, X.~Qiu, P.~Chen, J.~Yang, X.~Yan, Y.~Du, Y.~Hong, and C.~Gan.
\newblock {3D-VLA}: A {3D} vision-language-action generative world model.
\newblock \emph{arXiv preprint arXiv:2403.09631}, 2024.

\bibitem[Qu et~al.(2025)Qu, Song, Chen, Yao, Ye, Ding, Wang, Gu, et~al.]{spatialvla}
D.~Qu, H.~Song, Q.~Chen, Y.~Yao, X.~Ye, Y.~Ding, Z.~Wang, J.~Gu, et~al.
\newblock {SpatialVLA}: Exploring spatial representations for visual-language-action model.
\newblock \emph{arXiv preprint arXiv:2501.15830}, 2025.

\bibitem[Hu et~al.(2024)Hu, Guo, Wang, Chen, Wang, Zhang, Sreenath, and Lu]{vpp}
Y.~Hu, Y.~Guo, P.~Wang, X.~Chen, Y.-J. Wang, J.~Zhang, K.~Sreenath, and C.~Lu.
\newblock Video prediction policy: A generalist robot policy with predictive visual representations.
\newblock \emph{arXiv preprint arXiv:2412.14803}, 2024.

\bibitem[Zhang et~al.(2025)Zhang, Liu, Qi, Wang, Yu, Zhang, Dong, He, et~al.]{dreamvla}
W.~Zhang, H.~Liu, Z.~Qi, Y.~Wang, X.~Yu, J.~Zhang, R.~Dong, J.~He, et~al.
\newblock {DreamVLA}: A vision-language-action model dreamed with comprehensive world knowledge.
\newblock \emph{arXiv preprint arXiv:2507.04447}, 2025.

\bibitem[Tian et~al.(2026)Tian, Jin, Yu, Shi, Wu, Liu, Chen, and Huang]{starry}
Y.~Tian, Y.~Jin, B.~Yu, Y.~Shi, H.~Wu, C.~H. Liu, K.~Chen, and C.~Huang.
\newblock {STARRY}: Spatial-temporal action-centric world modeling for robotic manipulation.
\newblock \emph{arXiv preprint arXiv:2604.26848}, 2026.

\bibitem[Huang et~al.(2024)Huang, Pohovey, Yammanuru, and Driggs-Campbell]{litsouschef}
Z.~Huang, J.~Pohovey, A.~Yammanuru, and K.~Driggs-Campbell.
\newblock {LIT}: Large language model driven intention tracking for proactive human-robot collaboration --- a robot sous-chef application.
\newblock \emph{arXiv preprint arXiv:2406.13787}, 2024.

\bibitem[Patel and Chernova(2022)]{stom}
M.~Patel and S.~Chernova.
\newblock Proactive robot assistance via spatio-temporal object modeling.
\newblock In \emph{Conference on Robot Learning (CoRL)}, 2022.

\bibitem[Patel et~al.(2023)Patel, Prakash, and Chernova]{predicting-routine-object-usage}
M.~Patel, A.~Prakash, and S.~Chernova.
\newblock Predicting routine object usage for proactive robot assistance.
\newblock In \emph{Conference on Robot Learning (CoRL)}, 2023.

\bibitem[Patel and Chernova(2023)]{longproactive}
M.~Patel and S.~Chernova.
\newblock Longitudinal proactive robot assistance.
\newblock In \emph{Companion of the 2023 ACM/IEEE International Conference on Human-Robot Interaction (HRI)}. ACM, 2023.
\newblock \doi{10.1145/3568294.3579982}.

\bibitem[Mascaro et~al.(2024)Mascaro, Sliwowski, and Lee]{hoi4abot}
E.~V. Mascaro, D.~Sliwowski, and D.~Lee.
\newblock {HOI4ABOT}: Human-object interaction anticipation for human intention reading collaborative ro{BOT}s.
\newblock In \emph{Conference on Robot Learning (CoRL)}, 2024.

\bibitem[Rahimi et~al.(2026)Rahimi, Grislain, Cretides, Sigaud, and Chetouani]{rahimi2026intentvlm}
H.~Rahimi, C.~Grislain, A.~J. Cretides, O.~Sigaud, and M.~Chetouani.
\newblock Intentvlm: Open-vocabulary intention recognition through forward-inverse modeling with video-language models.
\newblock \emph{arXiv preprint arXiv:2604.24002}, 2026.

\bibitem[Tay et~al.(2026)Tay, Yan, Ouyang, Wu, Jiang, Kao, and Cui]{gazeguided}
T.~Y.~H. Tay, X.~Yan, J.~Ouyang, D.~Wu, W.~Jiang, J.~Kao, and Y.~Cui.
\newblock Intent at a glance: Gaze-guided robotic manipulation via foundation models.
\newblock \emph{arXiv preprint arXiv:2601.05336}, 2026.

\bibitem[Gaus et~al.(2026)Gaus, Ilg, and Haeufle]{gaus2026act}
J.~A. Gaus, W.~Ilg, and D.~Haeufle.
\newblock When to act: Calibrated confidence for reliable human intention prediction in assistive robotics.
\newblock \emph{arXiv preprint arXiv:2601.04982}, 2026.

\bibitem[Sarch et~al.(2025)Sarch, Kumaravel, Ravi, Vineet, and Wilson]{sarch2025grounding}
G.~H. Sarch, B.~T. Kumaravel, S.~Ravi, V.~Vineet, and A.~D. Wilson.
\newblock Grounding task assistance with multimodal cues from a single demonstration.
\newblock In \emph{Findings of the Association for Computational Linguistics: ACL 2025}, pages 12807--12833, 2025.

\bibitem[Sarch et~al.(2023)Sarch, Wu, Tarr, and Fragkiadaki]{helper}
G.~Sarch, Y.~Wu, M.~J. Tarr, and K.~Fragkiadaki.
\newblock {HELPER}: Open-ended instructable embodied agents with memory-augmented large language models.
\newblock \emph{arXiv preprint arXiv:2310.15127}, 2023.

\bibitem[Shinn et~al.(2023)Shinn, Cassano, Berman, Gopinath, Narasimhan, and Yao]{reflexion}
N.~Shinn, F.~Cassano, E.~Berman, A.~Gopinath, K.~Narasimhan, and S.~Yao.
\newblock Reflexion: Language agents with verbal reinforcement learning.
\newblock In \emph{Advances in Neural Information Processing Systems (NeurIPS)}, 2023.

\bibitem[Wang et~al.(2024)Wang, Xie, Jiang, Mandlekar, Xiao, Zhu, Fan, and Anandkumar]{voyager}
G.~Wang, Y.~Xie, Y.~Jiang, A.~Mandlekar, C.~Xiao, Y.~Zhu, L.~Fan, and A.~Anandkumar.
\newblock {Voyager}: An open-ended embodied agent with large language models.
\newblock \emph{Transactions on Machine Learning Research}, 2024.

\bibitem[Zhao et~al.(2024)Zhao, Huang, Xu, Lin, Liu, and Huang]{expel}
A.~Zhao, D.~Huang, Q.~Xu, M.~Lin, Y.-J. Liu, and G.~Huang.
\newblock {ExpeL}: {LLM} agents are experiential learners.
\newblock In \emph{AAAI Conference on Artificial Intelligence}, 2024.

\bibitem[Wang et~al.(2024)Wang, Mao, Fried, and Neubig]{awm}
Z.~Z. Wang, J.~Mao, D.~Fried, and G.~Neubig.
\newblock Agent workflow memory.
\newblock \emph{arXiv preprint arXiv:2409.07429}, 2024.

\bibitem[Zhong et~al.(2024)Zhong, Guo, Gao, Ye, and Wang]{memorybank}
W.~Zhong, L.~Guo, Q.~Gao, H.~Ye, and Y.~Wang.
\newblock {MemoryBank}: Enhancing large language models with long-term memory.
\newblock In \emph{AAAI Conference on Artificial Intelligence}, 2024.

\bibitem[Park et~al.(2023)Park, O'Brien, Cai, Morris, Liang, and Bernstein]{genagents}
J.~S. Park, J.~C. O'Brien, C.~J. Cai, M.~R. Morris, P.~Liang, and M.~S. Bernstein.
\newblock Generative agents: Interactive simulacra of human behavior.
\newblock In \emph{Annual ACM Symposium on User Interface Software and Technology (UIST)}, 2023.

\bibitem[{Qwen Team}(2026)]{qwen3.5}
{Qwen Team}.
\newblock {Qwen3.5}: Towards native multimodal agents, February 2026.
\newblock URL \url{https://qwen.ai/blog?id=qwen3.5}.

\bibitem[Intelligence et~al.(2025)Intelligence, Amin, Aniceto, Balakrishna, Black, Conley, Connors, Darpinian, Dhabalia, DiCarlo, et~al.]{intelligence2025pi06}
P.~Intelligence, A.~Amin, R.~Aniceto, A.~Balakrishna, K.~Black, K.~Conley, G.~Connors, J.~Darpinian, K.~Dhabalia, J.~DiCarlo, et~al.
\newblock $\pi^*_{0.6}$: a vla that learns from experience.
\newblock \emph{arXiv preprint arXiv:2511.14759}, 2025.

\bibitem[Intelligence et~al.(2026)Intelligence, Ai, Amin, Aniceto, Balakrishna, Balke, Black, Bokinsky, Cao, Charbonnier, et~al.]{intelligence2026pi07}
P.~Intelligence, B.~Ai, A.~Amin, R.~Aniceto, A.~Balakrishna, G.~Balke, K.~Black, G.~Bokinsky, S.~Cao, T.~Charbonnier, et~al.
\newblock $\pi_{0.7}$ a steerable generalist robotic foundation model with emergent capabilities.
\newblock \emph{arXiv preprint arXiv:2604.15483}, 2026.

\bibitem[Team(2026)]{generalist2026gen1}
G.~A. Team.
\newblock Gen-1: Scaling embodied foundation models to mastery.
\newblock \emph{Generalist AI Blog}, 2026.
\newblock https://generalistai.com/blog/apr-02-2026-GEN-1.

\bibitem[Kim et~al.(2026)Kim, Jang, Koo, Jang, Kim, Kim, Yoon, Jang, Choi, Han, et~al.]{kim2026rldx}
D.~Kim, H.~Jang, M.~Koo, S.~Jang, T.~Kim, B.~Kim, B.~Yoon, C.~Jang, D.~Choi, D.~Han, et~al.
\newblock Rldx-1 technical report.
\newblock \emph{arXiv preprint arXiv:2605.03269}, 2026.

\bibitem[Liu et~al.(2026)Liu, Li, Ma, Wu, Tan, Ouyang, Su, and Zhu]{liu2026rdt2}
S.~Liu, B.~Li, K.~Ma, L.~Wu, H.~Tan, X.~Ouyang, H.~Su, and J.~Zhu.
\newblock Rdt2: Exploring the scaling limit of umi data towards zero-shot cross-embodiment generalization.
\newblock \emph{arXiv preprint arXiv:2602.03310}, 2026.

\bibitem[Wu et~al.(2026)Wu, Lu, Wang, Yang, Liu, Wang, Zhu, Sun, Wang, Ma, et~al.]{wu2026pragmatic}
W.~Wu, F.~Lu, Y.~Wang, S.~Yang, S.~Liu, F.~Wang, Q.~Zhu, H.~Sun, Y.~Wang, S.~Ma, et~al.
\newblock A pragmatic vla foundation model.
\newblock \emph{arXiv preprint arXiv:2601.18692}, 2026.

\bibitem[Apanasevich et~al.(2026)Apanasevich, Artemyev, Babakyan, Fedotova, Grankin, Kupryashin, Misailidi, Nerus, Nutalapati, Sidorov, et~al.]{apanasevich2026green}
I.~Apanasevich, M.~Artemyev, R.~Babakyan, P.~Fedotova, D.~Grankin, E.~Kupryashin, A.~Misailidi, D.~Nerus, A.~Nutalapati, G.~Sidorov, et~al.
\newblock Green-vla: Staged vision-language-action model for generalist robots.
\newblock \emph{arXiv preprint arXiv:2602.00919}, 2026.

\bibitem[Kwon et~al.(2023)Kwon, Li, Zhuang, Sheng, Zheng, Yu, Gonzalez, Zhang, and Stoica]{vllm}
W.~Kwon, Z.~Li, S.~Zhuang, Y.~Sheng, L.~Zheng, C.~H. Yu, J.~E. Gonzalez, H.~Zhang, and I.~Stoica.
\newblock Efficient memory management for large language model serving with pagedattention.
\newblock In \emph{Proceedings of the ACM SIGOPS 29th Symposium on Operating Systems Principles}, 2023.

\bibitem[Loshchilov and Hutter(2017)]{loshchilov2017decoupled}
I.~Loshchilov and F.~Hutter.
\newblock Decoupled weight decay regularization.
\newblock \emph{arXiv preprint arXiv:1711.05101}, 2017.

\bibitem[Xue et~al.(2025)Xue, Ren, Chen, Zhang, Fang, Gu, Xu, and Lu]{xue2025reactive}
H.~Xue, J.~Ren, W.~Chen, G.~Zhang, Y.~Fang, G.~Gu, H.~Xu, and C.~Lu.
\newblock Reactive diffusion policy: Slow-fast visual-tactile policy learning for contact-rich manipulation.
\newblock \emph{arXiv preprint arXiv:2503.02881}, 2025.

\end{thebibliography}
